\documentclass[]{siamart220329}

\usepackage[margin=1in]{geometry}
\usepackage{lipsum}
\usepackage{amsfonts}
\usepackage{multirow}
\usepackage{graphicx}
\usepackage{epstopdf}
\usepackage{amsmath} 
\usepackage{hyperref}

\usepackage{algorithm} 
\usepackage{algpseudocode} % for algorithmic block

\usepackage{pgfplots}
\pgfplotsset{compat=1.16}
\usepgfplotslibrary{groupplots}
\usepgfplotslibrary{statistics}
\usepackage{color}
\usepackage{colortbl}
\usepackage{xcolor}

\usepackage{tikz}
\usetikzlibrary{fit}
\usetikzlibrary{shapes.multipart}
\usetikzlibrary{shapes, arrows.meta, positioning}
\usetikzlibrary{matrix}

\usepackage{booktabs} % For better looking tables

\definecolor{tblrowcolor}{rgb}{0.9, 0.9, 0.9}  % Gray color for table highlighting

\makeatletter
\def\pann@Cref@sec{Section}
\def\pann@Cref@fig{Figure}
\def\pann@Cref@eq{Equation}
\def\pann@Cref@tab{Table}
\def\pann@Cref@table{Table}
\def\pann@Cref@alg{Algorithm}
\def\pann@cref@sec{section}
\def\pann@cref@fig{figure}
\def\pann@cref@eq{equation}
\def\pann@cref@tab{table}
\def\pann@cref@table{table}
\def\pann@cref@alg{algorithm}
\def\pann@Cref#1:#2\@nil{\@nameuse{pann@Cref@#1}~\ref{#1:#2}}
\def\pann@cref#1:#2\@nil{\@nameuse{pann@cref@#1}~\ref{#1:#2}}
\renewcommand{\Cref}[1]{\expandafter\pann@Cref#1\@nil}
\renewcommand{\cref}[1]{\expandafter\pann@cref#1\@nil}
\makeatother

\newcommand{\methodname}{PANN}
\newcommand{\methodnames}{PANNs}

\newcommand{\probdim}{d}
\newcommand{\polydeg}{\ell}

\newcommand{\nnsize}{w}
\newcommand{\augsize}{m}

\newcommand{\nnsum}{\sum_{\nnind=1}^{\nnsize}}
\newcommand{\augsum}{\sum_{\augind=1}^{\augsize}}

\newcommand{\nnind}{j}
\newcommand{\augind}{k}

\newcommand{\nnbasiscoeff}{a_\nnind}

\newcommand{\augbasiscoeff}{b_\augind}

\newcommand{\onennbasiscoeff}{a}
\newcommand{\oneaugbasiscoeff}{b}

\newcommand{\nnbasisfunc}{ \psi_\nnind{\left(x; \theta^h \right)}}
\newcommand{\augbasisfunc}{\phi_\augind{\left(x \right)}}

\newcommand{\onennbasisfunc}{\psi}
\newcommand{\oneaugbasisfunc}{\phi}

\newcommand{\nnbasis}{  \nnsum \nnbasiscoeff  \nnbasisfunc}
\newcommand{\augbasis}{ \augsum  \augbasiscoeff \augbasisfunc}

\newcommand{\neuralnet}{\mathcal{N}}
\newcommand{\augnet}{\mathcal{P}}

\newcommand{\polynnpred}{u_\theta}

\newcommand{\x}{{ x}}

\newcommand{\argmin}{\operatornamewithlimits{argmin}}

\newcommand{\Fcal}{{\mathcal{F}}}

\newcommand{\rupdates}[1]{\textcolor{black}{#1}}

\title{Polynomial-Augmented Neural Networks (PANNs) with Weak Orthogonality Constraints for Enhanced Function and PDE Approximation
\thanks{
\funding{The first and second authors were supported by Army Research Office (ARO) grant W911NF-22-2-0158. The third and fourth authors were supported by Air Force Office of Scientific Research (AFOSR) LRIR grant FA9550-25-1-0042.}
}}

\author{Madison Cooley\footnotemark[1]\thanks{University of Utah (\email{mcooley220@gmail.com}, \email{zhe@cs.utah.edu}, \email{kirby@cs.utah.edu}, \email{shankar@cs.utah.edu}).}
\and Shandian Zhe\footnotemark[1]
\and Robert M. Kirby\footnotemark[1]
\and Varun Shankar\footnotemark[1]}

\headers{}{Madison Cooley, Shandian Zhe, Robert M. Kirby, and Varun Shankar}

\ifpdf
\hypersetup{ pdftitle={Polynomial-Augmented Neural Networks (PANNs)} }
\fi
\begin{document}
\maketitle

%% ------------------------------------------------------------------
%% ABSTRACT
%% ------------------------------------------------------------------
\begin{abstract}
We present polynomial-augmented neural networks (PANNs), a novel machine learning architecture that combines deep neural networks (DNNs) with polynomial expansions. PANNs combine the strengths of DNNs (flexibility and efficiency in higher-dimensional approximation) with those of polynomial approximation (rapid convergence rates for smooth functions). To aid in both stable training and enhanced accuracy over a variety of problems, we present (1) a family of orthogonality constraints that impose mutual orthogonality between the polynomial and the DNN within a PANN; (2) a simple basis pruning approach to combat the curse of dimensionality introduced by the polynomial component; and (3) an adaptation of a polynomial preconditioning strategy to both DNNs and polynomials. We test the resulting architecture for its polynomial reproduction properties, ability to approximate both smooth functions and functions of limited smoothness, and as a method for the solution of partial differential equations (PDEs). Through these experiments, we demonstrate that PANNs offer superior approximation properties to DNNs for both regression and the numerical solution of PDEs, while also offering enhanced accuracy over both polynomial and DNN-based regression (each) when regressing functions with limited smoothness.
\end{abstract}



%% ------------------------------------------------------------------
%% END HEADER
%% ------------------------------------------------------------------

\section{Introduction}\label{sec:intro}
Recent advancements in machine learning, particularly within deep neural networks (DNNs), have significantly impacted various scientific fields due to their broad applicability and flexibility~\cite{jumper2021highly, lam2023learning, zhang2018deep}.
DNNs are popular primarily due to their expressiveness, scalability, and efficient optimization with gradient descent methods through the use of automatic differentiation. 
DNNs are versatile tools capable of solving diverse problems ranging from classification and regression to the solution of partial differential equations (PDEs) and image recognition.
Recently, DNNs have also been applied to both forward and inverse PDEs in the form of physics-informed neural networks (PINNs), which extend the capabilities of standard DNNs by incorporating a physics-based loss term into the data loss~\cite{raissi2018jmlr, raissi2019jcp}.
DNNs are also generalizable on diverse data and domain types without requiring {\em a priori} knowledge of solution characteristics~\cite{Liu2023D2IFLN,Meng2020Mutual}.
This property is especially beneficial in the context of PINNs, as they eliminate the need for mesh generation mandated by many traditional numerical methods for PDEs.
Furthermore, DNNs arguably break the curse of    dimensionality~\cite{beck2021solving,Cheridito2019Efficient,Hu2023Tackling,Jentzen2018A}, meaning that as the problem dimension increases, the required network size for accurate approximations does not grow exponentially with dimension.
These traits allow DNNs to approximate complicated functions effectively~\cite{demanet2007acha}. 

Despite the many advantages of DNNs, their use comes with significant challenges in model initialization and training~\cite{Ioffe2015Batch,Larochelle2009Exploring}.
For instance, DNNs encounter issues like vanishing or exploding gradients during training, where the back-propagated gradients either approach zero or increase exponentially, respectively; this could either result in prolonged training or high generalization errors~\cite{Haber2017Stable, Pascanu2012Understanding}.
Spectral bias is an additional challenge, manifesting as quick convergence to the low-frequency components of the target solution while struggling with high-frequency components~\cite{basri2020icml, rahaman2019icml, xu2019arxiv}---an issue that also extends to PINNs~\cite{wang2021cmame}.
Furthermore, DNNs are prone to overfitting the training data~\cite{Srivastava2014Dropout}, compromising their generalizability.

Traditional approximation methods, especially those involving polynomials, remain a strong choice for both function approximation and the solution of PDEs. 
However, while robust in many applications, polynomial least-squares methods also face many challenges. 
Polynomial least-squares typically require oversampling to achieve stability~\cite{adcock2020approximating,adcock2016generalized,Li2015Approximation,Migliorati2015Analysis} even on tensor-product grids. 
Polynomial approximation can also be generalized to non-tensor-product grids (and hence irregular domains), but this requires the use of sophisticated techniques such as on-the-fly basis function recomputation~\cite{Buhr2015Interactive,dolbeault2022optimal,Yan2017A} or localization~\cite{Chen2015Local,Jiang2016A,Shimoyama1996Localization}. 
In addition, naive polynomial approximation is subject to the curse of dimensionality, where the number of polynomial basis functions grows exponentially with dimension. 
Common techniques to combat this explosive growth of the number of basis functions include compressive sensing~\cite{adcock2021oracle,Li2016Compressed} (which induces sparsity in the polynomial coefficients), Smolyak/sparse grids (which utilize sparse tensor-product grids)~\cite{Bungartz2003Multivariate,Gerstner2004Numerical,Lauvergnat2014Quantum,Winschel2010Solving}, or hyperbolic cross approximation (which utilizes only a subset of the total degree polynomial basis)~\cite{Dung2015Hyperbolic,Dung2016Hyperbolic,Shen2010Sparse}. 
In general, (global) polynomial methods are well-suited to approximating smooth target functions, while DNNs often perform better approximating non-smooth functions~\cite{demanet2007acha,elbrachter2021deep}, at least partly due to their connections to piecewise polynomial approximation~\cite{opschoor2020aa,opschoor2023deep}. 

Motivated by these observations, we introduce Polynomial-Augmented Neural Networks (\methodnames), which combine the strengths of both DNNs and polynomials.
Specifically, we augment a standard DNN with a preconditioned polynomial layer containing trainable coefficients and mutually optimize the two approximations using a novel family of eight orthogonalization constraints that enforce weak orthogonality between the polynomial and DNN bases; we also precondition the DNN itself. While a naive addition of a polynomial layer can re-introduce the curse of dimensionality into this augmented approximation, we leverage basis truncation to control the number of polynomial basis terms for increasing dimension and high polynomial degrees. From the DNN perspective, the PANN architecture can be viewed as a residual block with a set of transformed skip connections containing trainable strength connection parameters. 

Our results show that the resulting PANN architecture significantly improves DNN approximations of polynomial target functions (unsurprisingly). More specifically, we present empirical results showing that the PANN architecture is superior to either DNNs or polynomials on tasks such as approximating functions with finite smoothness, high-dimensional function approximation, and approximating noisy functions drawn from a high-dimensional housing dataset. Further, when PANNs are used as physics-informed networks to solve PDEs, we observe relative $\ell_2$ errors that are \textbf{orders of magnitude} lower than traditional PINNs. We also show that the choice of orthogonality constraint can affect approximation quality and wall-clock training times in an application-dependent fashion.

Other works have focused entirely on polynomial neural networks (PNNs) and their variants~\cite{yang2022polynomial,chrysos2023regularization,cheng2024multilinear,fronk2023interpretable}. 
PNNs typically refer to neural networks composed of polynomials~\cite{yang2022polynomial}. 
For example, PNNs have been successfully used for generative modeling tasks~\cite{chrysos2019polygan, chrysos2021conditional, wu2022adversarial}. 
In another instance of a polynomial-based approach, $\Pi$-Nets~\cite{chrysos2022ieee,chrysos2020ieee,chrysos2023regularization} modify convolutional neural nets to output polynomials of the input variables via high-order tensors. 
Similarly, multilinear operator networks (MONets) \cite{cheng2024multilinear} and interpretable polynomial neural ODEs \cite{fronk2023interpretable} integrate polynomials into the core components of the networks.
In contrast to the methods above, our approach does not rely solely on polynomials. 
Instead, we augment traditional DNNs with a polynomial layer, outputting a linear combination of DNN and polynomial components. 
The PANN architectural design leverages the strengths of DNNs while enhancing expressivity through the augmented polynomials.

Many other works directly replace DNNs with polynomials~\cite{chrysos2022eccv,goyal2020ieee,kileel2019expressive,li2019powernet}. 
Modern radial basis function-finite difference (RBF-FD) methods combine RBFs and polynomials with orthogonality constraints that enforce polynomial reproduction. 
There, since closed-form expressions are available for the RBF basis, no training is required, and the orthogonality constraint is enforced by treating the polynomial coefficients as Lagrange multipliers~\cite{BarnettPHS,bayona2019role,fasshauer2007meshfree,overlappedrbffd_2017,movingdomain_2021}.
RBF-FD is primarily used to generate finite difference weights on scattered points. 
However, barring the freedom to handle irregular point sets and domains, it suffers from many of the same issues as traditional polynomial approximation, such as sensitivity to noise, the curse of dimensionality, and difficulties tackling data sampled from finitely-smooth functions. 
In contrast, PANNs are global approximators that must be trained; the presence of a DNN introduces an entire family of orthogonality constraints, each presenting different cost-accuracy tradeoffs on different problems, but the overall method is robust to noise and inherits the benefits of DNNs.

The remainder of this paper is structured as follows:~\Cref{sec:background} provides essential background and notation. 
The new PANN architecture, its training, preconditioning, and the new orthogonality constraints are described in~\Cref{sec:methods}. 
Then Sections~\ref{sec:experiments} and~\ref{sec:pde-experiments} present our numerical results on regression and PDE problems, respectively, which include an assessment of computational cost and accuracy compared to baseline methods. 
Finally,~\Cref{sec:conclusion} discusses the results and outlines future research directions.

\section{Background} \label{sec:background}
In this section, we define the general optimization problems we are interested in, along with a brief review of DNNs and certain classes of polynomial approximation methods. 
Table~\ref{tab:notation} lists notation used throughout.
The problem dimension is denoted by $\probdim$, and $\polydeg$ signifies the polynomial degree used to generate the polynomial bases. 
The total number of training points is represented by $N_{data}$, and $\nnsize$ refers to the width of the last layer of a DNN. 
The total number of polynomial bases, which is also the width of the polynomial layer, is denoted by $\augsize$. 
The DNN basis coefficients are symbolized by $\nnbasiscoeff$ for $j=1, ..., \nnsize$, while the polynomial layer basis coefficients are represented by $\augbasiscoeff$ for $k=1, ..., \augsize$. 
The DNN basis functions are indicated by $\onennbasisfunc_\nnind$ for $j=1, ..., \nnsize$, and the polynomial layer basis functions are represented by $\oneaugbasisfunc_\augind$ for $k=1, ..., \augsize$.
In this paper, we primarily focus on the supervised regression problem with the form, 
\begin{align}
    \argmin_{\theta} \sum_{i=1}^{N_{data}} | u_{\theta}(x_i) - u(x_i) |^2, \label{eq:prob-stmt}
\end{align}
for $N_{data}$ training points $x \in \mathbb{R}^d$ such that $u$ is the true solution we aim to approximate and $u_{\theta}$ is the model parameterized by its weights and biases $\theta$. 
In the context of the solution of PDEs, we also leverage semi-supervised approaches of the form,
\begin{align}
    \argmin_{\theta} \sum_{i=1}^{N_{data}} | u_{\theta}(x_i^b) - u(x_i^b) |^2 
    + \lambda \sum_{j=1}^{N_{PDE}} | \Fcal[u_\theta](x_j^r) - f(x_j^r) |^2, \label{eq:pde-prob-stmt}
\end{align}
where $\Fcal$ is some (linear or non-linear) differential operator operating on $N_{PDE}$ collocation points $\{\x^r_i\}_{i=1}^{N_{PDE}}$ from the domain $\Omega$ and $N_{data}$ points 
$\{\x^b_i\}_{i=1}^{N_{data}}$ 
from the boundary of the domain ($\partial \Omega$). 
$\lambda$ is a Lagrange multiplier that balances the learning between the data and residual loss terms. 
Subsequent sections detail extensions of Equations~\ref{eq:prob-stmt} and~\ref{eq:pde-prob-stmt}, which incorporate custom orthogonalizing regularization terms, along with polynomial preconditioning.

\subsection{Deep Neural Networks}
Following~\cite{cyr2020robust}, we represent the family of DNNs, 
$\neuralnet_{a,\theta^h} \in \mathbb{R}^d \rightarrow \mathbb{R}$ of width $w$, as a linear combination of adaptive basis functions given by
\begin{align}
    \neuralnet(\x; a, \theta^h) = \nnbasis, \label{eq:NN-output-lincomb}
\end{align} 
where each $a_j$ for $j=1,.., w$ and $\theta^h$ constitute the weights and biases in the last layer and hidden layers, respectively, forming the set of all network parameters $\theta$; for readability, we henceforth refer to this DNN by $\mathcal{N}(x)$.
Then, each $\psi_j$ are non-linear activation functions (such as ReLU or tanh) acting on the outputs of the hidden layers. The parameters $\theta$ are computed through some iterative optimization technique; in this work, we use variants of gradient descent methods such as ADAM~\cite{kingma2015iclr} and L-BFGS~\cite{liu1989MP}.

\begin{figure}[!htpb]
\begin{minipage}{0.4\textwidth}
\begin{tikzpicture} %%%%%%%%%%%%%%%%%%%%%%%%%%%%
\begin{groupplot}[
    group style={
        group size=1 by 1,
        horizontal sep=0.01cm
    },
    legend style={at={(0.05,0.5)}, anchor=west},
    ylabel style={yshift=-0.2cm} ,
    width=0.9\textwidth, 
    height=1.2\textwidth, 
]

\nextgroupplot[
    xlabel={\small Input dimension $d$},
    ylabel={\small Number of basis functions},
    legend pos=north west,
    ymajorgrids=true,
    xmajorgrids=true,
    xtick=data,
    ymode=log,
    xticklabels={1,2,3,4,5},
    legend entries={TP-8, TD-8, HC-8}]
    
% Data for tensorproduct with max total degree=8
\addplot[color=blue, mark=diamond] coordinates {
    (1, 9)
    (2, 81)
    (3, 792)
    (4, 6561)
    (5, 59049)
};

% Data for total-degree with max total degree=8
\addplot[color=red, mark=diamond] coordinates {
    (1, 9)
    (2, 45)
    (3, 165)
    (4, 495)
    (5, 1287)
};

% Data for hyper-cross with max total degree=8
\addplot[color=green, mark=diamond] coordinates {
    (1, 9)
    (2, 23)
    (3, 44)
    (4, 73)
    (5, 111)
};

\end{groupplot}  
\end{tikzpicture}
\end{minipage}
\begin{minipage}{0.55\textwidth}
\hspace{-1.1cm}
\begin{minipage}{\textwidth}
\begin{tikzpicture}
{\scriptsize
  \matrix (mat) [matrix of math nodes, left delimiter={[}, right delimiter={]}] {
    |[fill=blue!20]| p_0(x)p_0(y) & |[fill=blue!20]| p_0(x)p_1(y) & |[fill=blue!20]| p_0(x)p_2(y) \\ 
    |[fill=blue!20]| p_1(x)p_0(y) & |[fill=blue!20]| p_1(x)p_1(y) & |[fill=blue!20]| p_1(x)p_2(y) \\ 
    |[fill=blue!20]| p_2(x)p_0(y) & |[fill=blue!20]| p_2(x)p_1(y) & |[fill=blue!20]| p_2(x)p_2(y) \\ 
  }; 
  \node[above=-0.1cm of mat](title) {{\footnotesize Tensor product (TP) bases.}};
}
\end{tikzpicture}
\end{minipage}\\
\begin{minipage}{\textwidth}
\hspace{-1.1cm}
\begin{tikzpicture}
{\scriptsize
  \matrix (mat) [matrix of math nodes, left delimiter={[}, right delimiter={]}] {
    |[fill=red!20]| p_0(x)p_0(y) & |[fill=red!20]| p_0(x)p_1(y) & |[fill=red!20]| p_0(x)p_2(y) & |[fill=red!20]| p_0(x)p_3(y) & |[fill=red!20]| p_0(x)p_4(y) \\
    |[fill=red!20]| p_1(x)p_0(y) & |[fill=red!20]| p_1(x)p_1(y) & |[fill=red!20]| p_1(x)p_2(y) & |[fill=red!20]| p_1(x)p_3(y) & \\
    |[fill=red!20]| p_2(x)p_0(y) & |[fill=red!20]| p_2(x)p_1(y) & |[fill=red!20]| p_2(x)p_2(y) &  &  \\
    |[fill=red!20]| p_3(x)p_0(y) & |[fill=red!20]| p_3(x)p_1(y) &  &  &  \\
    |[fill=red!20]| p_4(x)p_0(y) &  &  &  & \\
  }; 
  \node[above=-0.1cm of mat](title) {{\footnotesize Total degree (TD) bases.}};
}
\end{tikzpicture}
\end{minipage}\\
\begin{minipage}{\textwidth}
\hspace{-1.1cm}
\begin{tikzpicture}
{\scriptsize
 \matrix (mat) [matrix of math nodes, left delimiter={[}, right delimiter={]}] {
   |[fill=green!20]| p_0(x)p_0(y) & |[fill=green!20]| p_0(x)p_1(y) & |[fill=green!20]| p_0(x)p_2(y) & |[fill=green!20]| p_0(x)p_3(y) & |[fill=green!20]| p_0(x)p_4(y) \\
   |[fill=green!20]| p_1(x)p_0(y) & |[fill=green!20]| p_1(x)p_1(y) &  &  &  \\
   |[fill=green!20]| p_2(x)p_0(y) &  &  &  &  \\
   |[fill=green!20]| p_3(x)p_0(y) &  &  &  &  \\
   |[fill=green!20]| p_4(x)p_0(y) &  &  &  &  \\
 }; 
 \node[above=-0.1cm of mat](title) {{\footnotesize Hyperbolic cross (HC) bases.}};
}
\end{tikzpicture}
\end{minipage}
\end{minipage}
\vspace{-0.15cm}
\caption{\textit{\small Visual depiction of bases sets using different generation techniques on the right and the total number of basis function each method produces for increasing problem dimension.}} 
\label{fig:basisgen}
\vspace{-0.25cm}
\end{figure}
\subsection{Polynomial Methods}
We define a family of polynomial models, $\augnet : \mathbb{R}^d \rightarrow \mathbb{R}$, in a similar form to~\Cref{eq:NN-output-lincomb}, as a linear combination of $m$ orthogonal polynomials $\phi_k$ and parameters $b_k$. 
Specifically, 
\begin{align}
    \augnet(x) = \augbasis, \label{eq:AN-output-lincomb}
\end{align}
where $\phi_k$ for $k=1, ..., m$ constitutes a polynomial basis in $d$ dimensions. 
Each multivariate polynomial basis $\phi_k : \mathbb{R}^d \to \mathbb{R}$ maps a point $x \in \mathbb{R}^d$ to a scalar through a product of univariate basis terms, such as  $\phi_{p_1, ..., p_d}(x) = \prod_{j=1}^d P_{p_j}(x_j)$, where, in this work, $P_{p_j}$ denotes the univariate Legendre polynomial of degree $p_j$.
For a given polynomial of degree $\ell$ in dimension $d$, there are many natural choices of basis functions including tensor-product, total-degree, and hyperbolic-cross bases. 
\Cref{fig:basisgen} illustrates each type of basis, comparing their cardinality to the function space spanned by each. 
The number of tensor-product basis functions grows exponentially with dimension, often rendering this choice impractical for computation in high-dimensions~\cite{Petras2003Smolyak}. In contrast, the use of total-degree basis functions enforces that total degree of each polynomial basis function never exceeds $\ell$, thereby ensuring a more gradual increase in cardinality (though still exponential). Finally, hyperbolic-cross bases exhibit the slowest growth relative to dimension, albeit at the expense of expressivity and approximation power.

In this work, we start with the total-degree setting and leverage data-dependent basis pruning during training to ensure parsimony and computational efficiency. We restrict ourselves to Legendre polynomials as the basis for the polynomial expansion as they are a natural choice for the $\ell_2$ loss functions typically used in ML, but our approaches also carry over straightforwardly to other polynomial bases. More importantly, depending on the experiment at hand, we employ polynomials in purely spatial coordinates, e.g., in spacetime (2D in space and 1D in time, e.g., $d=3$) and high-dimensional settings as well. For example, we  present results on a synthetic regression problem in six ($d=6$) dimensions and a real-world noisy dataset in eight ($d=8$) dimensions.
  
\section{Polynomial Augmented Neural Networks (\methodnames)} \label{sec:methods}
This section outlines our enhanced neural network architecture incorporating polynomials, expanding on the preconditioning methods in ~\Cref{sec:precond}, and our unique discrete orthogonality constraints in~\Cref{sec:orth}. 
The algorithmic framework, including the selection and truncation of polynomial bases, is detailed in~\Cref{sec:algorithm}. 

\begin{figure}[ht]
\begin{center}
\begin{minipage}{0.50\textwidth}
\begin{tikzpicture}[scale=0.43, transform shape, font=\large,
  neuron/.style={font=\large, circle, draw, fill=blue!20, minimum size=0.7cm},
  polyneuron/.style={font=\large, circle, draw, fill=red!20, minimum size=0.7cm},
  outputneuron/.style={circle, draw, fill=purple!40, minimum size=1.0cm},
  nnoutputneuron/.style={circle, draw, fill=blue!20, minimum size=1.0cm},
  augnoutputneuron/.style={circle, draw, fill=red!20, minimum size=1.0cm},
  inputneuron/.style={circle, draw, fill=gray!20, minimum size=0.75cm},
  input/.style={coordinate},
  arrow/.style={->, >=stealth, thin},
  bordered/.style={
        draw=white,
        line width=2pt,
        thick,
        inner sep=1pt,
        outer sep=0pt, % Ensure the label position remains consistent
        text=black
    }
]

%%%%%%%%%%%%%%%%%%%%%%%%%%%%%%%%%%%%%% NN
% Input layer
\node[inputneuron] (I1) at (0.5, -2.5) {};

% Hidden-1 layer
\foreach \i in {1,2,4}
  \node[neuron] (H1\i) at (2.0, 2.5-\i) {};
\node (H13) at (2.0, 2.5-2.75) {};
  
% Hidden-2 layer
\foreach \i in {1,2,4}
  \node[neuron] (H2\i) at (4, 2.5-\i) {};
\node (H23) at (4, 2.5-2.75) {};
  
% Hidden-3 layer
\foreach \i in {1,2,4}
  \node[neuron] (H3\i) at (6.0, 2.5-\i) {};
\node (H33) at (6.0, 2.5-2.75) {};
  
% Hidden-4 layer
\foreach \i in {1,2,4}
  \node[neuron] (H4\i) at (8, 2.5-\i) {};
\node (H43) at (8, 2.5-2.75) {};

% Output layer
\node[nnoutputneuron] (O1) at (11.5, -0.75) {};

%%%%%%%%%%%%%%%%%%%% NN Connections
% Connect Input layer to Hidden1 layer
\foreach \i in {1}
  \foreach \j in {1,2,4}
    \draw[arrow] (I\i) -- (H1\j);

% Connect Hidden1 layer to Hidden2 layer
\foreach \i in {1,2,4}
  \foreach \j in {1,2,4}
    \draw[arrow] (H1\i) -- (H2\j);
        
% Connect Hidden3 layer to Hidden4 layer
\foreach \i in {1,2,4}
  \foreach \j in {1,2,4}
    \draw[arrow] (H3\i) -- (H4\j);

% Connect Hidden4 layer to Output layer
\draw[arrow] (H41) -- (O1) node[above, midway] {$a_0$};
\draw[arrow] (H42) -- (O1) node[above, midway] {$a_1$};
\draw[arrow] (H44) -- (O1) node[above, midway] {$a_w$};
    
%%%%%%%%%%%%%%%%%%%%%%%%%%%%%%%%%%%%%%
% poly layer
\foreach \i in {1,2,4}
  \node[polyneuron] (P1\i) at (8.0, -1.8-\i) {};
\node (P13) at (8.0, -1.8-2.75) {};
  
% Poly-Output layer
\node[augnoutputneuron] (OP1) at (11.5, -3.5) {};
  
% Connect input layer to poly-layer
\foreach \i in {1}
  \foreach \j in {1,2,4}
    \draw[arrow] (I\i) -- (P1\j);
    
% Connect poly-layer to output
\draw[arrow] (P11) -- (OP1) node[above, midway, xshift=-0.1em] {$b_0$};
\draw[arrow] (P12) -- (OP1) node[above, midway, xshift=-0.1em] {$b_1$};
\draw[arrow] (P14) -- (OP1) node[above, midway, xshift=-0.1em] {$b_m$};
% \draw[arrow] (P15) -- (OP1) node[above, midway, xshift=-0.1em] {$b_m$};
%%%%%%%%%%%%%%%%%%%%%%%%%%%%%%%%%%%%%%

% Connect bot Output layers
% \node[outputneuron] (OFull1) at (13.0, -1.25-1.0) {};
\node[circle, draw, minimum size=0.55cm] (plus) at (13.0, -2.25) {$+$};
\node (OFull1) at (14.5, -2.25) {$u_\theta(x)$};
  
%% Create a node for the solid box behind the top graph
%\node[draw, fill=gray!10, inner sep=10pt, fit=(I1)(H41)(OP1)] (box) {};

% \draw[arrow] (O1) -- (OFull1);
% \draw[arrow] (OP1) -- (OFull1);

\draw[arrow] (O1) -- (plus);
\draw[arrow] (OP1) -- (plus);
\draw[arrow]  (plus) -- (OFull1);

% Add labels
\node at (I1) {$x$};

\node at (H13) {$\vdots$};
\node at (H23) {$\vdots$};
\node at (H33) {$\vdots$};

\node (H13mid) at (5, 2.5-1.00) {$\dots$};
\node (H33mid) at (5, 2.5-2.75) {$\dots$};
\node (H43mid) at (5, 2.5-4.00) {$\dots$};

\node at (H41) {$\psi_1$};
\node at (H42) {$\psi_2$};
\node at (H43) {$\vdots$};
\node at (H44) {$\psi_w$};

\node at (P11) {$\phi_1$};
\node at (P12) {$\phi_2$};
\node at (P13) {$\vdots$};
\node at (P14) {$\phi_m$};
% \node at (P15) {$\phi_m$};

\node at (O1)     {\small $\neuralnet(x)$};
\node at (OP1)    {\small $\augnet(x)$};
% \node at (OFull1) {\small $u_\theta(x)$};

% Draw the bracket and label above hidden layers
% \draw[decorate,decoration={brace, amplitude=10pt,raise=6pt}] (H11.north west) -- (H41.north east) node[midway,above=12pt] {\footnotesize Hidden Layers with Depth $L$};
% \draw[decorate,decoration={brace, amplitude=2pt, raise=10pt}] (I1.south) -- (I1.north) node[midway, left=20pt, rotate=90, anchor=center] {\footnotesize Input Dimension $d$};

% \draw[decorate,decoration={brace, amplitude=6pt, raise=10pt, mirror}] (H11.north) -- (H14.south) node[midway, left=22pt, rotate=90, anchor=center, fill=white] {\footnotesize Network Width $W$};
% \draw[decorate,decoration={brace, amplitude=6pt, raise=10pt, mirror}] (P11.north) -- (P15.south) node[midway, left=22pt, rotate=90, anchor=center, fill=white] {\footnotesize Basis Functions $M$};

\end{tikzpicture}
\end{minipage} 
\begin{minipage}{0.49\textwidth}
\begin{tikzpicture}[scale=0.43, transform shape,
  nnweightlayer/.style={draw, rectangle, thin, fill=blue!20, minimum width=0.65cm, minimum height=3.5cm, rounded corners},
  pnweightlayer/.style={draw, rectangle, thin, fill=red!20, minimum width=0.65cm, minimum height=4.25cm, rounded corners},
  neuron/.style={circle, draw, fill=blue!20, minimum size=0.65cm},
  polyneuron/.style={circle, draw, fill=red!20, minimum size=0.65cm},
  outputneuron/.style={circle, draw, fill=purple!40, minimum size=1.0cm},
  nnoutputneuron/.style={circle, draw, fill=blue!20, minimum size=1.0cm},
  augnoutputneuron/.style={circle, draw, fill=red!20, minimum size=1.0cm},
  inputneuron/.style={circle, draw, fill=gray!20, minimum size=0.70cm},
  input/.style={coordinate},
  arrow/.style={->, >=stealth, thin},
  bordered/.style={
        draw=white,
        line width=2pt,
        thick,
        inner sep=1pt,
        outer sep=0pt, % Ensure the label position remains consistent
        text=black
    }
]

%%%%%%%%%%%%%%%%%%%%%%%%%%%%%%%%%%%%%% NN
% Input layer
\node (I1) at (0.5, 1.5) {$x$};

% Hidden-1 layer
\node[nnweightlayer] (H1) at (2.0, 1.5) {$H_1$};
\node (A1) at (2.9, 1.7) {$\sigma$};

% Hidden-2 layer
\node[nnweightlayer] (H2) at (4.0, 1.5) {$H_2$};
  
% Hidden-3 layer
\node (H3) at (5.5, 1.5) {$\mathbf{\dots}$\;\;\;\;\;};
  
% Hidden-4 layer
\node[nnweightlayer] (H4) at (8.0, 1.5) {$H_L$};
\node (A2) at (6.9, 1.7) {$\sigma$};

% Poly layer
\node[pnweightlayer] (P1) at (5.5, -2.75) {$P$};

%%%%%%%%%%%%%%%%%%%% NN Connections
% Connect Input layer to Hidden1 layer
\draw[-latex, arrow] (I1) -- (H1);

% Connect Hidden1 layer to Hidden2 layer
\draw[arrow] (H1) -- (H2);
        
% Connect Hidden3 layer to Hidden4 layer
\draw[arrow] (H3) -- (H4);

\node[circle, draw, minimum size=0.55cm] (plus) at (10.0, 1.5) {$+$};
\node (OFull1) at (12.0, 1.5) {$u_\theta(x)$};

\draw[arrow] (I1.south) to[bend right=40] (P1);
% \draw[arrow] (P11) -- (OP1) node[above, midway, xshift=-0.1em] {$b_0$};
\draw[arrow] (P1) to[bend right=20] node[midway, above] {$c_1$} (plus);
% \draw[arrow] (P1) to[bend right=35] node[midway, above] {$...$} (plus);
\draw[arrow] (P1) to[bend right=55] node[midway, above] {$c_m$} (plus);
\draw[arrow]  (H4) -- (plus);
\draw[arrow]  (plus) -- (OFull1);

\node (CM) at (8.75, -1.2) {$...$};

% \draw[node] (c1) at (8.0, 1.0) {$c_1$};

\end{tikzpicture}
\end{minipage}
\end{center}
\caption{
\textit{\small 
Both figures show the proposed neural network architecture with polynomial layer (PANN). 
The left figure demonstrates the architecture from the adaptive basis viewpoint where each $\psi_i$ and $a_i$ for $i=1,...w$ are the DNN bases and coefficients, while $\phi_j$ and $b_j$ for $j=1,...m$ are the polynomial layer bases and coefficients respectively. $u_\theta$ is the model output which is a linear combination of the DNN and polynomial layer bases and coefficients.
Alternatively, the right figure demonstrates the architecture as a residual block with transformed skip connections such that each $H_k$ for $k=1,..,L$ represent the hidden layers of the DNN, $\sigma$ are non-linear activations, $P$ is the polynomial layer and $c_j$ for $j=1,..,m$ are the transformed and adaptive skip connections. 
Unlike traditional residual blocks aimed at resolving the vanishing gradient issue, our residual interpretation focuses on augmenting the DNN’s output with additional polynomial-based transformations, enriching the function approximation.}
}
\label{fig:architecture}
\end{figure}

The \methodname~architecture involves augmenting a standard DNN with a trainable polynomial expansion; see~\Cref{fig:architecture} (left) for an illustration. 
We define the model's prediction, $\polynnpred$, as the sum of the DNN's output, $\neuralnet(\x)$ (defined in~\Cref{eq:NN-output-lincomb}), and the output of the polynomial layer, $\augnet(\x)$  (defined in~\Cref{eq:AN-output-lincomb}) as follows:
\begin{align}
    \polynnpred(\x) &= \neuralnet(\x) + \augnet(\x) \;=\; \nnbasis + \augbasis, \label{eq:polynn}
\end{align}

In this paper, we primarily present \methodnames~ from the vantage point of an adaptive basis method~\cite{cyr2020robust}. 
However, one can also interpret \methodnames~as a type of non-traditional residual network where the DNN output is combined with a set of polynomial-transformed skip connections that have trainable strength parameters; see ~\Cref{fig:architecture} (right). In contrast to traditional interpretations of residual networks such as ResNet, where skip connections are primarily used to prevent vanishing gradients and enhance gradient flow, our polynomial-enhanced residual block simply serves to improve the model’s function approximation capabilities by enriching the set of functions being used for approximation. 
Such residual connections have been used recently within operator learning architectures such as the Fourier neural operator (FNO)~\cite{li2020fourier}.  
We note that in this work, we use a single polynomial layer rather than a ``deep'' polynomial architecture for two primary reasons. 
First, for the types of PDE-based problems considered in this work, a single polynomial layer provides sufficient expressivity while maintaining computational efficiency. 
The static nature of the polynomial bases allows us to precompute and store them, significantly reducing the computational overhead during training (as detailed in~\Cref{sec:algorithm}). 
Second, while deeper polynomial networks can offer advantages in high-dimensional or highly nonlinear settings, previous findings have shown that additional depth in the polynomial layers does not necessarily improve the performance~\cite{leshno1993multilayer} in settings like ours, where the DNN component already provides a flexible and nonlinear representation. 
Third, we explored alternative architectures where the polynomial or the DNN were fit to the target function, and the other component fit to the residual; we found this alternative underperformed our approach outlined in this paper.
Our experiments demonstrate that our jointly-trained ``single polynomial layer'' approach allows us to balance representation power with efficiency.

\begin{figure}[!ht]
\centering
\includegraphics[width=0.4\textwidth]{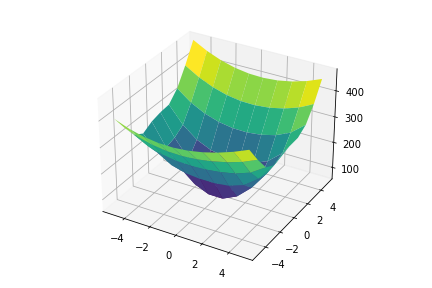}
\includegraphics[width=0.4\textwidth]{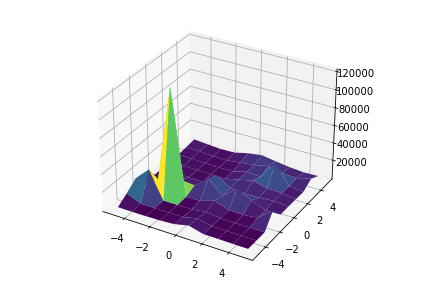}
\caption{\textit{\small
Loss landscapes of a physics-informed PANN (left) and standard PINN (right) on a 2D Poisson problem.
}}
\label{fig:poisson-losslandscapes}
\end{figure}

\paragraph{A loss landscape perspective} As a brief aside, we present an experiment to provide further intuition on the \methodname~architecture. 
In ~\Cref{fig:poisson-losslandscapes} we show a comparison between the loss landscapes of a PANN used as a physics-informed neural network (PINN) and a standard PINN.
For this experiment, we perturbed the two dominant eigenvectors ($\bf \delta$, $\bf \nu$) of the loss Hessian, 
and evaluated the adjusted loss $\mathcal{L'}$ across a specified range for $\alpha$ and $\beta$ such that, 
$\mathcal{L'}(\alpha, \beta)=\mathcal{L}( \theta + \alpha \delta + \beta \nu )$ 
and $\alpha, \beta \in [-\alpha_0, \alpha_0 ] \times [ -\beta_0, \beta_0 ]$~\cite{krishnapriyan2021neurips, liu2022loss}.
~\Cref{fig:poisson-losslandscapes} demonstrates that polynomial augmentation smooths the loss landscape, suggesting that it helps avoid local minima, hence simplifying the optimization process and possibly boosting model accuracy. 
This qualitative observation is consistent with the training behavior shown in~\Cref{fig:poisson-equalcost}. 
Specifically, when trained under comparable runtime budgets, PANNs reduce the PDE loss, boundary loss, and relative $\ell_2$ error more rapidly than the standard PINNs of varying depth. 
Taken together with the additional quantitative results in~\Cref{sec:experiments}, these findings suggest that polynomial augmentation contributes to a more favorable optimization landscape. 

\paragraph{Matrix Notation}
For clarity in this and later sections, we introduce the matrix notation for the basis evaluations associated with the DNN and polynomial components. 
Specifically, given the set of $N$ training or collocation points $\{x_i\}_{i=1}^N$, we define the $N \times m$ matrix $\Phi$ and the $N \times w$ matrix $\Psi$
such that, 
\[
\Psi_{i,j} = \psi_j(x_i), \quad \Phi_{i,k} = \phi_k(x_i),
\]
where $\psi_j(\cdot)$ and $\phi_k(\cdot)$ denote the $j$th DNN basis function (activation of the last hidden layer) and the $k$th polynomial basis function respectively.  
The matrix form equivalent of~\Cref{eq:polynn} can therefore be written compactly as
\begin{align}
    u_\theta = \Psi a + \Phi b,
\end{align}
where $a$ and $b$ are the corresponding coefficient vectors. 

\subsection{Preconditioning} \label{sec:precond}
The convergence rates of many numerical methods rely on problem conditioning, where ill-conditioned problems typically exhibit slower convergence. 
Problem conditioning not only influences traditional methods but is also a critical factor in the training difficulties of DNNs, including PINNs~\cite{krishnapriyan2021neurips}.
In this work, we apply the polynomial preconditioning techniques from~\cite{jakeman2017generalized, narayan2017christoffel}. 
This preconditioning technique was developed for least-squares approximation scenarios with extensive undersampling, \emph{i.e.}, with far fewer function samples ($N$) than polynomial basis functions ($m$), where it is vital to maximize sparsity in the under-determined polynomial coefficients. 
The optimization problem considered in~\cite{jakeman2017generalized} aims to solve the inequality-constrained $\ell_1$-minimization problem defined as: 
\begin{align}
\argmin_{b} ||b||_1 \;\;\;\text{such that} \;\; || K (\Phi b  - f) ||_2 \leq \epsilon. \label{eq:poly-precond}
\end{align}
Here, the diagonal preconditioning matrix $K \in \mathbb{R}^{N \times N}$ is constructed to improve the $\ell_1$-minimization problem's tractability and aids in recovering sparse solutions. 
Formally, 
\begin{equation}
K_{N,N} = \sqrt{ \frac{m}{\sum_{i \in \Lambda} \phi_i^2(x)}}, 
\label{eq:poly-precond-const}
\end{equation}
where $\Lambda$ is the set of all polynomial bases. In the context of polynomial least-squares approximation, this preconditioning attempts to rescale polynomial bases with large norms, leading to a system where each basis contributes more equally. 

In this work, we consider a more general form of~\Cref{eq:poly-precond-const} such that the objective function constrains both the preconditioned error norm and parameter norm. 
Furthermore, we apply this preconditioning to both the DNN ($\Psi$) and the polynomial bases ($\Phi$). 
For DNN bases coefficients $a$, and polynomial bases coefficients $b$, the
updated optimization problem considered is therefore,
\begin{align}
\min_{a, b, \theta} \;\; || K ( \Psi a + \Phi b - f ) ||^2_2 + \lambda_r||\theta||_1, \label{eq:poly-precond-ours}
\end{align}
where $\lambda_r$ is a prescribed Lagrange multiplier. 
For PDE-based problems, the preconditioning matrix $K$ is applied to both the boundary and residual terms, ensuring that both the data and physics losses are normalized under the same preconditioning.

\subsection{Discrete orthogonality constraints} \label{sec:orth}
In this work, we utilize a novel family of orthogonality constraints designed to induce a ``weak"  orthogonality between the DNN basis and the polynomial layer within the \methodname. 
This orthogonality was inspired by a philosophically similar approach utilized in modern radial basis function-finite difference (RBF-FD) methods~\cite{BarnettPHS,FlyerPHS}, where RBF expansions are computed in such a way that they are orthogonal to some polynomial bases (typically total-degree); this orthogonality constraint endows RBF-FD weights with polynomial reproduction properties, thereby controlling their convergence rates. While deriving this constraint in RBF-FD methods is straightforward, it is significantly more challenging in the context of \methodnames, which require training to determine not only polynomial coefficients but DNNs coefficients \emph{and} bases. 

Fortunately, enforcing orthogonality is nevertheless possible through several means.  
For instance, a \emph{continuous} orthogonality constraint between the DNN output $\neuralnet(\x)$ and the polynomial layer $\augnet(\x)$ can be expressed as $\int_{\Omega} \neuralnet(\x) \augnet(\x) = 0$. 
While it is generally impossible to compute this integral analytically, it can be approximated to arbitrary accuracy by a suitable quadrature formula provided the DNN is continuous~\cite{davis2007methods}. 
However, this approach does not generalize straightforwardly to irregular domains or higher dimensional problems without sacrificing the meshless nature of DNNs. 

Our approach involves replacing the continuous orthogonality constraint with discrete alternatives that obviate the need for quadrature. 
To see how, consider more carefully the continuous constraint \\ $\int_{\Omega} \neuralnet(\x) \augnet(\x)=0$.
This constraint yields many equivalent forms, some of which include, 
\begin{align} 
    \int\limits_{\Omega}\nnbasis \augnet(\x) = 0 \Longleftrightarrow  \int\limits_{\Omega}\augbasis\neuralnet(\x; \theta) = 0.
\end{align}
One straightforward way for these integrals to be zero is to enforce that the integrands themselves be zero. This can be enforced by forcing the summands to be zero, which in turn can be done by enforcing that each term in the summands be zero. This chain of reasoning leads to the following two distinct families of discrete constraints:
\begin{align}
    a_j \psi_j (\x) \augnet(\x) &= 0, \;\;\; j=1,\ldots, w, \label{eq:NNconstraint} \\
    b_j \phi_k (\x) \neuralnet(\x)&=0, \;\;\; k=1,\ldots,m. \label{eq:polyconstraint}
\end{align}
It is important to note that the discrete ``index-wise'' constraints in Equations~\ref{eq:NNconstraint} and~\ref{eq:polyconstraint} imply that the continuous constraint holds, but the converse does not necessarily hold. 
\begin{table}[!htpb]  
% \vspace{-0.35cm}
\footnotesize
\caption{\textit{A family of orthogonality constraints}} \label{table:orthoconstraints}
\begin{center}
\begin{tabular}{>{\raggedright\arraybackslash}p{2.0cm}
                >{\raggedright\arraybackslash}p{2.3cm}
                >{\raggedright\arraybackslash}p{6.0cm}}
\toprule
\textbf{Method} & \textbf{Objective} & \\
\midrule
$C_A$ & $\neuralnet(x) \augnet(x)$ & \\
$C_B$ & $\neuralnet(x) \oneaugbasiscoeff_k \augbasisfunc$ & for all $k=1, ..., m$ \\
$C_C$ & $\neuralnet(x) \augbasisfunc$ & for all $k=1, ..., m$ \\
$C_D$ & $\neuralnet(x) \oneaugbasiscoeff_k$ & for all $k=1, ..., m$ \\
$C_E$ & $\augnet(x) \onennbasiscoeff_j \onennbasisfunc_j(x)$ & for all $j=1, ..., w$ \\
$C_F$ & $\augnet(x) \onennbasisfunc_j(x)$ & for all $j=1, ..., w$ \\
$C_G$ & $\augnet(x) \onennbasiscoeff_j$ & for all $j=1, ..., w$ \\
$C_H$ & $\onennbasiscoeff_j \onennbasisfunc_j(x) \oneaugbasiscoeff_k \augbasisfunc$ & for all $k=1, ..., m$ and $j=1, ..., w$ \\
\bottomrule
\end{tabular}
\end{center}
\end{table}

Expanding either \eqref{eq:NNconstraint} or \eqref{eq:polyconstraint} leads to a family of orthogonality constraints; these are shown in ~\Cref{table:orthoconstraints}. 
Intuitively, constraint $C_A$ is the ``weakest'' of our constraints, only enforcing that the products of the DNN and polynomial in the PANN be zero. 
The other constraints are stronger. For instance, constraints $C_B$, $C_C$, and $C_D$ ensure that during any training epoch, for a given DNN $\neuralnet(\x)$, we find weights $b_k$ and functions $\phi_k(\x)$ in the polynomial $\augnet$ such that their projection onto $\neuralnet(\x)$ is zero. 
In a similar vein, constraints $C_E$, $C_F$, and $C_G$ ensure that for a given polynomial $\augnet(\x)$, we determine weights $a_j$ and functions $\psi_j(\x)$ in the neural net $\neuralnet$ such that their projection onto $\augnet(\x)$ is zero.
Constraint $C_H$ imposes the strictest orthogonality, ensuring a higher level of independence between all pairs of polynomial and DNN bases but at a greater computational cost. 

Constraints $C_B$ and $C_D$ also help regularize polynomial basis coefficients, which is beneficial when the polynomial bases contain a large number of terms. 
In our experiments, we found that these constraints, coupled with our basis truncation routine, address the curse of dimensionality by eliminating unneeded bases during training. 
However, through experimentation, we found that stringently enforcing any of these constraints often results in difficulties in training PANNs and is also computationally expensive in high-dimensional settings. 
Therefore, in this work, this discrete orthogonality is enforced through an additional weighted loss term optimized during gradient descent in conjunction with the error norm as, 
\begin{align}
\min_{\theta} \;\; || K \Psi a + K \Phi b - K f ||^2_2 + \lambda_r||\theta||_1 + \lambda_c||C||_F, \label{eq:polynn-orth-opt}
\end{align}
where $C$ is some constraint listed in~\Cref{table:orthoconstraints}, $\lambda_c$ is a Lagrange multiplier modulating the strength in which the orthogonality constraint is enforced, and $\|.\|_F$ is the Frobenius norm~\footnote{The Frobenius norm was used purely for efficiency. 
We also found that the spectral norm produced similar errors, albeit at greater expense during training}. 

\paragraph{Choosing Constraints}
A natural question is how to choose the ``right" constraint (e.g., the one that works best in a given situation). While this is in general problem-dependent, we observed some useful trends which we document here. 
The \methodname~architecture is a combination of DNNs and polynomials, both of which are likely to have overlapping approximation spaces in practice. Our weakly-enforced orthogonality constraints attempt to separate out these spaces during training by either imposing restrictions on DNN terms, polynomial terms, or both. 
For instance, in our experiments, constraint $C_H$ appeared to produce the lowest errors and the highest training times as it simultaneously attempts to control both DNN bases and coefficients and polynomial coefficients. 
In contrast, constraint $C_E$ appeared to produce the best tradeoffs between accuracy and training times, likely because $C_E$ only attempts to control the DNN coefficients and basis functions.\footnote{Constraint $C_G$ produced similar results with slightly different accuracy-training-time tradeoffs, again likely because it controls the DNN coefficients alone.} 
However, $C_E$ produced these favorable results primarily when the polynomial degree was sufficiently high (for that particular experiment); note that a higher polynomial degree corresponds to a richer polynomial space. 
In~\Cref{sec:experiments}, we present evaluations and comparisons between \methodnames~trained using the constraints listed in~\Cref{table:orthoconstraints} on both real and synthetic high-dimensional problems and show that solution accuracy is generally improved through their use with minor computational expense. We further demonstrate that the optimal choice of constraint is likely problem-dependent.

\subsection{Algorithm} \label{sec:algorithm}
We now outline key implementation details that enhance efficiency in dynamic back-propagation frameworks like PyTorch. 
\paragraph{Poisson Example}
To illustrate our algorithmic details, we consider the Poisson equation defined by $f = \Delta u$. 
The following loss function describes the forward pass of our algorithm:
\begin{align}
    \mathcal{L} &= \sum_{i=0}^{N_b} || u(x_i) - 
        \sum_{\nnind=1}^{\nnsize} a_j \psi_j{\left(\x; \theta^h \right)}
        - \augsum b_k \phi_\augind{\left(\x_i \right)} 
       ||_2^2 \notag \\
       &+ \sum_{i=0}^{N_r} || f(x_i) - \Delta 
        \sum_{\nnind=1}^{\nnsize} a_j \psi_j{\left(\x; \theta^h \right)}
        - \Delta \augsum b_k \phi_\augind{\left(\x_i \right)} 
       ||_2^2, \label{eq:forward}
\end{align}
where $a_j$ and $\psi_j$ represent the neural network basis coefficients and functions, respectively, while $b_k$ and $\phi_k$ denote the polynomial basis coefficients and functions, as detailed in~\Cref{eq:polynn}.

% Specifically, given the set of $N$ training or collocation points $\{x_i\}_{i=1}^N$, we define the $N \times m$ matrix $\Phi$ with entries $\Phi_{i,k} = \phi_k(x_i)$ as the set of all polynomial basis, and the $N \times w$ matrix $\Psi$ with entries $\Psi_{i,j} = \psi_j(x_i)$ as the set of all neural network bases.}
\paragraph{Precomputation of Polynomial Bases}
The central idea is to exploit the fact that polynomial basis functions $\phi_\augind$ for $k=1, \ldots,m$ remain constant throughout training, although the coefficients $b_k$ may change.
We enhance efficiency by precomputing and storing evaluations of these polynomial basis functions and their derivatives at the training points. 
Specifically, prior to the onset of training, given the set of collocation points $\mathbf{x}^r \in \mathbb{R}^{N^r \times d}$ and boundary points $\mathbf{x}^b \in \mathbb{R}^{N^b \times d}$, we store the polynomial bases evaluations $\Phi^r \in \mathbb{R}^{N^r \times m}$ and $\Phi^b \in \mathbb{R}^{N^b \times m}$. 
We additionally compute and store the PDE-specific derivatives, which for the Poisson problem include $\Delta \Phi^r \in \mathbb{R}^{N^r \times m}$ and $\Delta \Phi^b \in \mathbb{R}^{N^b \times m}$. 
For classical regression problems that do not incorporate network derivatives in the forward pass, this precomputation routine is consistent across various problems. 
However, in the context of PDE approximation, the precomputation of polynomial bases is tailored to specific problems. 
This precomputation eliminates the need to reevaluate basis functions or their derivatives during each training iteration, thereby reducing computational overhead and ensuring numerical consistency. 
Although standard automatic differentiation can compute derivatives of both the neural and polynomial components, our custom implementation explicitly exploits the analytical derivatives of the polynomial bases to avoid redundant differentiation. 
Automatic differentiation is used for the neural network components evaluated at the residual and boundary points ($\Psi$), while analytical derivatives are used for the polynomial component ($\Phi$), yielding efficiency gains in PDE residual evaluation.

\paragraph{Custom Automatic Differentiation}
We developed custom forward and backward passes for the PANN architectures for the efficient computation of derivatives of the polynomials in the architecture, since these can be computed analytically. 
The pseudocode for these custom routines, tailored for the Poisson problem as discussed, is presented in Algorithms~\ref{alg:forward} and~\ref{alg:backward} (although our approach is applicable to any PDE). 
In particular, line 7 of~\Cref{alg:forward} and lines 5 and 9 of~\Cref{alg:backward} utilize the precomputed polynomial basis gradients at each iteration, thus enhancing efficiency. 
This customized approach is critical for ensuring the model remains scalable and computationally feasible, especially in high-resolution PDEs or large datasets.

%%%%%%%%%%%%%%%%%%%%%%%%%%%%%%%%%%%%%%%%%%%%%%%%%%%
\begin{algorithm} 
\caption{Custom Forward Pass for PANN Optimization. \\
\textit{\small 
The forward pass evaluates the DNN and polynomial components at boundary and residual points, 
computes the corresponding data and PDE losses, and returns the total loss. 
The DNN derivatives ($\Delta \Psi$) are obtained via automatic differentiation, while polynomial derivatives ($\Delta \Phi^r$ and $\Delta \Phi^b$) are precomputed analytically and passed as input.
}}
\begin{algorithmic}[1] 
\Function{ForwardPass}{$\mathbf{x}^r$, $\mathbf{x}^b$, $\Phi^r$, $\Phi^b$, $\Delta \Phi^r$, $\Delta \Phi^b$}
\State $\mathcal{L} \gets 0$ \Comment{\footnotesize Initialize the loss function}
\State $\mathbf{u}_{\text{nn}}^b \gets \Psi(\mathbf{x}^b; \theta^h) \mathbf{a}$ 
\Comment{\footnotesize Evaluate DNN output at boundary points}
\State $\mathbf{u}_{\text{poly}}^b \gets \Phi^b \mathbf{b}$ 
\Comment{\footnotesize Evaluate polynomial output at boundary points}
\State $\mathcal{L} \gets \mathcal{L} + \| \mathbf{u}^b - (\mathbf{u}_{\text{nn}}^b + \mathbf{u}_{\text{poly}}^b) \|_2^2$ 
\Comment{\footnotesize Add boundary (data) loss terms}
\State $\mathbf{f}_{\text{nn}}^r \gets \Delta \Psi(\mathbf{x}^r; \theta^h) \mathbf{a}$ 
\Comment{\footnotesize Evaluate Laplacian of DNN output at collocation points (via autograd)} 
\State $\mathbf{f}_{\text{poly}}^r \gets \Delta \Phi^r \mathbf{b}$ 
\Comment{\footnotesize Evaluate Laplacian of polynomial output at collocation points (precomputed)}
\State $\mathcal{L} \gets \mathcal{L} + \| \mathbf{f}^r - (\mathbf{f}_{\text{nn}}^r + \mathbf{f}_{\text{poly}}^r) \|_2^2$ 
\Comment{\footnotesize Add PDE residual loss term}
\State \textbf{return} $\mathcal{L}$ 
\Comment{\footnotesize Return total loss (boundary $+$ PDE residual)} 
\EndFunction
\label{alg:forward}
\end{algorithmic} 
\end{algorithm}
%%%%%%%%%%%%%%%%%%%%%%%%%%%%%%%%%%%%%%%%%%%%%%%%%%%

%%%%%%%%%%%%%%%%%%%%%%%%%%%%%%%%%%%%%%%%%%%%%%%%%%%
\begin{algorithm}
\caption{Custom Backward Pass for PANN Optimization. \\ 
\textit{\small
The backward pass computes analytic gradients for the polynomial layer and autograd-based gradients for the DNN component. Boundary losses update coefficients and DNN weights using $\mathbf{g}^b$, while PDE residual losses propagate through Laplacian terms via $\mathbf{g}^r$. The total parameter gradients $(\nabla \theta^h, \nabla \mathbf{a}, \nabla \mathbf{b})$ are returned for updating.
}}
\label{alg:backward}
\begin{algorithmic}[1]
\Function{BackwardPass}{$\mathbf{x}^r$, $\mathbf{x}^b$, $\Phi^r$, $\Phi^b$, $\Delta \Phi^r$, $\Delta \Phi^b$}

\footnotesize
\State Initialize $\nabla \theta^h$, $\nabla \mathbf{a}$, $\nabla \mathbf{b}$ to zeros 
    \Comment{Initialize gradient accumulators for all trainable parameters.}
\vspace{0.1cm}

\State $\mathbf{g}^b \gets 2\,(\mathbf{u}^b - (\Psi(\mathbf{x}^b; \theta^h)\mathbf{a} + \Phi^b\mathbf{b}))$ 
    \Comment{Gradient of boundary loss $\partial \mathcal{L}_b / \partial u_\theta(\mathbf{x}^b)$.}
\vspace{0.1cm}

\State {$\nabla \mathbf{a} \gets -(\Psi(\mathbf{x}^b; \theta^h))^T \mathbf{g}^b$}
    \Comment{Accumulate gradient of DNN coefficients $a_j$ from boundary loss.}
\vspace{0.1cm}

\State $\nabla \mathbf{b} \gets -(\Phi^b)^T \mathbf{g}^b$ 
    \Comment{Accumulate gradient of polynomial coefficients $b_k$ from boundary loss}
\vspace{0.2cm}

\State $\nabla \theta^h \gets$ updated gradients based on $\mathbf{g}^b$ and derivative computations for $\Psi$
    \Comment{Backpropagate boundary loss through DNN weights via autograd}
\vspace{0.1cm}

\State $\mathbf{g}^r \gets 2 (\mathbf{f}^r - (\Delta \Psi(\mathbf{x}^r; \theta^h) \mathbf{a} + \Delta \Phi^r \mathbf{b}))$ 
    \Comment{Gradient of PDE residual loss $\partial \mathcal{L}_r / \partial \mathcal{F}[u_\theta](\mathbf{x}^r)$}
\vspace{0.1cm}
    
\State $\nabla \mathbf{a} \gets \nabla \mathbf{a} - (\Delta \Psi(\mathbf{x}^r; \theta^h))^T \mathbf{g}^r$ 
    \Comment{Update DNN coefficient gradients using PDE residual loss}
\vspace{0.1cm}

\State $\nabla \mathbf{b} \gets \nabla \mathbf{b} - (\Delta \Phi^r)^T \mathbf{g}^r$ 
    \Comment{Update polynomial coefficient gradients using PDE residual loss}
\vspace{0.1cm}

\State $\nabla \theta^h \gets $ updated gradients based on $\mathbf{g}^r$ and derivatives of $\Delta \Psi(\mathbf{x}^r; \theta^h)$
    \Comment{Backpropagate PDE residual loss through DNN weights (via autograd)}
\vspace{0.1cm}

\State \textbf{return} $\nabla \theta^h, \nabla \mathbf{a}, \nabla \mathbf{b}$ 
    \Comment{Return accumulated gradients for parameter updates}
\EndFunction
\end{algorithmic}
\end{algorithm}
%%%%%%%%%%%%%%%%%%%%%%%%%%%%%%%%%%%%%%%%%%%%%%%%%%%

\paragraph{Basis Pruning}
In order to combat the curse of dimensionality, we prune the DNN and (total-degree) polynomial expansions by first zeroing out any DNN coefficients $a_j$ and polynomial $b_k$ coefficients that fall below a specified threshold $t$, then discarding the corresponding basis functions. Specifically, we set the coefficients defined by $\{a_j, b_k \; | \; a_j<t \text{ for }j=1,...,w \text{ and }b_k<t \text{ for }k=1,...,m \}$ to zero and discard their corresponding basis functions. This pruning strategy helps us control the number of \methodname~basis functions in the context of high-dimensional problems; see~\Cref{fig:basisgen} for an illustration. Further, this pruning is also beneficial when solving PDEs as it allows us to require fewer derivative computations than if one had not pruned the bases.

\paragraph{Computational Complexity}
The \methodname~framework, without basis pruning, introduces additional training and evaluation complexity (over DNNs) through the inclusion of polynomials. For the Legendre polynomials used in this work, recursive evaluation of the $m-$th Legendre polynomial (or its derivative) at a single point has a computational complexity of $O(m)$, and $O(Nm)$ for $N$ points. 
In~\Cref{eq:forward}, if we set $N=\max(N_r, N_b)$, the additional computational cost of polynomial augmentation in both the forward and backward passes is $O(Nm)$ for each polynomial basis function. 
Our precomputation strategy's asymptotic computational cost remains $O(Nm)$, but is about half as expensive in the constant factors. The basis pruning strategy defined above further reduces the number of active bases while enhancing computational efficiency during training, though this is data-dependent and hard to estimate \emph{a priori}. Though we restrict ourselves to the aforementioned strategies, it is also worth noting that implementing parallel evaluation methods such as those in~\cite{bogaert20121} for $O(1)$ computation of the Legendre polynomial could improve total complexity to just $O(N)$.

\paragraph{PyTorch C++}
Our code was written in C++ using the PyTorch C++ library~\cite{paszke2019pytorch}, offering several compelling benefits over other pure Python methods for developing machine learning and deep learning applications. 
PyTorch C++ inherits many of the strengths of the Python version of PyTorch, such as its dynamic computational graph, while bypassing Python's often slow interpretation. 
The PyTorch C++ library, as of this writing, is an underutilized tool in the research community despite being significantly faster than its Python counterpart.  
Therefore, all C++ source code for our methods (and baselines) is open-source and publicly 
available at \href{https://github.com/madicooley/KernelPack}{https://github.com/madicooley/KernelPack}, facilitating extensions and further investigation for the research community.

\section{Numerical Experiments for Regression Problems}\label{sec:experiments}
In this section, we evaluate the performance of \methodnames~on a suite of carefully chosen regression benchmark problems. 
Since \methodnames~explicitly incorporate polynomials into their architecture, we begin in~\Cref{sec:exact_solution_recovery} by testing their ability to recover known polynomial functions exactly.
This test is motivated by modern RBF-FD methods, where exact orthogonality between radial and polynomial components results in polynomial reproduction and drives high-order convergence with a rate dictated by the polynomial degree~\cite{bayona2019insight}. 
To explore whether a similar relationship holds in our setting, our first experiment in~\Cref{sec:exact_solution_recovery_p1} varies both the polynomial degree and the number of training points to evaluate whether PANNs can exactly recover a known polynomial solution---despite relying on weakly-enforced, optimization-based orthogonality constraints. 
In~\Cref{sec:orths_comparison}, we extend this analysis by comparing the various orthogonality constraints in terms of their effect on polynomial recovery accuracy, training cost, and basis truncation behavior.

Next, in \Cref{sec:non_smooth_recovery}, we consider a challenging non-smooth function to test the complementary strength of the DNN in capturing features with which the polynomial bases struggle.
We follow these tests with an interpretability analysis in \Cref{sec:learned_solutions}, which dissects the learned contributions of the DNN and polynomial components in both spatial and frequency domains.
Finally, in \Cref{sec:high_dim_exps}, we assess the scalability of PANNs by applying them to high-dimensional synthetic problems and a real-world dataset. 

\begin{table}[!htpb]    
\scriptsize
\caption{\textit{Comparative analysis of various function approximation methods, delineating their objectives, advantages, and disadvantages}}\label{table:baselines} 
\vspace{-0.1cm}
\begin{center}
\resizebox{0.999\textwidth}{!}{% 
\begin{tabular}{>{\raggedright\arraybackslash}p{3.0cm} 
                  >{\raggedright\arraybackslash}p{2.0cm} 
                  >{\raggedright\arraybackslash}p{3.1cm} 
                  >{\raggedright\arraybackslash}p{3.1cm}}
\toprule
\textbf{Method} & \textbf{Objective} & \textbf{Pros} & \textbf{Cons} \\
\midrule
a. Neural Network (Tanh Activation) 
    & Minimize mean squared error (MSE). 
    & Efficient training (less risk of vanishing gradients);\newline handles high-dimensional data well. 
    & High computational load with larger networks or dimensions. \\
\addlinespace
b. Neural Networks (ReLU Activation)~\cite{nair2010rectified} 
    & Minimize mean squared error (MSE). 
    & Efficient training; handles complex, high-dimensional problems. 
    & Risk of vanishing/exploding gradients. Requires careful initialization and optimization methods~\cite{he2015delving}. \\ 
\addlinespace
c. Neural Networks (RePU Activation)~\cite{li2019better, shen2023differentiable} 
    & Minimize mean squared error (MSE). 
    & Excels at approximating smooth functions. 
    & Works poorly for non-smooth functions. \\
\addlinespace
d. Projected Polynomial ($L^2$) with Legendre Basis
    & Solve \( \langle f - \hat{f}, p \rangle = 0 \) for all \( p \) in \( P_d^{\ell} \), using quadrature. 
    & Provides high accuracy for smooth functions; efficient for regular domains and low-dimensional problems.
    & Not suited for non-smooth functions~\cite{demanet2007acha,elbrachter2021deep};
    \newline requires exponentially more points with increasing dimensions. \\
\addlinespace
e. Optimized Polynomial (PL) with Legendre Basis. 
    & Minimize mean squared error (MSE) using gradient descent. 
    & Adaptable to a broad range of problems; no structured point requirements.
    & May also struggle with very high-dimensional data; however, stochastic optimization~\cite{hu2024tackling} may ameliorate. \\
\bottomrule
\end{tabular}
}
\end{center}
\end{table}

\paragraph{Baseline Methods and Experimental Details}
All experiments in this section use the supervised loss term defined in~\cref{eq:prob-stmt}, which minimizes the squared error between the model predictions $u_\theta(x)$ and the true solution $u(x)$.
We compare our approach against a set of baseline methods summarized in \cref{table:baselines} to evaluate the effectiveness and generality of our polynomial augmentation strategy.
Our goal is to demonstrate that the polynomial layer in PANNs can enhance a variety of DNN architectures, particularly in function approximation tasks. 
To this end, we include standard DNNs with Tanh, ReLU, and RePU activation functions, capturing a range of nonlinear behaviors and expressive capacities. 
Comparing across activation functions allows us to assess how the DNN expressivity interacts with the augmented polynomial structure. 
In addition, we include two polynomial regression baselines.
The first, labeled `PL', uses a fixed Legendre basis -- similar to traditional least-squares approaches -- but computes the polynomial coefficients through gradient descent following the same training setup as the DNNs (e.g., optimizer, training iterations, learning rate), rather than using a linear solve; the goal here was to mimic the training techniques required by \methodnames.
The second, labeled $L^2$, instead computes the coefficients analytically through $L^2$ projection onto the Legendre basis (via quadrature).
This setup allows us to evaluate whether gradient-based optimization is a viable alternative to traditional projection for estimating the polynomial coefficients.  

These experiments serve not only as a benchmark for accuracy, but also as a means for highlighting the trade-offs between accuracy and modeling flexibility --- specifically, the ability to handle non-smooth target functions, unstructured domains, and high-dimensional inputs. 

We assess model performance using the relative $\ell_2$ error $\|\hat{y} - y\|_2/ \|y\|_2$, where $y$ is the vector of samples of the true solution and $\hat{y}$ contains samples of the predicted output.
Unless otherwise specified, we run five independent trials for each experiment and report the mean and standard deviation of the relative $\ell_2$ error. 
We also include wall-clock training times reported in seconds.
All experiments involving \methodnames~and the optimized polynomial (PL) incorporate both preconditioning and basis truncation. 
The $L^2$ projection tests also apply preconditioning. 
\Cref{sec:apx-expdetails} includes further information on each experiment's implementation and training details including software and hardware specifics, while~\Cref{sec:apx-additresults} lists results using conventional regression baseline methods.

\begin{figure}[!htpb]
\centering
\includegraphics[width=0.8\textwidth]{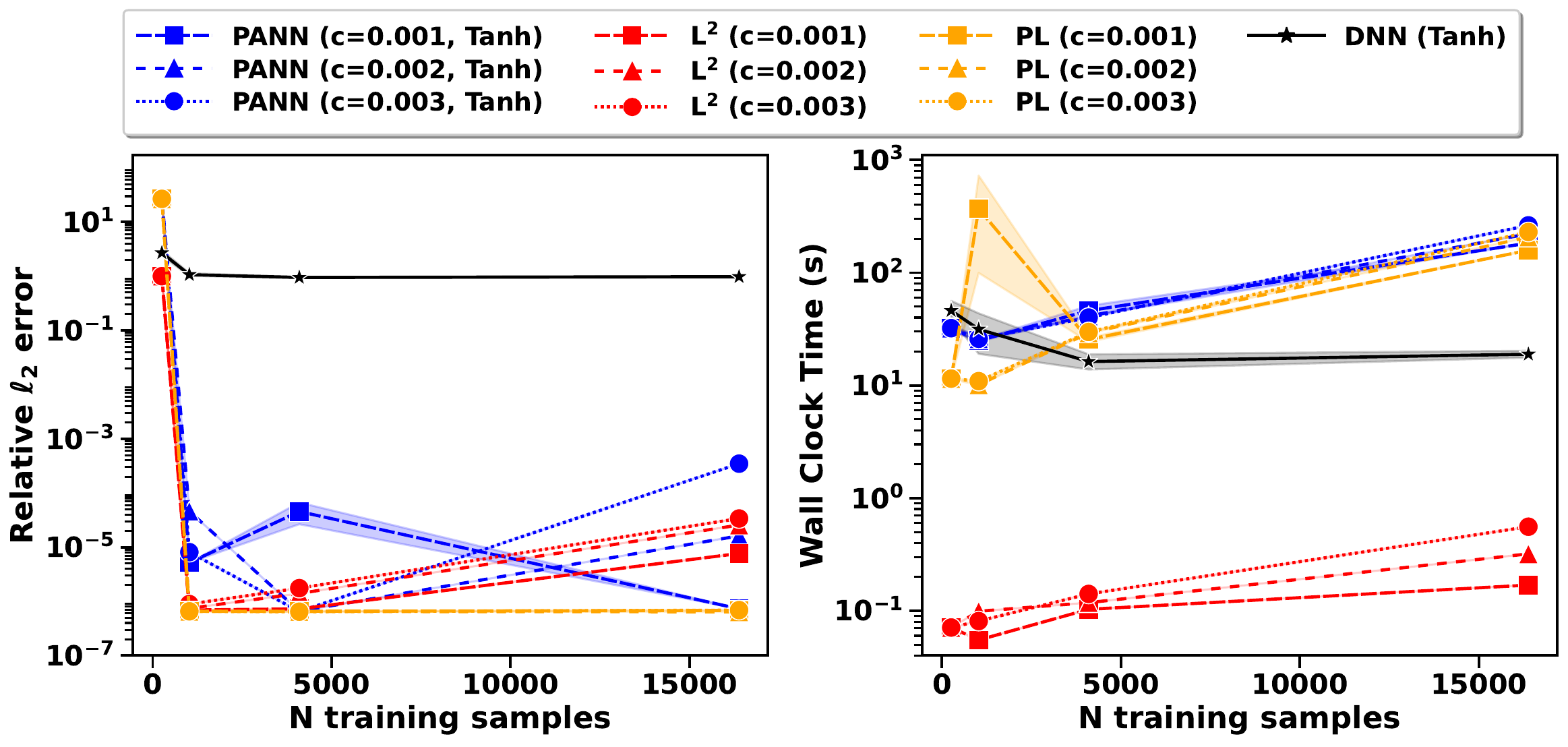}
\includegraphics[width=0.8\textwidth]{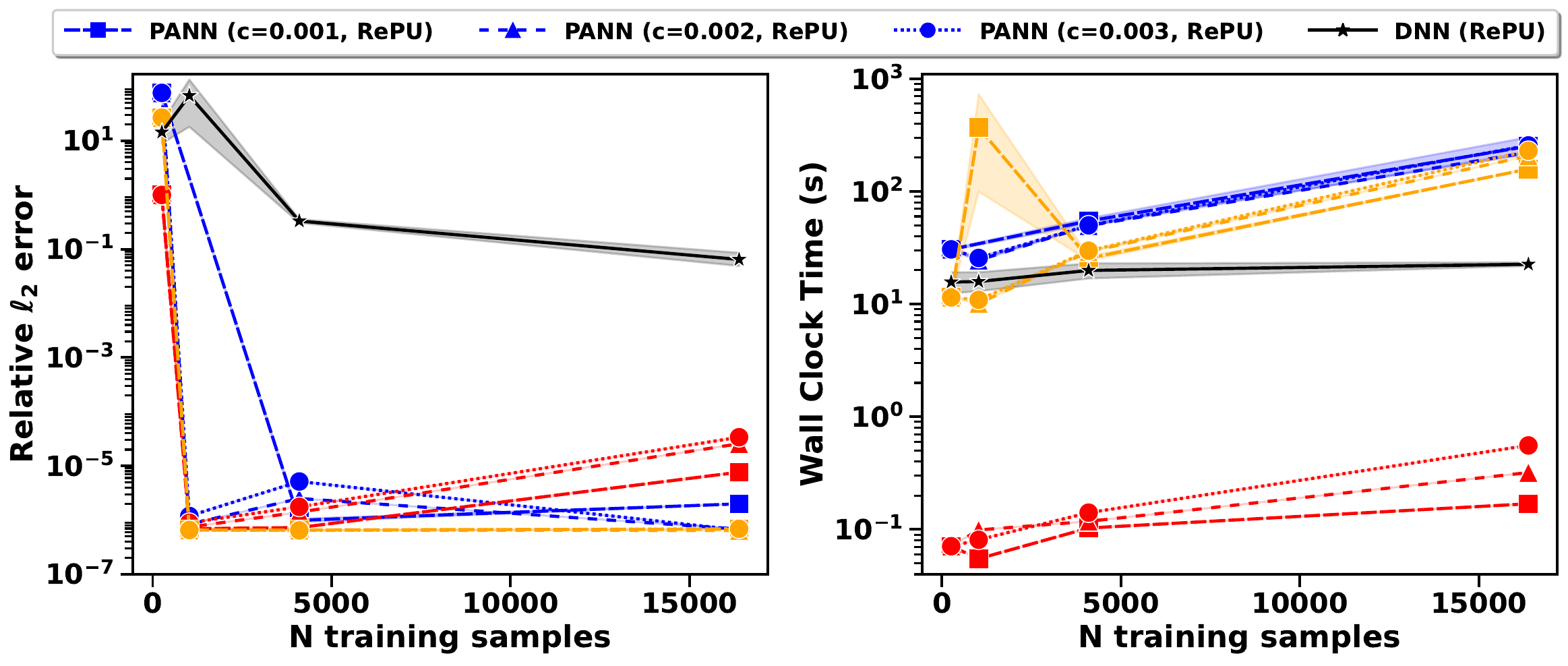}
\includegraphics[width=0.8\textwidth]{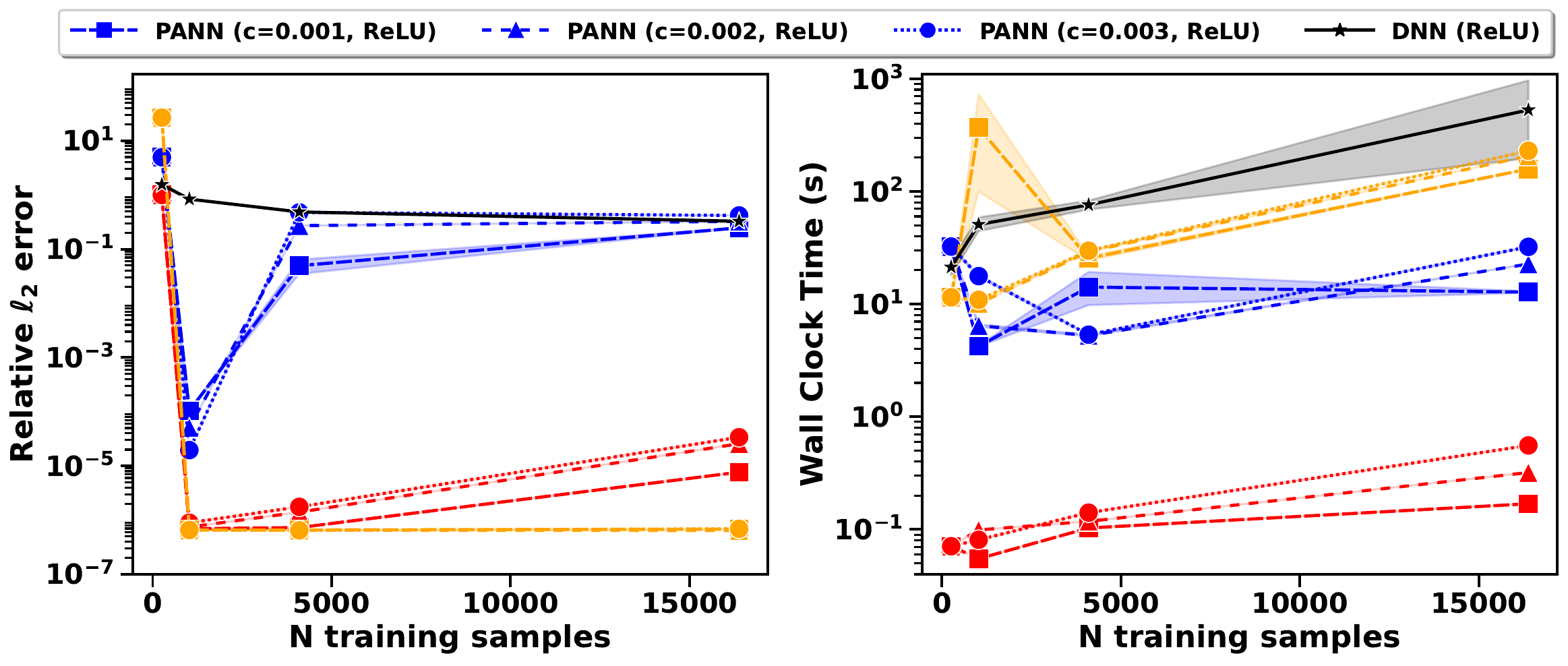}
\caption{\textit{(Left) Relative $\ell_2$ errors and (right) wall clock time in seconds for different network types using the Tanh (top), RePU (middle), and ReLU (bottom) activation function. PL and $L^2$ projection results are repeated in each figure for easy comparison. }}\label{fig:exact_exp_convergence_errors}
\vspace{-0.3cm}
\end{figure}

\subsection{Polynomial Reproduction in Neural Architectures} \label{sec:exact_solution_recovery}
In this experiment, we evaluate whether \methodnames~can recover known polynomial solutions exactly -- a property that is fundamental in many numerical approximation methods for establishing convergence behavior and important in certain engineering applications such as solid mechanics (\emph{e.g.}, the patch test for finite element methods~\cite{dvorkin2001convergence}). This test also helps us tease out the relative roles of the DNN and the polynomial expansion in the \methodname~architecture. 

Specifically, in this experiment, we consider a synthetic test case where the ground truth is a bivariate Legendre polynomial, defined as:
\begin{align}
u(x) = u(x_1,x_2) = P_{10}(x_1) P_{10}(x_2), \label{eq:test1_groundtruth}
\end{align}
such that $P_{10}$ is the degree-10 Legendre polynomial given by
\begin{align*}
P_{10}(z)&=\frac{46189}{256} z^{10} - \frac{109395}{256} z^{8} + \frac{90090}{256} z^{6}
- \frac{30030}{256} z^{4} + \frac{3465}{256} z^{2} - \frac{63}{256}. 
\end{align*} 

\subsubsection{Effect of polynomial degree and dataset size} \label{sec:exact_solution_recovery_p1}
This experiment tests whether \methodnames~can isolate and recover a high-degree polynomial solution under varying levels of approximation complexity. 
Specifically, we vary both the polynomial degree and the number of training points to assess how well the model converges to a known ground truth.
Successful recovery would indicate that the augmented polynomial remains dominant and unaffected by the DNN component, provided the basis sufficiently spans the target function (and provided the target function is itself a polynomial). 

\paragraph{Experimental Details}
We evaluate the relative $\ell_2$ error as a function of the number of training points $N \in [256, 1024, 4096, 16384]$.
Training points are sampled using Poisson disk sampling over the domain $[-1, 1]^2$ following~\cite{SFKSISC2018}.
For \methodname, the optimized polynomial (PL), and $L^2$ projection methods, we varied the total degree of Legendre bases based on $N$, with $\ell=2(\lceil cN \rceil +b)$ and $b=8$ for constants $c \in [0.001, 0.002, 0.003]$.  
Detailed configurations of total degrees $\ell$ and the corresponding widths of the polynomial layer $m$ are documented in ~\Cref{tab:exact_totaldegrees}. 
Note that the total degree of the true solution $u$ is $\ell=20$, meaning for $\ell<20$, the polynomial bases do not contain the true solution, which occurs for $N=256$ and $c=0.001, 0.002$.
For both the standard DNNs and the DNN component of \methodnames, we use three hidden layers, each with $100$ neurons, and compare RePU, ReLU, and tanh activation functions.
We also apply the orthogonality constraint $C_E$ in \methodname; we defer a detailed comparison across constraints to \Cref{sec:orths_comparison}.
Comparisons with conventional regression models are listed in~\Cref{tab:test2_baseline_comparisons}.

\paragraph{Results} 
Results in~\Cref{fig:exact_exp_convergence_errors} confirm that \methodnames~recover polynomial solutions to high accuracy, often reaching near-machine precision when using tanh or RePU activations.
This success likely stems from the smooth nature of both activations and the true solutions.  
However, \methodnames~with ReLU show reduced accuracy, likely due to the interaction between the non-smooth DNN activation and the orthogonality constraint $C_E$. 
As discussed in \Cref{sec:orth}, constraint $C_E$ focuses on controlling the DNN basis functions and coefficients, which suggests that in this example, the emphasis on DNN basis function optimization might over-complicate the learning by trying to match non-smooth DNN bases with a smooth target, thereby compromising the optimization of the polynomial layer.

In general, when equipped with smooth activation functions and an adequate number of polynomial bases (that is, when $\ell>20$), \methodnames~ can almost perfectly replicate the true solutions.
Notably, with tanh and RePU activations, \methodnames~ outperform the accuracy of $L^2$ projection, with a larger number of training points and polynomial basis functions.
The higher error in the $L^2$ projection method is likely due to numerical issues related to the large number of unnecessary polynomial bases.
The spike in training time for the PL method when $N=1024$ and $c=0.001$ is likely due to resource contentions within the system. 
The standard DNNs exhibit relatively high errors irrespective of the activation function, underscoring their inherent challenges in approximating highly oscillatory solutions. 

Remarkably, optimizing coefficients using gradient descent in the PL method in conjunction with collocation at Poisson disk samples achieves comparable accuracy to traditional $L^2$ projection methods. 
This suggests that gradient descent optimization can deliver solutions as precise as projection methods without requiring structured training points. 
In general, these results indicate that jointly optimizing both the DNN and polynomial coefficients in \methodnames~ through gradient descent is a valid approach. 
This experiment further confirms that adding a trainable DNN component does not compromise the approximation power of the polynomial layer. 

\subsubsection{Comparisons of Orthogonality Constraints}\label{sec:orths_comparison}
Next, we evaluate the impact of each discrete orthogonality constraint described in~\Cref{table:orthoconstraints} on solution accuracy and computational overhead.
These are compared against two baseline configurations: one without any constraint (`None') and one using $\ell_1$ regularization on the basis coefficients; the latter is a common choice when using approximation with high-degree polynomials or very large numbers of degrees of freedom.
For all experiments, we use $N=4096$ training points, a polynomial layer of total-degree $\ell=26$, and width $m=378$.

\begin{figure}[!htpb]
\centering
\includegraphics[width=0.9\textwidth]{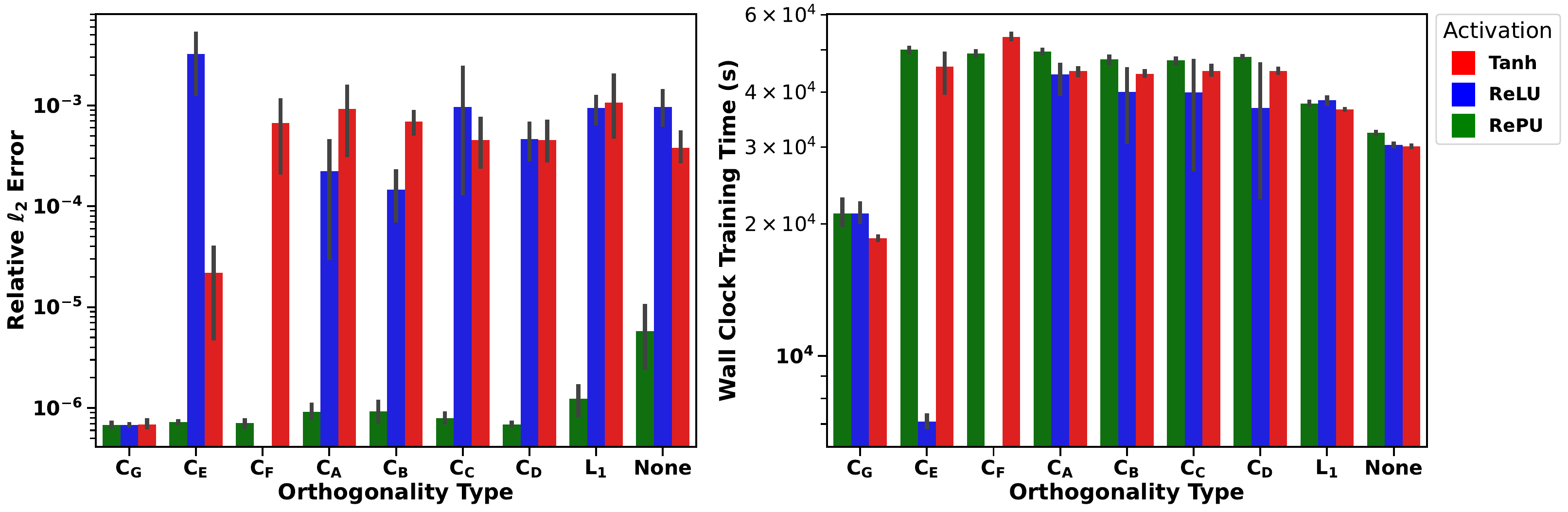}
\vspace{-0.2cm}
\caption{\textit{{(Left) Barplot showing the relative $\ell_2$ errors for each orthogonality constraint and activation, compared to using no constraint (None) and using standard $\ell_1$ regularization. The (right) barplot shows the wall clock training times of the associated method in seconds.}}} \label{fig:exact_exp_orth_comparison}
\end{figure}

\paragraph{Results}
The results in~\Cref{fig:exact_exp_orth_comparison} demonstrate that applying orthogonality generally improves approximation accuracy relative to the unconstrained or $L_1$-regularized baselines.
Across the tanh activation results, constraints $C_E$ and $C_G$ produced the lowest relative $\ell_2$ errors, while ReLU performed best with constraints $C_G$ and $C_B$.
ReLU activation paired with constraint $C_F$ did not converge, likely due to training difficulties arising from the interaction between a $C^0$ activation and the particular nature of the constraint itself. $C_F$ enforces orthogonality without directly regularizing the associated DNN coefficients $a_j$. 
As a result, $C_F$ likely exerts the weakest regularization effect on the DNN basis weights, which in turn leads to fewer DNN basis functions being pruned during training, as observed in~\Cref{tab:exact_exp_orth_comparison}.
Conversely, constraint $C_G$ achieved the highest accuracy across all activations,
shortest training times, and also exactly recovers the intended basis functions through truncation;~\Cref{tab:exact_exp_orth_comparison} shows that $100\%$ of the DNN bases were pruned, and $99.7\%$ of the polynomial bases were pruned. 
Constraints $C_E$ and $C_G$ appeared to have the strongest regularization effect on the DNN basis coefficients, leading to the highest DNN basis pruning rates across all activations.

\begin{table}[!htpb]
\footnotesize
\caption{\textit{The percentage of truncated NN and polynomial bases coefficients in the form \%NN/\%PL. Bold values represent methods who truncated the expected number of bases.}} \label{tab:exact_exp_orth_comparison}
\vspace{-0.15cm}
\begin{center}
\begin{tabular}{|l||c|c|c|c|c|c|}
\hline
Act. & $C_G$      & $C_E$       & $C_F$       & $C_A$       & $C_B$       & $C_C$         \\ \hline
Tanh & $\textbf{100/99.7}$ & $97.6/\textbf{99.7}$ & $52.4/\textbf{99.7}$ & $51.8/\textbf{99.7}$ & $52.8/99.6$ & $50.0/\textbf{99.7}$ \\[4pt]
ReLU & $\textbf{100/99.7}$ & $\textbf{100}/73.8$  & --- & $59.0/\textbf{99.7}$ & $58.6/\textbf{99.7}$ & $60.6/92.9$ \\[4pt]
RePU & $\textbf{100/99.7}$ & $53.4/\textbf{99.7}$ & $46.8/\textbf{99.7}$ & $51.0/\textbf{99.7}$ & $53.2/\textbf{99.7}$ & $50.0/\textbf{99.7}$  \\[4pt]
\hline
\end{tabular} \\[4pt]
\begin{tabular}{|l||c|c|c|}
\hline
Act. & $C_D$       & $L_1$       & None         \\ \hline
Tanh & $52.0/\textbf{99.7}$ & $47.8/99.4$ & $47.4/97.1$  \\[4pt]
ReLU & $71.8/\textbf{99.7}$ & $49.4/99.6$ & $50.6/84.2$  \\[4pt]
RePU & $53.8/\textbf{99.7}$ & $46.2/\textbf{99.7}$ & $48.2/99.6$ \\[4pt]
\hline
\end{tabular}
\end{center}
\end{table}

While the RePU activation performs best across all constraint variants, it often incurs slightly higher training times and, interestingly, does not recover the same basis sets as the true solution.
Overall, our results indicate that the activations in the DNN and the polynomial layer both model different portions of the target, and the improved error results of all constraint variations (over $\ell_1$ normalization and no constraints) indicates that the orthogonality aids optimization even in settings where the DNN and polynomial expressivity are comparable (such as those where the activation functions themselves are piecewise-polynomial).  

\begin{figure}[!htpb]
\vspace{-0.2cm}
    \centering
    \includegraphics[width=0.8\textwidth]{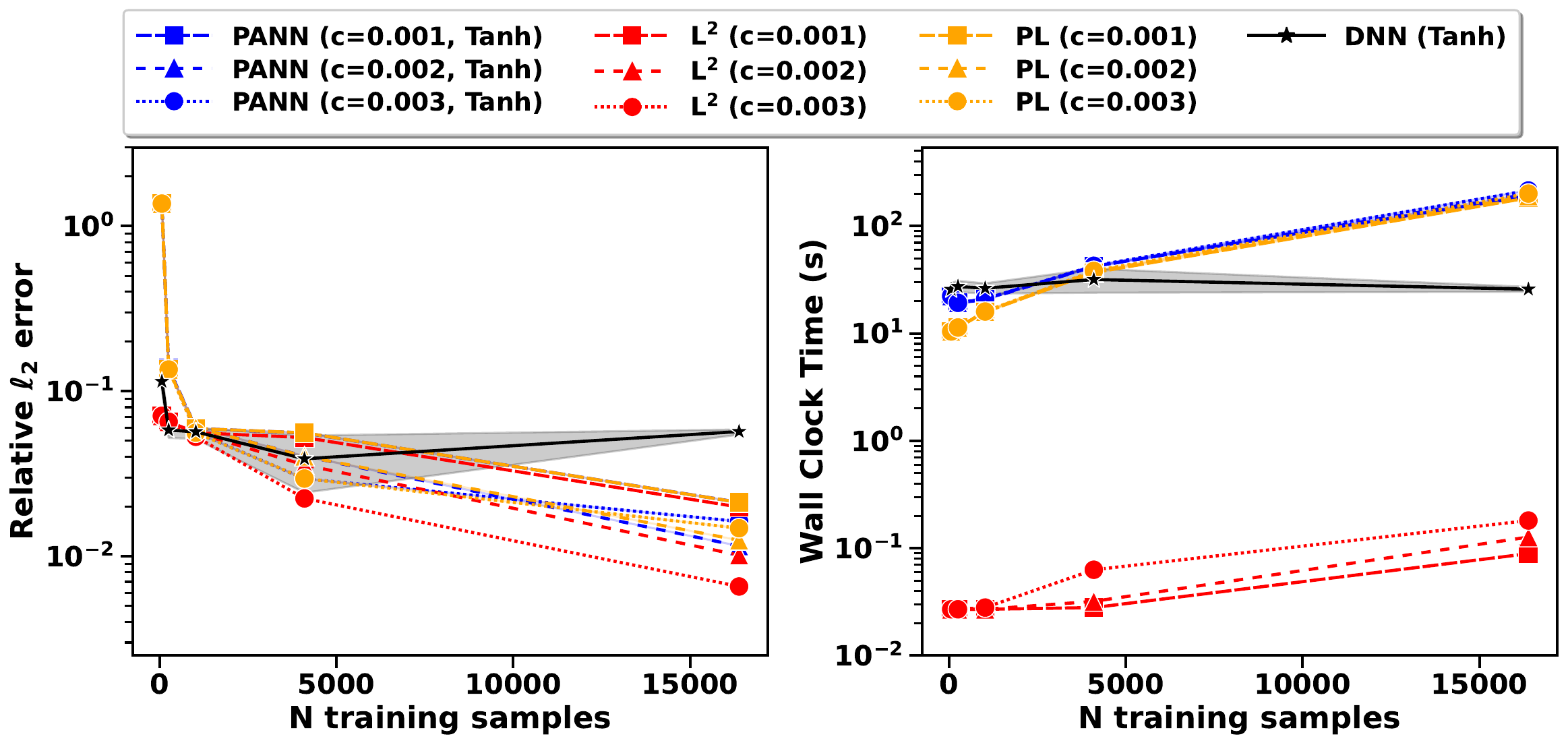}
    \includegraphics[width=0.8\textwidth]{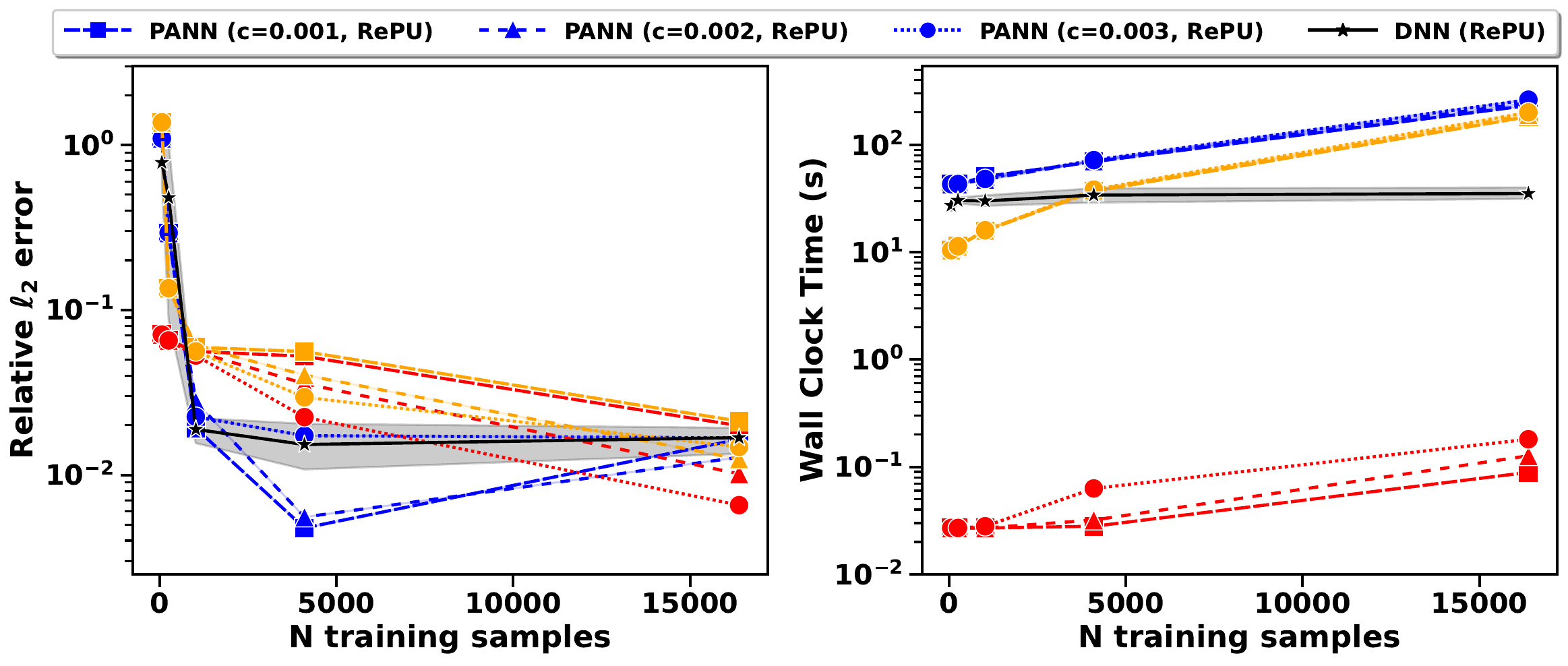}
    \includegraphics[width=0.8\textwidth]{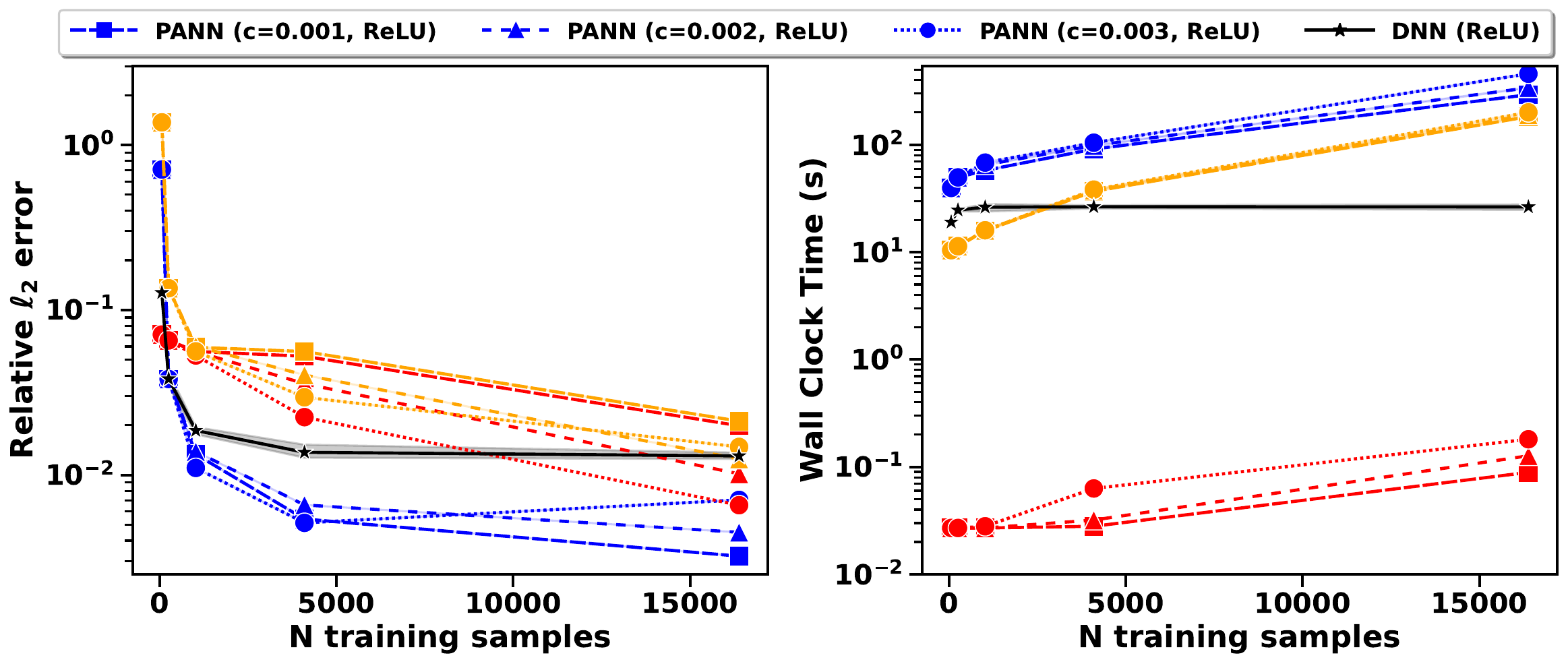}
\caption{\textit{(Left) Relative $\ell_2$ errors and (right) wall clock time in seconds for different network types using the Tanh (top), RePU (middle), and ReLU (bottom) activation function. PL and $L^2$ projection results are repeated in each figure for easy comparison. }}\label{fig:diff_exp_convergence_errors}
\vspace{-0.3cm}
\end{figure}

\subsection{Approximating Non-Smooth Functions}\label{sec:non_smooth_recovery}
The previous experiment demonstrated that PANNs recover polynomial solutions as effectively as traditional methods. 
It also confirmed that the DNN component within PANNs does not compromise polynomial solution recovery. 
Given the known flexibility of DNNs to handle complex and nonlinear functions, this test aims to explore the converse of our previous findings. 
Specifically, we want to ensure that the polynomial layer in PANNs, typically less adept at managing non-smooth functions, does not hinder the DNN portion's ability to effectively approximate these functions.
\rupdates{The results below show that PANNs remain competitive with, and often improve upon, either DNNs or polynomial approximations when approximating continuous non-smooth target functions.}
To that end, we evaluate the performance of PANNs against baseline methods by approximating the target function:
\begin{align}
u(x) = u(x_1,x_2) = x_1^2 x_2^2 \sin(1/x_1) \sin(1/x_2), \label{eq:test4_difficult}
\end{align}
which is in \rupdates{$C^0 (\mathbb{R}^2)$ and exhibits highly oscillatory behavior with vanishing amplitude near the coordinate axis due to the $x_1^2x_2^2$ factor, while its derivatives are not well behaved near $x_1=0$ and $x_2=0$.}
We use a consistent experimental setup to~\Cref{sec:exact_solution_recovery}, but varied the total degree of the Legendre bases by $\ell = \lceil cN \rceil +b$ for $b=8$. 
All total degree and polynomial layer width configurations are documented in ~\Cref{tab:diff_totaldegrees}. 

\rupdates{
As shown in~\Cref{fig:diff_exp_convergence_errors}, standard DNNs outperform $L^2$ projection at lower spatial resolutions when a piecewise-smooth activation is used, while \methodnames~remain competitive and often achieve the lowest errors, particularly for the ReLU activation. 
These results indicate that the appended polynomial layer does not prevent the DNN component from approximating continuous non-smooth targets.
}
This trend holds when \methodnames~are compared to other methods as well; see ~\Cref{tab:test4_baseline_comparisons} for these results.

\rupdates{
At the same time, the oscillatory component in~\Cref{eq:test4_difficult} is strongly damped by the factor $x_1^2x_2^2$, so although the oscillations are most rapid near the coordinate axes, their amplitudes become small there. 
To test a less damped and therefore more challenging family of continuous oscillatory targets, we additionally consider:
\begin{align}
u_\alpha(x_1,x_2) = 
|x_1|^\alpha |x_2|^\alpha \sin(1/x_1)\sin(1/x_2),
\label{eq:test4_difficult_alpha}
\end{align}
where decreasing $\alpha$ weakens the amplitude damping while preserving continuity. 
The original target~\Cref{eq:test4_difficult} corresponds to $\alpha=2$. 
\Cref{tab:diff_alpha_sensitivity} reports results for $N=16384$ across several values of $\alpha$.
}

\begin{table}[t]
\centering
\small
\caption{\rupdates{Results for the target family in~\Cref{eq:test4_difficult_alpha} using \(N=16384\) training samples. Relative \(\ell_2\) errors are reported as mean with standard deviation in parentheses, except for the deterministic \(L^2\) projection baseline, for which a single value is reported.}}
\label{tab:diff_alpha_sensitivity}
\begin{tabular}{l|lll}
\hline
Method & \(\alpha=2\) & \(\alpha=1.5\) & \(\alpha=1\) \\
\hline
\multicolumn{4}{c}{Tanh activation} \\
\hline
DNN                 & 0.040860 (0.000259) & 0.030773 (0.001344) & 0.036613 (0.000581) \\
PANN (\(c=0.001\))  & 0.020306 (0.000073) & \textbf{0.025108} (0.000710) & \textbf{0.032573} (0.000711) \\
PANN (\(c=0.002\))  & \textbf{0.012908} (0.002035) & 0.030429 (0.000239) & 0.046035 (0.000910) \\
PANN (\(c=0.003\))  & 0.014497 (0.002834) & 0.034226 (0.000633) & 0.046316 (0.000581) \\
\hline
\multicolumn{4}{c}{RePU activation} \\
\hline
DNN                 & 0.012838 (0.002565) & \textbf{0.021332} (0.003331) & 0.046655 (0.003624) \\
PANN (\(c=0.001\))  & 0.019525 (0.001445) & 0.025357 (0.000175) & \textbf{0.041454} (0.000225) \\
PANN (\(c=0.002\))  & \textbf{0.012809} (0.000745) & 0.029679 (0.000275) & 0.044312 (0.000895) \\
PANN (\(c=0.003\))  & 0.015151 (0.001645) & 0.056526 (0.020780) & 0.051870 (0.000981) \\
\hline
\multicolumn{4}{c}{ReLU activation} \\
\hline
DNN                 & 0.009701 (0.000575) & 0.020773 (0.001488) & 0.046888 (0.002877) \\
PANN (\(c=0.001\))  & \textbf{0.003260} (0.000239) & \textbf{0.005808} (0.000890) & \textbf{0.021454} (0.000225) \\
PANN (\(c=0.002\))  & 0.003720 (0.000046) & 0.017036 (0.000175) & 0.033393 (0.000145) \\
PANN (\(c=0.003\))  & 0.007710 (0.000172) & 0.019830 (0.008301) & 0.069718 (0.000611) \\
\hline
\multicolumn{4}{c}{Activation-independent baselines} \\
\hline
PL (\(c=0.001\))    & 0.021179 (0.000432) & 0.045675 (0.000670) & 0.104135 (0.000257) \\
PL (\(c=0.002\))    & 0.012744 (0.002252) & 0.037550 (0.002411) & 0.097370 (0.003306) \\
PL (\(c=0.003\))    & 0.015360 (0.006130) & 0.032664 (0.004616) & 0.150175 (0.005338) \\
\(L^2\) (\(c=0.001\)) & 0.020276 & 0.04430000 & 0.10053659 \\
\(L^2\) (\(c=0.002\)) & 0.010453 & 0.02743145 & 0.07312599 \\
\(L^2\) (\(c=0.003\)) & 0.007969 & 0.02325550 & 0.06796853 \\
\hline
\end{tabular}
\end{table}

\rupdates{
The ablation in~\Cref{tab:diff_alpha_sensitivity} shows that decreasing $\alpha$ generally makes the approximation problem more difficult, but the effect on the relative performance between the PANN and DNN depends on both the activation and the polynomial degree. 
A consistent trend is that smaller or moderate polynomial subspaces remain competitive across the tested values of $\alpha$, whereas larger values of $c$ become less reliable as $\alpha$ decreases. 
This behavior is consistent with the structure of the target family: although $u_\alpha$ retains a broad low-frequency, quadrant-scale component, its most difficult features remain localized near the coordinate axes, and these oscillatory features account for a larger fraction of the signal as the damping weakens. 
Since the polynomial component is built from global Legendre bases, increasing $c$ adds expressive power in a global manner, which is not the most efficient way to resolve localized high-frequency structure. 
As a result, larger polynomial subspaces do not necessarily improve accuracy and can even make the pruning problem more difficult, while smaller PANNs can still benefit from using the polynomial branch to capture the coarse global component and leaving the localized oscillatory residual to the DNN.
}

Regarding PANN’s sensitivity to the polynomial degree parameter $c$ (which controls the expressiveness of the polynomial subspace), ~\Cref{fig:exact_exp_convergence_errors} and ~\Cref{fig:diff_exp_convergence_errors} show that model accuracy remains stable for moderate values, with two distinct scenarios: small $c$ under-represents the target function, while large $c$ introduces redundant bases that the pruning procedure may struggle to remove. 

Once the polynomial subspace becomes sufficiently expressive and the training set size is large relative to the number of polynomial bases, performance saturates and becomes effectively insensitive to further increases in $c$ \rupdates{for problems whose dominant structure is well represented by the global polynomial component. 
For the less damped family~\Cref{eq:test4_difficult_alpha}, however, the results suggest that larger polynomial subspaces are less effective when the dominant difficult features are localized oscillations rather than broad smooth structure. 
Overall, our results indicate that \methodnames~retain the flexibility needed to approximate continuous non-smooth targets, and that their benefit is strongest when the target contains global structure that is well aligned with the polynomial subspace, together with additional features that are more naturally handled by the DNN.
}

\subsection{Frequency Analysis}\label{sec:learned_solutions}
\begin{figure}[!htpb]
\centering
\includegraphics[width=0.75\textwidth]{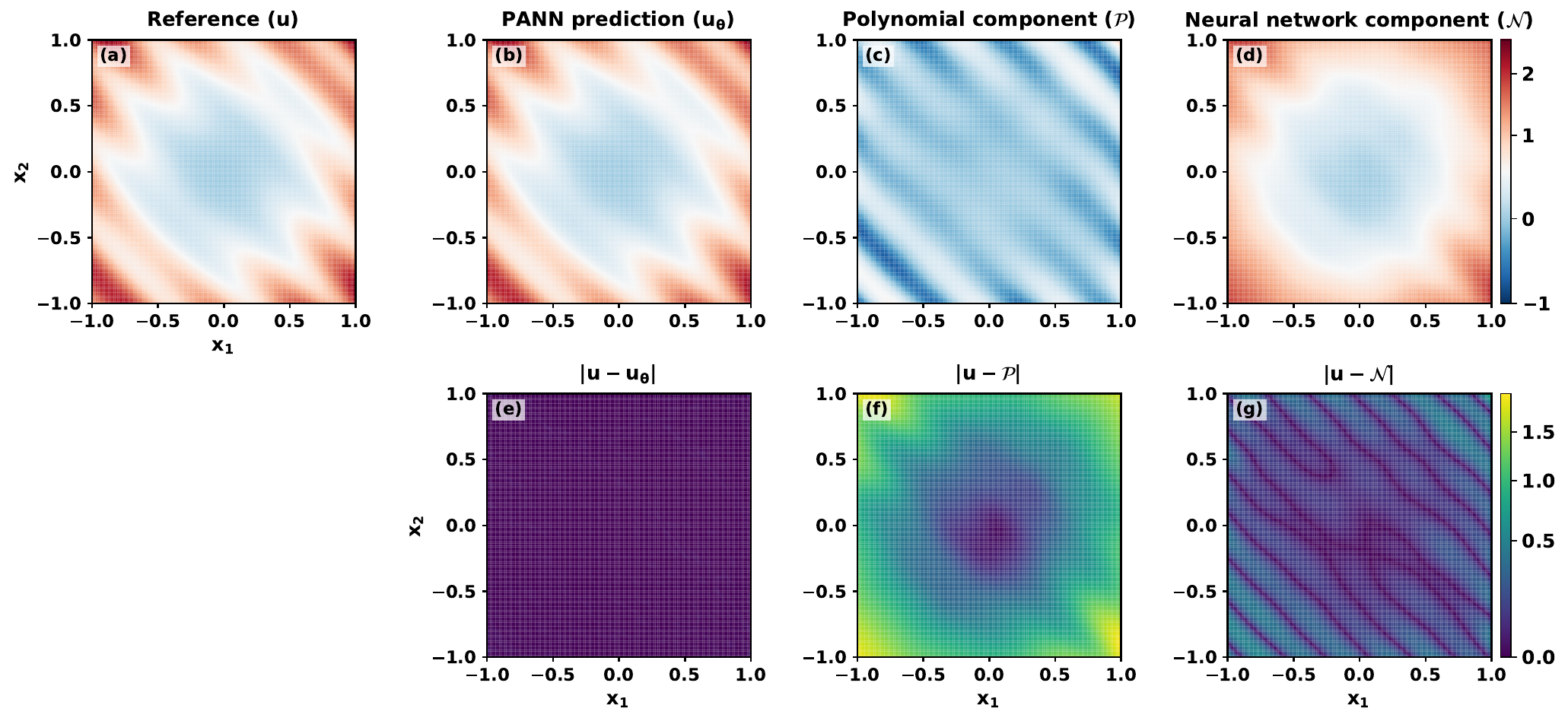}
\caption{\textit{ 
\rupdates{
Spatial-domain decomposition of the learned PANN solution for the target function in \Cref{eq:test4v2_difficult}.
Top row (left to right): reference solution, full PANN prediction, polynomial component, and neural component.
Bottom row: absolute error in the full PANN prediction, polynomial component relative to the reference, and neural component relative to the reference.
}}}\label{fig:predicted_solutions_test4}
\includegraphics[width=0.75\textwidth]{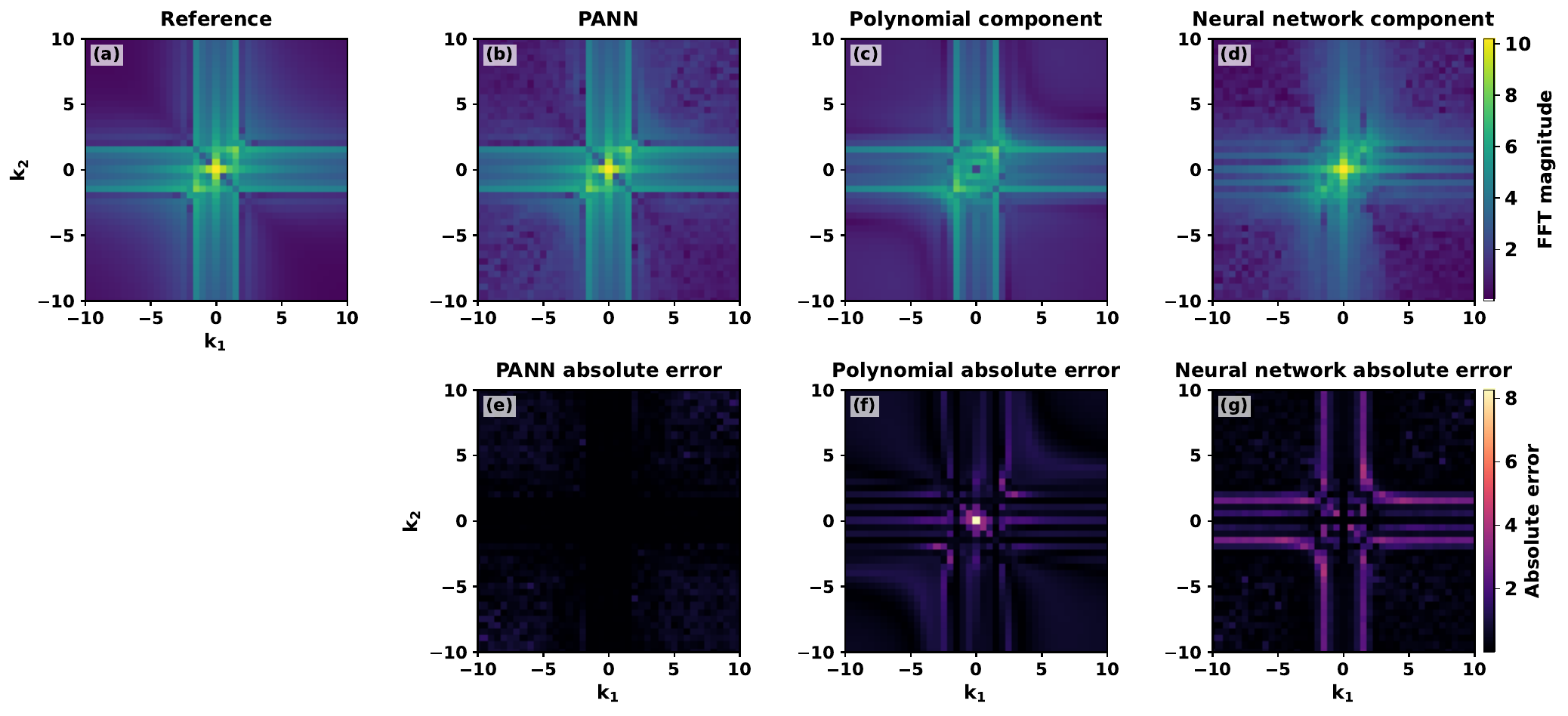}
\vspace{-0.20cm}
\caption{\textit{
\rupdates{
Frequency-domain analysis of the learned PANN decomposition.
Top row: magnitude of the 2D FFT for (a) the reference solution, (b) the full PANN prediction, (c) the polynomial component, and (d) the neural component.
Bottom row: absolute spectral error relative to the reference for (e) the full PANN prediction, (f) the polynomial component, and (g) the neural component.
$k_1$ and $k_2$ denote the Fourier mode indices in the two coordinate directions.
}}}\label{fig:predicted_solutions_fft_test4}
\end{figure}

\rupdates{
The results in \Cref{sec:non_smooth_recovery} suggest that the benefit of the polynomial branch depends not only on the frequency content of the target, but also on how its difficult features are distributed across the domain. 
In particular, the less-damped family in \Cref{sec:non_smooth_recovery} showed that when the most challenging oscillations are localized, enlarging the global polynomial subspace does not necessarily improve performance. 
To better understand the complementary roles of the polynomial and neural components in a setting that is more naturally aligned with a global polynomial basis, we now consider a target whose oscillatory structure is distributed more broadly across the domain and analyze the learned PANN decomposition in both the spatial and frequency domains.
Specifically, we consider
\begin{align}
u(x) = u(x_1,x_2) = r^2 + 0.35\, r^2 \sin\!\big(10(x_1+x_2)\big), 
\qquad r^2 = x_1^2 + x_2^2. \label{eq:test4v2_difficult}
\end{align}
This function combines a smooth quadratic with an oscillatory term that is present throughout the domain. 
As a result, the target exhibits a nontrivial combination of broad low-frequency structure and globally distributed oscillatory behavior, as shown in \Cref{fig:predicted_solutions_test4}a.
 We train a PANN to approximate~\Cref{eq:test4v2_difficult} using $4{,}096$ data points, a polynomial total degree of $\ell=21$, a ReLU activation in the DNN component, and three hidden layers of width $100$. 
 \Cref{fig:predicted_solutions_test4} shows the resulting learned decomposition into the full PANN prediction and its polynomial and neural components. 
 For this example, the polynomial component captures much of the broad spatial structure of the target, including a substantial portion of the globally distributed oscillatory pattern, while the neural component provides a complementary correction that improves the overall approximation.
}

\rupdates{
\Cref{fig:predicted_solutions_fft_test4} shows the resulting frequency-domain representations. 
The polynomial component (\Cref{fig:predicted_solutions_fft_test4}c) reproduces much of the dominant spectral structure and retains relatively stronger energy away from the central modes, indicating that, for this example, the polynomial branch captures a substantial share of the globally distributed oscillatory content. 
In contrast, the neural component (\Cref{fig:predicted_solutions_fft_test4}d) is more concentrated near the central modes, suggesting that it contributes more prominently to the lower-frequency portion of the learned representation. 
The spectral error plots support this interpretation: the polynomial component (\Cref{fig:predicted_solutions_fft_test4}f) exhibits larger error near the central low-frequency modes, whereas the neural component (\Cref{fig:predicted_solutions_fft_test4}g) shows larger error farther from the center and along the stronger oscillatory spectral structures. 
Taken together, these results suggest that when oscillatory features are distributed more globally across the domain, the polynomial layer can participate much more directly in representing them, while the neural network provides a complementary correction to the remaining lower-frequency and residual structure.
}

\rupdates{
To further analyze this learned decomposition, we examine the solutions in frequency space by computing the two-dimensional fast Fourier transform (FFT) of the reference solution, the full PANN prediction, and the individual polynomial and neural components. Although the target is not strictly periodic on $[-1,1]^2$, the FFT is still useful here as a diagnostic tool for comparing the relative spectral footprints of the learned components. 
Our goal is not to interpret individual Fourier coefficients as exact modal amplitudes on a periodic domain, but rather to determine whether the polynomial and neural branches emphasize different portions of the spectrum. 
Because the same transformation is applied uniformly to the reference, full prediction, and both learned components, any boundary-induced spectral leakage affects all of them in a comparable way, so relative comparisons remain informative.
}

\rupdates{
\Cref{fig:predicted_solutions_fft_test4} shows the resulting frequency-domain representations. 
The polynomial component (\Cref{fig:predicted_solutions_fft_test4}c) reproduces much of the dominant spectral structure and retains relatively stronger energy away from the central modes, indicating that, for this example, the polynomial branch captures a substantial share of the globally distributed oscillatory content. 
In contrast, the neural component (\Cref{fig:predicted_solutions_fft_test4}d) is more concentrated near the central modes, suggesting that it contributes more prominently to the lower-frequency portion of the learned representation. 
The spectral error plots support this interpretation: the polynomial component (\Cref{fig:predicted_solutions_fft_test4}f) exhibits larger error near the central low-frequency modes, whereas the neural component (\Cref{fig:predicted_solutions_fft_test4}g) shows larger error farther from the center and along the stronger oscillatory spectral structures. 
Taken together, these results suggest that when oscillatory features are distributed more globally across the domain, the polynomial layer can participate much more directly in representing them, while the neural network provides a complementary correction to the remaining lower-frequency and residual structure.
}

\rupdates{
While this observed division of spectral content is informative, it is not fixed a priori and depends strongly on the expressiveness of the polynomial subspace. 
In particular, the total degree $\ell$ and the chosen polynomial family determine how effectively the polynomial layer can represent oscillatory behavior in the target function. 
When the polynomial subspace is sufficiently expressive, the polynomial layer can capture a larger share of the oscillatory content. 
If the polynomial subspace is too limited, however, the neural network must compensate for unresolved structure, weakening this separation of roles. 
Conversely, an overly rich polynomial subspace may introduce redundancy and degrade training efficiency. 
This trade-off motivates the basis-pruning strategy described in \Cref{sec:algorithm}, which removes inactive basis functions during training and helps balance expressiveness with computational cost.
}

\rupdates{
Because the target function is not periodic on $[-1,1]^2$, the FFT-based analysis should be interpreted with some caution due to boundary-induced spectral leakage. 
To provide a complementary view that is better aligned with the bounded non-periodic approximation setting, we also analyzed the learned decomposition in a two-dimensional total-degree Legendre basis on $[-1,1]^2$ by computing least-squares coefficients and aggregating modal energy by total degree (not shown).
This polynomial-mode analysis was broadly consistent with the FFT results: the polynomial component carries relatively more high-order content, while the neural component is more concentrated in lower-order content, and the full PANN combines both contributions. 
}

\subsection{High-Dimensional Approximation} \label{sec:high_dim_exps}
We now present results for approximating target functions defined on high-dimensional spaces. Specifically, we
present results for both a synthetic target function with spatial dimension ranging from 2 to 6 and a real-world target function dependent on 8 input variables.

\subsubsection{High-Dimensional Synthetic Example} \label{sec:highdim-results}
In this section, we explore the capabilities of PANN and DNNs using tanh, ReLU, and RePU activations, and optimized polynomials (PLs), in approximating a smooth target function $u: \mathbb{R}^{d+1} \to \mathbb{R}$ of the form:
\vspace{-0.25cm}
\begin{align}
    u(\x) = 5 \pi^2 \sin(2\pi \x_0) \prod_{i=1}^d \sin(\pi \x_i),\label{eq:highdim_truesolution}
\end{align} 
where the dimension $d$ is allowed to vary from $d=2$ to $d=6$ and the domain was $[-1,1]^{d+1}$.
We set the number of training points to $N=1536, 3072, 6144,$ $12288$, and $26576$ for $d=2,3,4,5,$ and $6$, respectively. 
For the \methodnames, we used the orthogonality constraint $C_E$, with a fixed total degree for the polynomial bases set at $\ell=8$. 
This corresponds to polynomial layers of width $m=45, 165, 495, 1287, 3003$ for $d=2$ through $d=6$, respectively. 
For both the standard DNNs and the DNN component of \methodnames, we use three hidden layers each with $100$ neurons.
Comparisons against various popular regression models are also presented in~\Cref{tab:highdim_baseline_comparisons}. 
We excluded $L^2$ projections from the comparison as they became prohibitively expensive in this setting and required high-dimensional quadrature; while there are established techniques to ameliorate these issues, such comparisons are beyond the scope of this work. 
Our focus remains on straightforward, scalable methodologies that require minimal preprocessing and are suitable for various high-dimensional contexts. 

\begin{figure}[!htpb]
\vspace{-0.2cm}
\centering
\includegraphics[width=0.9\textwidth]{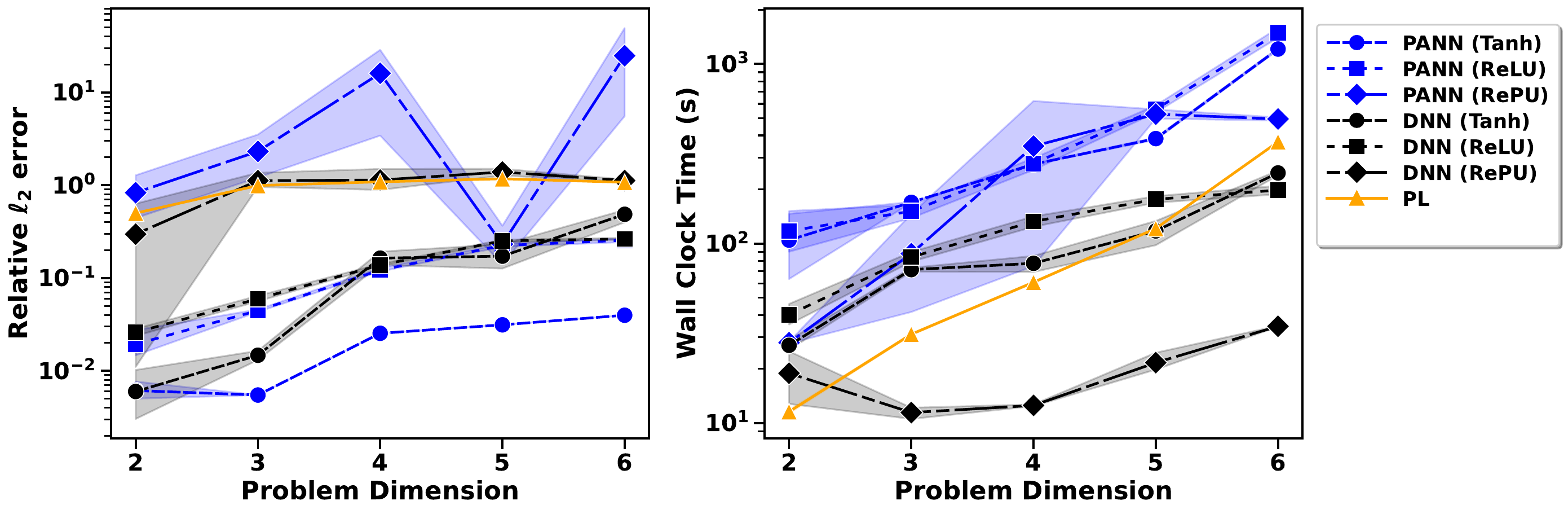}
\vspace{-0.2cm}
\caption{\textit{(Left) Relative $\ell_2$ error by problem dimension and (right) wall clock time in seconds by problem dimension for DNNs and \methodnames~using Tanh, ReLU, and RePU activation functions compared to the polynomial layer (PL).}}\label{fig:highdim_exp_convergence_errors}
\vspace{-0.4cm}
\end{figure}

As illustrated in~\Cref{fig:highdim_exp_convergence_errors}, \methodname~with tanh activation once again consistently produced the lowest relative $\ell_2$ errors across all tested dimensions, albeit with a modest increase in both error and computational time as the problem dimension increased. 
In contrast, \methodname~equipped with RePU activation demonstrated higher errors and more significant increases in computational time, making it less suited for scaling to larger dimensions. 

\subsubsection{Noisy Real-World Example} \label{sec:realworld-results}
We also tested the effectiveness of \methodnames~in predicting housing prices using the California housing dataset~\cite{califhousingdataset1997}, which includes $20,640$ instances and eight features, such as the number of bedrooms and occupancy rates. 
We compare different orthogonality constraints (as described in~\Cref{table:orthoconstraints}) under preconditioned and non-preconditioned settings and benchmark against standard DNNs with ReLU and Tanh activations and PLs with varying polynomial bases. 
RePU performed uniformly poorly on this real-world dataset and was excluded from the results.
For both the standard DNNs and PANNs, we use three hidden layers of $100$ neurons, with each feature normalized to the range $[-1,1]$. 
We performed four-fold cross-validation with $15,480$ training and $5,152$ test samples per fold. 
\Cref{tab:realworld_baseline_comparisons_full} in the Appendix presents the error results compared to various popular regression models.
\begin{table}[!htpb]
\scriptsize
\caption{\textit{Relative $\ell_2$ errors and training times for \methodnames~for each constraint with and without preconditioning, DNNs and polynomial layers (PLs) under various settings.}}\label{tab:realworld_baseline_comparisons}
\vspace{-0.15cm}
\begin{center}
\resizebox{0.75\textwidth}{!}{%
\begin{tabular}{|l|c|c|c|c|}
\toprule 
\textbf{Network} & \textbf{Act} & \textbf{Precond} & \textbf{Relative $\ell_2$ Error} & \textbf{Wall-clock Time (s)} \\ 
\toprule 
\multirow[c]{4}{*}{PANN-$C_A$} & \multirow[c]{2}{*}{ReLU} & \cellcolor{tblrowcolor} False & \cellcolor{tblrowcolor} $0.2256\pm0.002$ & \cellcolor{tblrowcolor} $165.5032\pm272.636$  
    \\ & & True & $0.2126\pm0.002$ &  $1075.4628\pm1861.269$  \\[4pt]
    & \multirow[c]{2}{*}{Tanh} & \cellcolor{tblrowcolor} False & \cellcolor{tblrowcolor} $0.2249\pm0.019$ & \cellcolor{tblrowcolor} $126.4284\pm187.072$  
    \\ & & True & $0.2202\pm0.003$ &  $1546.4286\pm2925.925$  \\ \midrule
\multirow[c]{4}{*}{PANN-$C_B$} & \multirow[c]{2}{*}{ReLU} & \cellcolor{tblrowcolor} False & \cellcolor{tblrowcolor} $0.2164\pm0.006$ & \cellcolor{tblrowcolor} $152.2832\pm242.852$ 
    \\ & & True & $0.2139\pm0.003$ &  $1084.5254\pm1884.902$  \\[4pt] 
    & \multirow[c]{2}{*}{Tanh} & \cellcolor{tblrowcolor} False & \cellcolor{tblrowcolor} $0.2200\pm0.009$ & \cellcolor{tblrowcolor} $114.0070\pm196.716$  
    \\ & & True & $0.2209\pm0.003$ &  $1063.0726\pm1855.294$  \\ \midrule
\multirow[c]{4}{*}{PANN-$C_C$} & \multirow[c]{2}{*}{ReLU} & \cellcolor{tblrowcolor} False & \cellcolor{tblrowcolor} $0.2213\pm0.004$ & \cellcolor{tblrowcolor} $151.7962\pm246.296$  
    \\ & & True & $0.2124\pm0.002$ &  $238.2235\pm4.155$  \\[4pt]
    & \multirow[c]{2}{*}{Tanh} & \cellcolor{tblrowcolor} False & \cellcolor{tblrowcolor} $0.2639\pm0.003$ & \cellcolor{tblrowcolor} $132.9082\pm206.539$  
    \\ & & True & $0.2239\pm0.004$ &  $231.4235\pm2.245$  \\ \midrule
\multirow[c]{4}{*}{PANN-$C_D$} & \multirow[c]{2}{*}{ReLU} & \cellcolor{tblrowcolor} False & \cellcolor{tblrowcolor} $0.2615\pm0.006$ & \cellcolor{tblrowcolor} $160.4952\pm266.624$  
    \\ & & True & $0.2120\pm0.003$ &  $241.8500\pm1.410$  \\[4pt]
    & \multirow[c]{2}{*}{Tanh} & \cellcolor{tblrowcolor} False & \cellcolor{tblrowcolor} $0.2645\pm0.003$ & \cellcolor{tblrowcolor} $142.3642\pm225.477$  
    \\ & & True & $0.2183\pm0.004$ &  $233.1430\pm2.675$  \\ \midrule
\multirow[c]{4}{*}{PANN-$C_E$} & \multirow[c]{2}{*}{ReLU} & \cellcolor{tblrowcolor} False & \cellcolor{tblrowcolor} $0.2122\pm0.003$ & \cellcolor{tblrowcolor} $190.5806\pm322.992$  
    \\ & & True & $\mathbf{0.2118\pm0.005}$ &  $242.8420\pm3.402$  \\[4pt]
    & \multirow[c]{2}{*}{Tanh} & \cellcolor{tblrowcolor} False & \cellcolor{tblrowcolor} $0.2319\pm0.013$ & \cellcolor{tblrowcolor} $128.1918\pm190.995$  
    \\ & & True & $0.2225\pm0.006$ &  $239.5772\pm1.154$  \\ \midrule
\multirow[c]{4}{*}{PANN-$C_F$} & \multirow[c]{2}{*}{ReLU} & \cellcolor{tblrowcolor} False & \cellcolor{tblrowcolor} $0.2153\pm0.002$ & \cellcolor{tblrowcolor} $\mathbf{44.3470\pm0.790}$  
    \\ & & True & $0.2123\pm0.003$ &  $242.0880\pm3.468$  \\[4pt]
    & \multirow[c]{2}{*}{Tanh} & \cellcolor{tblrowcolor} False & \cellcolor{tblrowcolor} $0.2291\pm0.002$ & \cellcolor{tblrowcolor} $157.4134\pm260.070$  
    \\ & & True & $0.2288\pm0.004$ &  $233.9955\pm4.150$  \\ \midrule
\multirow[c]{4}{*}{PANN-$C_G$} & \multirow[c]{2}{*}{ReLU} & \cellcolor{tblrowcolor} False & \cellcolor{tblrowcolor} $0.2450\pm0.020$ & \cellcolor{tblrowcolor} $92.6790\pm143.156$  
    \\ & & True & $0.2126\pm0.001$ &  $233.2752\pm4.080$  \\[4pt]
    & \multirow[c]{2}{*}{Tanh} & \cellcolor{tblrowcolor} False & \cellcolor{tblrowcolor} $0.2189\pm0.004$ & \cellcolor{tblrowcolor} $133.5222\pm221.093$  
    \\ & & True & $0.2214\pm0.003$ &  $225.1395\pm2.406$  \\ \midrule
\multirow[c]{4}{*}{PANN-$C_H$} & \multirow[c]{2}{*}{ReLU} & \cellcolor{tblrowcolor} False & \cellcolor{tblrowcolor} $0.2120\pm0.004$ & \cellcolor{tblrowcolor} $360.6397\pm465.925$  
    \\ & & True & $0.2123\pm0.006$ &  $330.7845\pm5.975$  \\[4pt] 
    & \multirow[c]{2}{*}{Tanh} & \cellcolor{tblrowcolor} False & \cellcolor{tblrowcolor} $0.2253\pm0.004$ & \cellcolor{tblrowcolor} $278.8388\pm345.442$  
    \\ & & True & $0.2268\pm0.001$ &  $323.1293\pm3.634$  \\ \midrule
\multirow[c]{2}{*}{DNN} & ReLU & \cellcolor{tblrowcolor} False & \cellcolor{tblrowcolor} $0.2132\pm0.003$ & \cellcolor{tblrowcolor} $317.4062\pm23.341$  \\[4pt] 
& Tanh & \cellcolor{tblrowcolor} False & \cellcolor{tblrowcolor} $0.2157\pm0.003$ & \cellcolor{tblrowcolor} $291.6603\pm7.065$  \\[4pt]  \midrule
PL ($l=2$) & --- & True & $0.2538\pm0.003$ &  $206.6065\pm2.536$  \\[4pt]  \midrule
PL ($l=4$) & --- & True & $0.2654\pm0.029$ &  $258.8638\pm7.315$  \\[4pt]  \midrule
PL ($l=6$) & --- & True & $0.8696\pm0.170$ &  $331.7095\pm13.724$ \\[4pt]  
\bottomrule
\end{tabular} 
}
\end{center}
\vspace{-0.25cm}
\end{table}

The results in ~\Cref{tab:realworld_baseline_comparisons} show that \methodnames~with the $C_E$ constraint, ReLU activation, and preconditioning achieved the lowest relative $\ell_2$ error ($0.2118$) slightly outperforming the DNN baseline ($0.2132$). 
While the performance gains are modest, they align with the findings in~\Cref{sec:orths_comparison}, where the polynomial layer in \methodnames~enhance expressivity when paired with orthogonality constraints and preconditioning.
While preconditioning increases wall-clock time, it generally reduces errors-- especially with constraints $C_E$ and $C_G$.
Taken together, the results thus far indicate that $C_E$ and $C_G$ can likely serve as general-purpose constraints, though the optimal constraints are likely problem specific.
However, the effectiveness of preconditioning varies, as seen in cases like $C_H$ with tanh, where errors increased. 
Despite these exceptions, the benefits of preconditioning are evident in most configurations, with $C_E$ and ReLU being the most effective combination.
Interestingly, PANN variants remain competitive even without preconditioning, though preconditioning provides additional error reduction, reinforcing the insights from~\Cref{sec:orths_comparison}.
We also note that the wall-clock times exhibit higher variance for some PANN configurations which arises because both preconditioning and the orthogonality penalties alter the curvature of the loss landscape, thereby influencing the line search behavior of the LBFGS optimizer. 
The resulting variation in internal evaluations affects runtime but not accuracy, as the relative errors remain stable across seeds.

%%%%%%%%%%%%%%%%%%%%%%%%%%%%%%%%%%%%%%%%%%%%%%%%%%%%%%%%%%%%%%%%%%%%%%%%%%
\section{Numerical Experiments for PDE Problems}\label{sec:pde-experiments}
In this section, we demonstrate how \methodnames~ extend to the numerical solution of both linear and nonlinear PDEs.
Following the standard PINN formulation~\cite{raissi2018jmlr}, all PDE experiments use the physics-informed loss defined in~\cref{eq:pde-prob-stmt}, which combines the supervised data mismatch loss with the (unsupervised) residual loss at collocation points.
Similar to the regression setting in~\Cref{sec:experiments}, we test each baseline PINN architecture with the PANN framework, which incorporates a trainable polynomial layer, preconditioning, coefficient truncation, and weak orthogonality constraints.

\paragraph{Experimental Outline}
We evaluate the performance of PANNs on a diverse set of benchmark PDEs, chosen to test the method under varying scenarios.
We start with the 2D Poisson problem in~\Cref{sec:pdeexp_poisson} --- a classical linear elliptic PDE. 
This example uses the same high-degree polynomial solution as in ~\Cref{sec:exact_solution_recovery} and serves to validate the method's ability to recover known polynomial solutions in a PDE setting. 
Next in ~\Cref{sec:pdeexp_ac}, we consider the 2D steady-state Allen-Cahn equation (a nonlinear elliptic PDE) and once again compare errors against a manufactured solution. 
This test assesses PANN's performance in capturing solutions with both polynomial and oscillatory components in a nonlinear setting.  
We then move to time-dependent PDEs, starting with the Korteweg-De Vries (KdV) equation in~\Cref{sec:pdeexp_ks}, a third order dispersive PDE with soliton solutions. The KdV equation is challenging even for classical numerical methods due to its localized wave characteristics, and hence serves as an excellent test case for PANNs. 
Then, in~\Cref{sec:pdeexp_ks}, we examine the Kuramoto-Sivashinksy (KS) equation in a chaotic regime. 
This equation introduces strong nonlinearity and instability, testing the model's ability to approximate chaotic spatiotemporal dynamics. 
\Cref{sec:pdeexp_gs} focuses on the Grey-Scott reaction-diffusion equation, a multi-variable PDE that exhibits rich pattern formation and tests the model's ability to scale to coupled nonlinear multi-output systems.
Then, we conclude with a challenging case for PANNs in~\Cref{sec:limitations}, where we examine the Darcy Flow problem with jump coefficients. 

\paragraph{Baseline Methods}
We compare the performance of PANNs against a representative set of PINN baselines, chosen for their demonstrated popularity and effectiveness across diverse PDE benchmarks. 
These baselines include both architectural modifications and training algorithm enhancements.
The architectural variants include the standard fully connected PINN~\cite{raissi2019jcp}, RFF~\cite{wang2021cmame}, which introduces random Fourier feature embeddings used to enrich the input space and enhance expressivity, and ModMLP~\cite{wang2021siam}, which modifies the MLP architecture to improve gradient flow.  
Further, we compare against two baselines that retain the standard PINN architecture but incorporate specialized training techniques to improve convergence. 
GradNorm~\cite{wang2021siam} balances gradient magnitudes from different loss terms to improve multi-objective training, and Causal~\cite{wang2022causality}, which employs causality-based training to improve convergence in time-dependent PDEs. 
As in~\Cref{sec:experiments}, we compare the non-time-dependent examples with the optimized polynomial (PL) baseline, which includes the polynomial expansion in the spatial dimension(s).
To assess the extendibility of our framework, we test each of the above baselines with the PANN components.
We implemented all baselines and their PANN counterparts following the architectural and training details provided in their respective references~\cite{raissi2019jcp, wang2023expert}.

\begin{figure}[!htpb]
\centering
\includegraphics[width=0.9\textwidth]{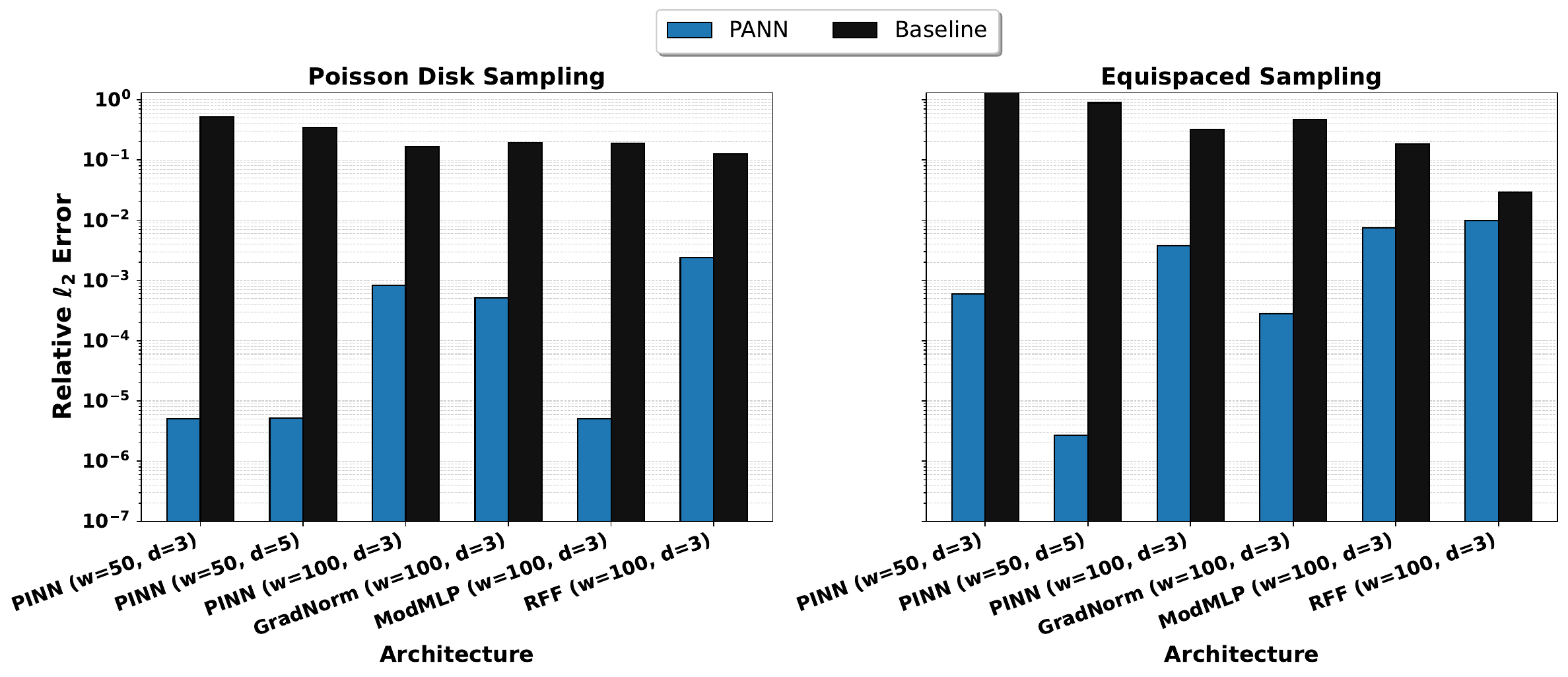}
\caption{\textit{(Poisson Equation)}
Select mean relative $\ell_2$ errors (log scale) for baseline PINN variants (black) and their PANN counterparts (blue), evaluated across multiple architectures: standard PINNs (with varying widths and depths), RFF, GradNorm, and ModMLP.
Each model is labeled with its hidden layer width $w$ and depth $d$ in the format (w=$\cdot$, d=$\cdot$). 
The polynomial layer size in each PANN is controlled by $c=0.003$ use orthogonality constraint $C_E$. 
Results are show for both Poisson (left) and equispaced (right) sampling, each with $N = 4096$ collocation points. Full error statistics are reported in~\Cref{tab:appendix-poisson-errorstd}.
}\label{fig:poisson}
\end{figure}

\paragraph{Experimental Details}
Unless otherwise specified, we adopted a similar experimental design to the one described in~\Cref{sec:exact_solution_recovery}.
We evaluated both standard PINNs and their PANN counterparts using three network configurations: (1) three hidden layers with $50$ nodes, (2) three hidden layers with $100$ nodes, and (3) five hidden layers with $50$ nodes.
For all non-PINN baselines, we use three hidden layers with $100$ nodes.
We denote the PANN variants using the format PANN (c=$\;\cdot$, Base [w=$\;\cdot$, d=$\;\cdot$]), where c=$\;\cdot$ represents the polynomial basis control parameter and Base refers to the corresponding architecture or training method---either PINN, RFF, ModMLP, GradNorm, or Causal---and [w=$\;\cdot$, d=$\;\cdot$] represent the width and depth of the architecture\footnote{In figures, we use abbreviated labels by referencing the baseline architecture and distinguishing between PANN variants through color or context.}.
We trained all models using varying collocation point set sizes,
comparing Poisson disk~\cite{SFKSISC2018} and equispaced points
and used the tanh activation function.
For PANNs, we varied the polynomial degree as a function of $N$ collocation points using predefined scaling rules, and we applied the orthogonality constraint $C_E$.
Specific experimental hyperparameters, along with software and hardware details, are outlined in the \Cref{sec:apx-expdetails}.

\begin{figure}[!ht]
\centering
\includegraphics[width=\textwidth]{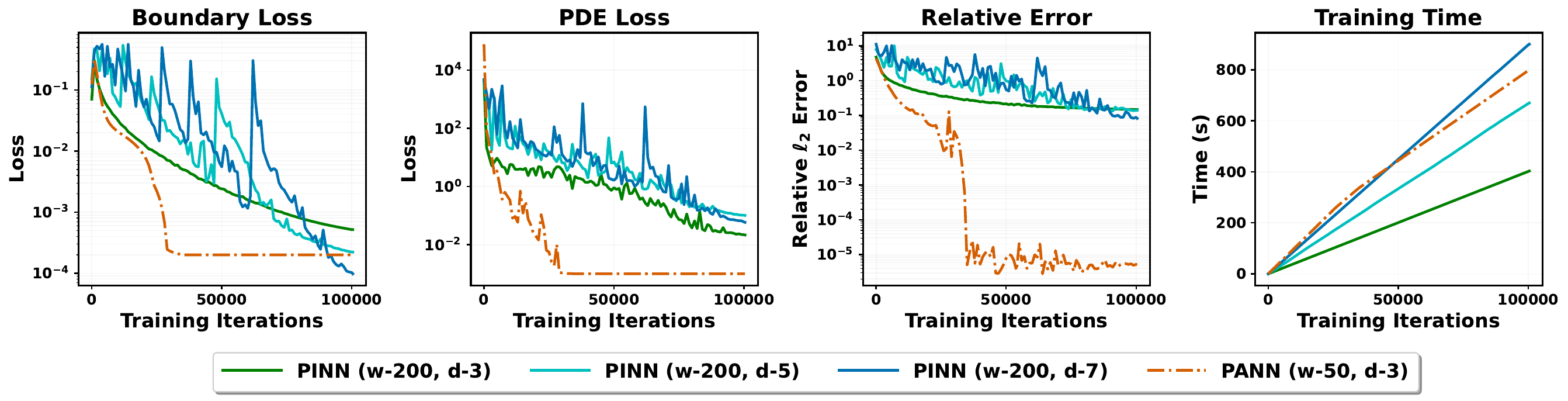}
\vspace{-0.5cm}
\caption{\textit{\small 
(Poisson Equation) Training behavior and computational cost for several PINN architectures compared with a PANN model under similar iteration and runtime budgets.
Left to right: (A) boundary-loss evolution, (B) PDE-loss evolution, (C) relative $\ell_2$ error during training, and (D) cumulative training time in seconds. 
Three PINN models (width=200, depths=3, 5, 7) are compared against a PANN (c=0.003, PINN [w=50, d=3]).
Results are shown for Poisson point sampling with $N=4096$ collocation points. 
}}
\label{fig:poisson-equalcost}
\end{figure}

\subsection{2D Poisson Equation} \label{sec:pdeexp_poisson}
We first consider the linear Poisson equation given by:
\begin{equation*}
\begin{aligned}
    \Delta u(\x) = f(\x), &&\text{ for } \x\in \Omega, \;\;\;\;
    u(\x) = g(\x), &&\text{ for } \x \in \partial\Omega,
\end{aligned}
\end{equation*}
with $\Omega = [-1,1]^2$. 
The ground truth solution is chosen as $u(x) = u(x_1,x_2) = P_{10}(x_1)P_{10}(x_2)$ (the product of degree 10 Legendre polynomials), and $f$, $g$ are derived accordingly.
This test mirrors the exact polynomial recovery task in \Cref{sec:exact_solution_recovery}, now posed as a PDE, and serves as a controlled setting to assess whether PANNs can recover known high-degree polynomial solutions. 
Training details and hyperparameters are listed in ~\Cref{tab:poisson_ac_impl_details}.

\paragraph{Results}
For clarity and brevity,~\Cref{fig:poisson} presents a selected subset of results from the full experimental sweep.
Comprehensive results---including mean errors, standard deviations, and training times---are reported in ~\Cref{tab:appendix-poisson-errorstd}.
~\Cref{fig:poisson} shows that PANNs provide consistent gains in prediction accuracy. 
Notably, the PANN model using the ModMLP architecture achieves near machine-precision accuracy on Poisson-disk points, with a relative $\ell_2$ error of $5.02 \times 10^{-6}$.
Interestingly, narrower PANN models using the standard PINNs achieved slightly lower accuracy than the wider baselines. 
Since the PINN basis is unnecessary for this problem, using narrower networks likely simplifies the task for the orthogonality constraint $C_E$, enabling more effective regularization and truncation. 
As expected, when the optimized polynomial layer (PL) is sufficiently expressive, it recovers the target solution with near-machine precision across both sampling strategies ($4.34 \times 10^{-6}$ for equispaced and $7.16 \times 10^{-6}$ for Poisson disk).
Finally, PANN (c=$0.003$, PINN [w=50, d=5]) with equispaced points achieved the overall lowest relative error of $2.68\times 10^{-6}$.

To complement the accuracy-based comparison in~\Cref{fig:poisson}, we include an additional diagnostic experiment in~\Cref{fig:poisson-equalcost} that examines the training behavior and computational cost of several PINN variants relative to a representative PANN model. We trained PINNs with width 200 and increasing depths 3, 5, and 7 until their total computational budgets were similar to that of the PANN (c = 0.003, PINN [w = 50, d = 3]). 
The purpose of this experiment is to determine whether the benefits of PANNs stem from the polynomial augmentation itself or whether a sufficiently enlarged PINN can achieve comparable performance under equal optimization effort. 
As shown in~\Cref{fig:poisson-equalcost}, deeper PINNs reduce the loss somewhat but do so at higher cost, whereas the PANN---despite using a smaller PINN backbone---achieves substantially faster reductions in both loss and relative $\ell_2$ error while using a (now) similar computational budget. 
This indicates that the performance gains of PANNs arise from the structured polynomial augmentation and orthogonality constraints rather than simply increasing network size or training effort.

\subsection{2D Steady-State Allen-Cahn Equation} \label{sec:pdeexp_ac} 
Next, we consider the nonlinear steady-state Allen-Cahn equation:
\begin{equation*}
\begin{aligned}\label{2dAllenCahn}
    \Delta u + u(u^2-1) &= f(\x) &&\text{ for } \x \in \Omega, \;\;\;\;\;
    u(\x) = g(\x) &&\text{ for } \x \in \partial\Omega,
\end{aligned}
\end{equation*}
with $\Omega = [-1,1]^2$. 
We choose the manufactured solution to be \\ $u(x) = u(x_1,x_2) = x_1^3 x_2^3 + 5x_1 \cos(2\pi x_1) \cos(2\pi x_2)$, and derive $f$ and $g$ accordingly. 
Training details and hyperparameters are listed in ~\Cref{tab:poisson_ac_impl_details}.

\begin{figure}[!htpb]
\centering
\includegraphics[width=0.9\textwidth]{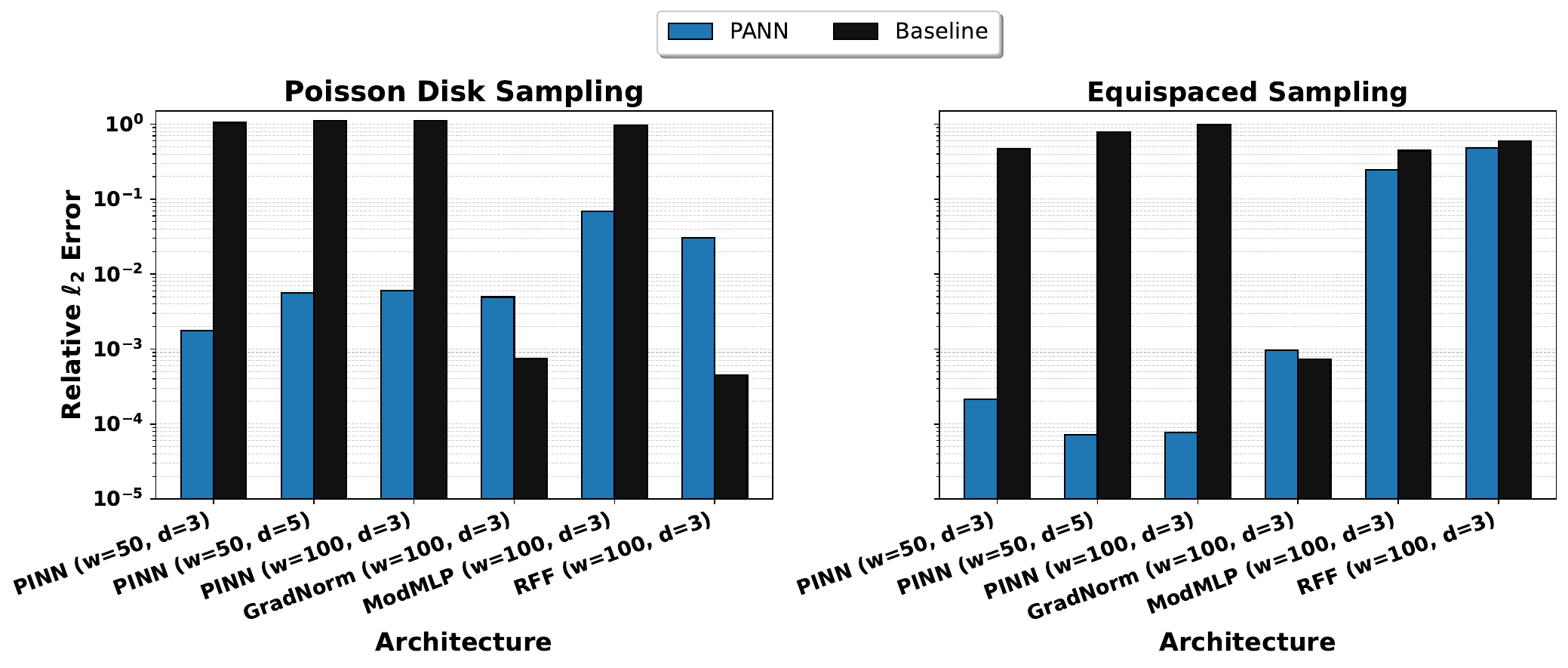}
\caption{\textit{(Allen-Cahn Equation)}
Select mean relative $\ell_2$ errors (log scale) for baseline PINN variants (black) and their PANN counterparts (blue), evaluated across multiple architectures: standard PINNs (with varying widths and depths), RFF, GradNorm, and ModMLP.
Each model is labeled with its hidden layer width $w$ and depth $d$ in the format (w=$\cdot$, d=$\cdot$). 
The polynomial layer size in each PANN is controlled by $c=0.003$ use orthogonality constraint $C_E$. 
Results are show for both Poisson (left) and equispaced (right) sampling, each with $N = 4096$ collocation points. Full error statistics are reported in~\Cref{tab:appendix-allencahn-errorstd}.
}\label{fig:allencahn}
\end{figure}

\paragraph{Results} \Cref{fig:allencahn} presents a representative subset of results from the broader experimental evaluation.
A comprehensive summary of the error statistics across all settings can be found in~\Cref{tab:appendix-allencahn-errorstd}. 
Consistent with the trends discussed in~\Cref{sec:pdeexp_poisson}, PANNs generally improve predictive accuracy across many configurations.  
However, the degree of improvement is more dependent on the baseline method in this setting.  
In particular, PANNs using the standard PINNs show significant accuracy improvement, whereas others like PANNs using ModMLP exhibit more modest improvements.
The PANN (PINN (c=0.003, w=50, d=5)) model with equispaced points achieved the overall lowest relative error of $7.2\times 10^{-5}$.
In a few cases, particularly for GradNorm and RFF using equispaced sampling, PANNs slightly under-perform their baselines. 
This may be due in part, to the orthogonality constraint suppressing beneficial feature interactions, resulting in premature coefficient truncation.

\subsection{Korteweg-De Vries Equation} \label{sec:pdeexp_kdv}
The Korteweg–de Vries (KdV) equation is a third-order nonlinear PDE modeling the propagation of shallow water waves:
\begin{equation*}
\begin{aligned}\label{kdv}
\frac{\partial u}{\partial t} + \nu u\frac{\partial u}{\partial x} + \mu^2 \frac{\partial^3 u}{\partial x^3} = 0, \quad x \in [-1, 1], \quad t \in [0, T].
\end{aligned}
\end{equation*}
We impose periodic boundary conditions and set the initial condition as:
\[
u(x, 0) = \cos(\pi x),
\] 
with $\nu=1$ and $\mu=0.022$ as in \cite{wang2024piratenets, zabusky1965interaction}.
Training details and hyperparameters are listed in ~\Cref{tab:kdv_ks_impl_details}.

\begin{figure}[!htpb]
\centering
\includegraphics[width=0.9\textwidth]{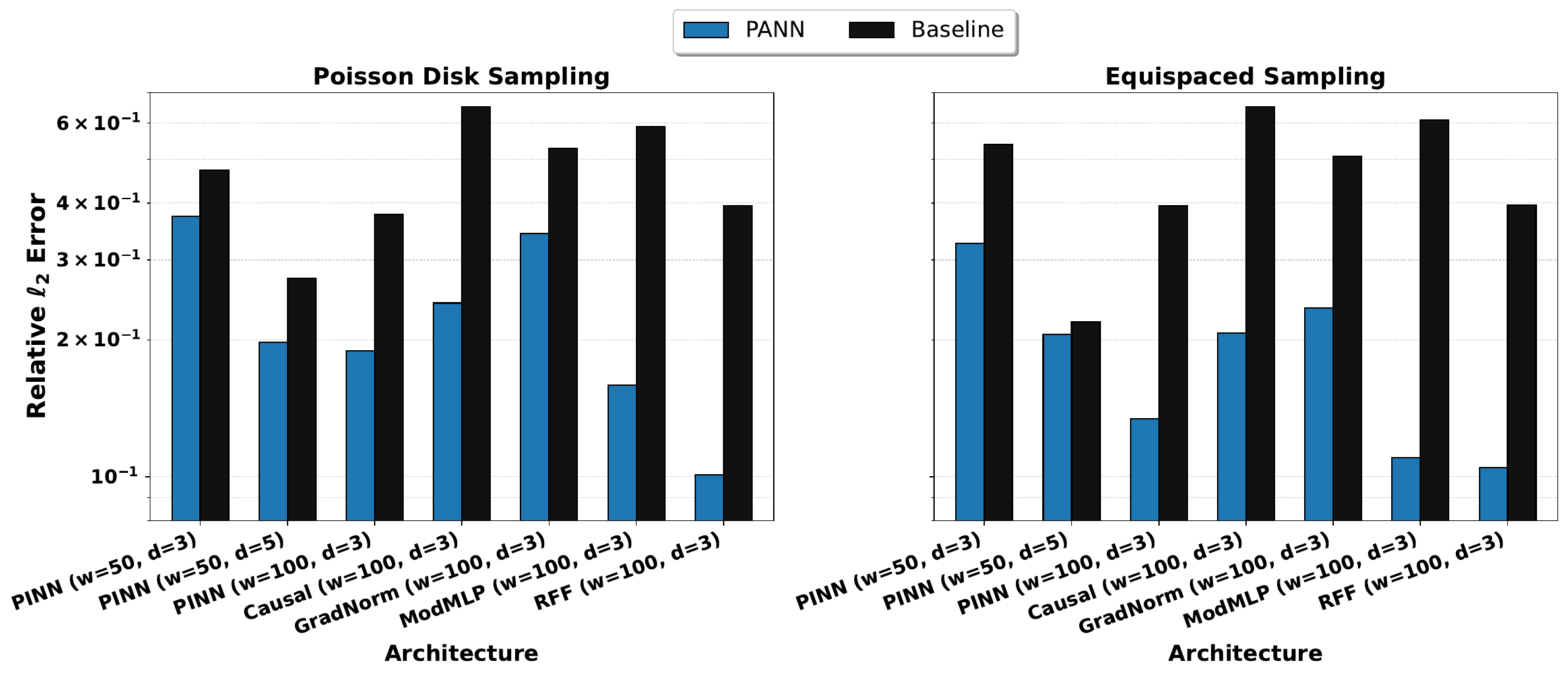}
\caption{\textit{(Korteweg-De Vries Equation)
Select mean relative $\ell_2$ errors (log scale) for baseline PINN variants (black) and their PANN counterparts (blue), evaluated across multiple architectures: standard PINNs (with varying widths and depths), RFF, GradNorm, Causal, and ModMLP.
Each model is labeled with its hidden layer width $w$ and depth $d$ in the format (w=$\cdot$, d=$\cdot$). 
The polynomial layer size in each PANN is controlled by $c=0.0002$ use orthogonality constraint $C_E$. 
Results are show for both Poisson (left) and equispaced (right) sampling, each with $N = 16384$ collocation points. Full error statistics are reported in~\Cref{tab:appendix-kdv-errorstd}.
}}\label{fig:kdv}
\end{figure}

\paragraph{Results} 
~\Cref{fig:kdv} illustrates that, for the Korteweg-De Vries equation, the effectiveness of PANNs is less dependent on the chosen baseline---PANNs provide consistent accuracy improvement across all baselines.
In particular, architectures such as RFF and ModMLP show significant benefits from the polynomial layer.
The PANN model using RFF achieved the overall lowest relative error of $0.100$ and $0.105$ for equispaced and Poisson disk sampled points respectively.
Full statistics including standard deviations and extended results are provided in~\Cref{tab:appendix-kdv-errorstd}.

\subsection{Kuramoto-Sivashinsky Equation} \label{sec:pdeexp_ks}
The Kuramoto–Sivashinsky (KS) equation is a PDE exhibiting chaotic solutions and arises in fluid dynamics and combustion modeling applications:
\begin{equation*}
\begin{aligned}\label{ks}
\frac{\partial u}{\partial t} + \frac{\partial^2 u}{\partial x^2} + \frac{\partial^4 u}{\partial x^4} + u\frac{\partial u}{\partial x} = 0, \quad x \in [0, 2\pi], \quad t \in [0, T].
\end{aligned}
\end{equation*}
We use periodic boundary conditions and an initial condition of the form:
\[
u(x, 0) = \cos\left(\frac{2\pi x}{L}\right),
\]
This benchmark evaluates the capacity of PANNs to approximate nonlinear, unstable dynamics and maintain temporal coherence.
Training details and hyperparameters are listed in ~\Cref{tab:kdv_ks_impl_details}.

\begin{figure}[!htpb]
\centering
\includegraphics[width=0.9\textwidth]{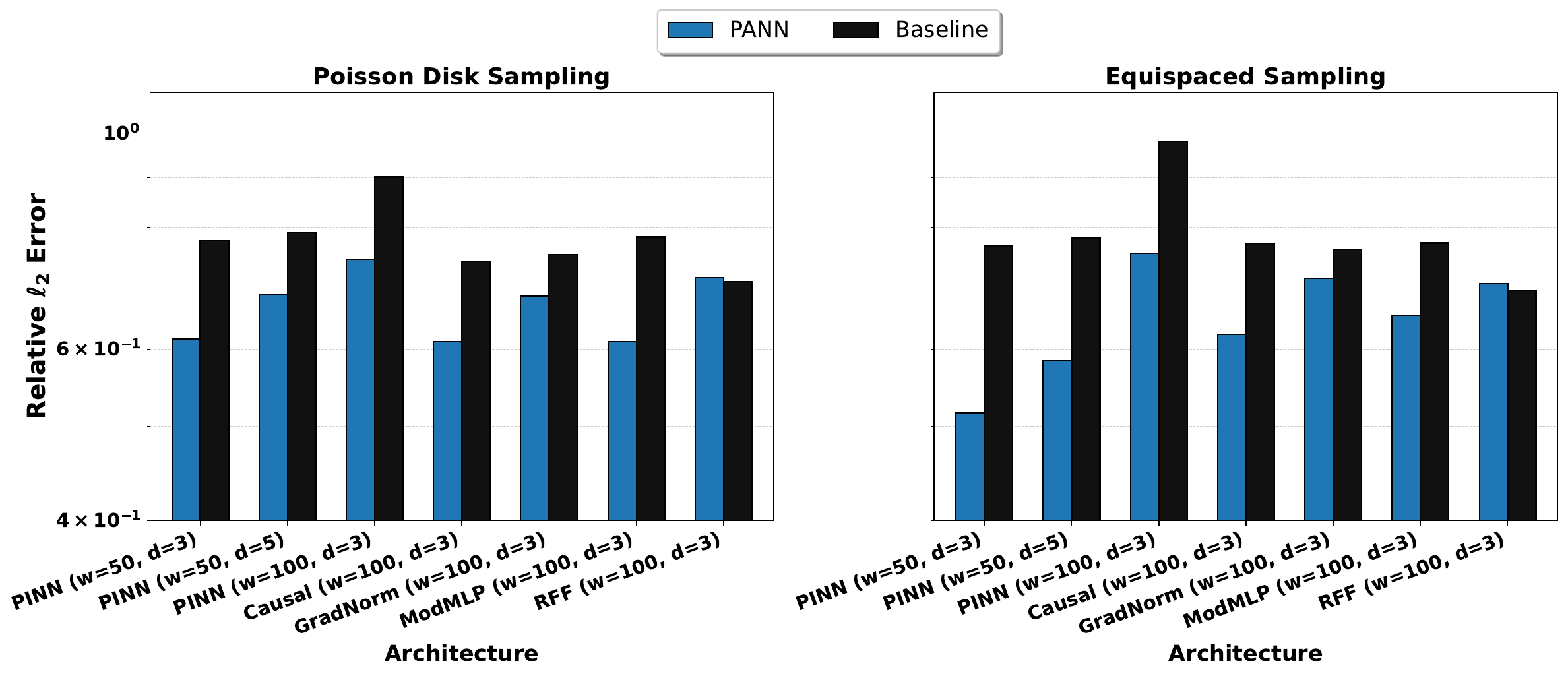}
\caption{\textit{(Kuramoto-Sivashinksy Equation)
Select mean relative $\ell_2$ errors (log scale) for baseline PINN variants (black) and their PANN counterparts (blue), evaluated across multiple architectures: standard PINNs (with varying widths and depths), RFF, GradNorm, Causal, and ModMLP.
Each model is labeled with its hidden layer width $w$ and depth $d$ in the format (w=$\cdot$, d=$\cdot$). 
The polynomial layer size in each PANN is controlled by $c=0.0002$ use orthogonality constraint $C_E$. 
Results are show for both Poisson (left) and equispaced (right) sampling, each with $N = 16384$ collocation points. Full error statistics are reported in~\Cref{tab:appendix-kuramoto-errorstd}.
}}\label{fig:ks}
\end{figure}

\paragraph{Results} \Cref{fig:ks} presents a representative subset of results from the broader experimental evaluation.
Due to the chaotic nature of the system, this benchmark remains difficult for all tested models, however, PANNs still provide consistent accuracy improvements. 
This highlights the limitations of current architectures and indicates that further work is needed to improve modeling chaotic spatiotemporal PDEs.
The PANN (PINN [c=0.0001, w=50, d=3]) model achieved the overall lowest relative error of $0.57$ for equispaced sampled points.
Full statistics including standard deviations and extended results are provided in~\Cref{tab:appendix-kuramoto-errorstd}.

\subsection{Grey-Scott Equation} \label{sec:pdeexp_gs}
Finally, we consider the Grey–Scott equation which describes the interaction of two chemical species undergoing diffusion and nonlinear reaction:
\begin{equation*}
\begin{aligned}\label{gs}
\frac{\partial u}{\partial t} &= \epsilon_1 \nabla^2 u + b_1(1 - u) - c_1 uv^2,  \\
\frac{\partial v}{\partial t} &= \epsilon_2 \nabla^2 v - b_2 v + c_2 uv^2,
\quad x = (x_1, x_2) \in \Omega, \quad t \in [0, T],
\end{aligned}
\end{equation*}
where $u$ and $v$ are concentration fields.
Following~\cite{wang2024piratenets}, we set the system parameters as:
\[
\epsilon_1 = 0.2, \quad \epsilon_2 = 0.1, \quad b_1 = 40, \quad b_2 = 100, \quad c_1 = c_2 = 1000.
\]
We define the spatial domain as $\Omega = [-1,1]^2$ and apply periodic boundary conditions.
The initial conditions consist of:
\begin{align}
    u_0(x_1, x_2) &= 1 - \exp\left(-10\left((x_1 + 0.05)^2 + (x_2 + 0.02)^2\right)\right), \tag{5.8} \\
    v_0(x_1, x_2) &= 1 - \exp\left(-10\left((x_1 - 0.05)^2 + (x_2 - 0.02)^2\right)\right), \tag{5.9}
\end{align}
where $u$ and $v$ represent the concentrations of the two species. 
This example evaluates PANN’s ability to approximate coupled, multi-output spatiotemporal systems that exhibit nonlinear instabilities. 
Training details and hyperparameters are listed in ~\Cref{tab:gs_impl_details}.

\begin{figure}[!htpb]
\centering
\includegraphics[width=0.9\textwidth]{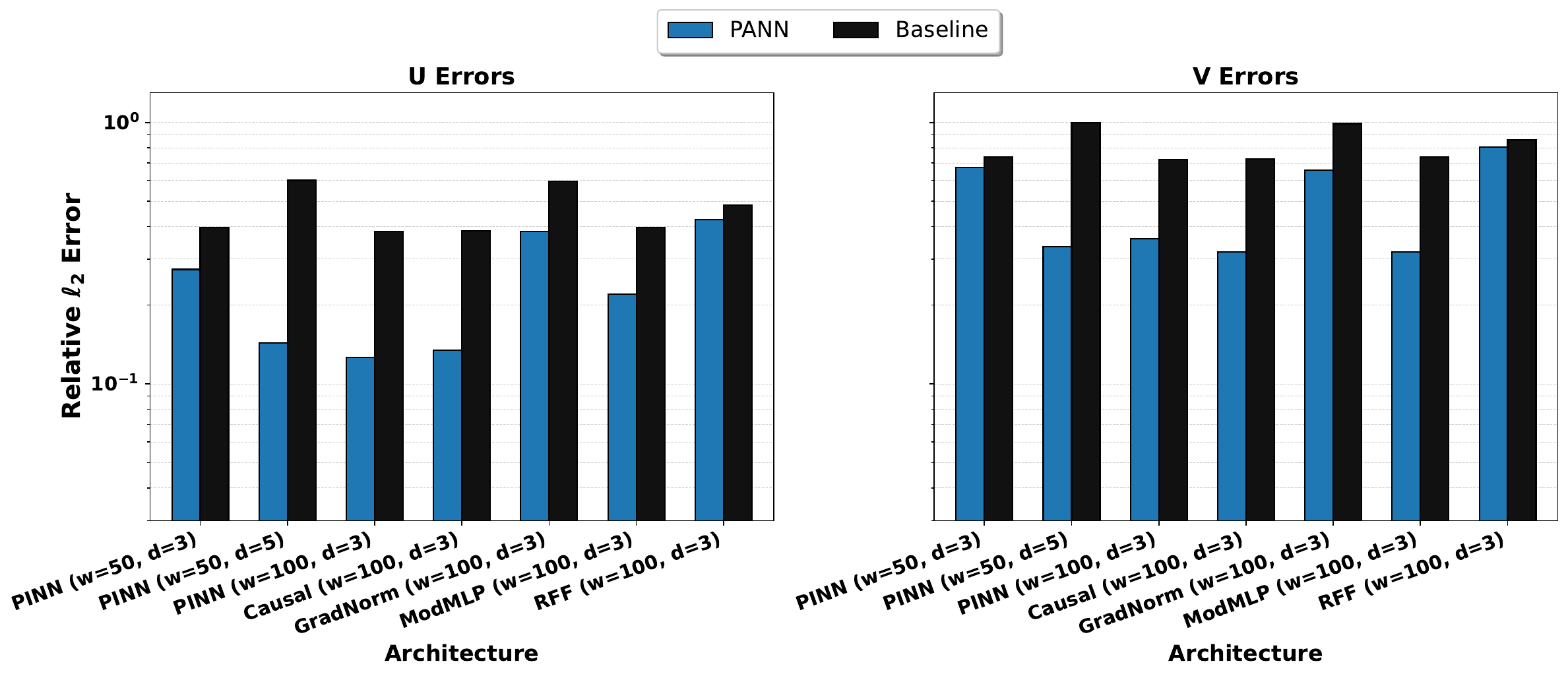}
\caption{\textit{(Grey-Scott Equation)
Select mean relative $\ell_2$ errors (log scale) for baseline PINN variants (black) and their PANN counterparts (blue), evaluated across multiple architectures: standard PINNs (with varying widths and depths), RFF, GradNorm, Causal, and ModMLP.
Each model is labeled with its hidden layer width $w$ and depth $d$ in the format (w=$\cdot$, d=$\cdot$). 
The polynomial layer size in each PANN is controlled by $c=0.001$ use orthogonality constraint $C_E$. 
Results are show for both Poisson (left) and equispaced (right) sampling, each with $N = 131072$ collocation points. Full error statistics are reported in~\Cref{tab:appendix-greyscott-errorstd}.
}}\label{fig:gs}
\end{figure}

\paragraph{Results} \Cref{fig:gs} displays a representative subset of results from the broader experimental evaluation. 
Overall, the effectiveness of PANNs is partially determined by the choice of underlying baseline architecture and learning routines. 
The PANN (PINN [c=0.001, w=50, d=5]) model achieved the overall lowest relative error of $0.143$ for $u$ and $0.306$ for $v$.
Full statistics including standard deviations and extended results are provided in~\Cref{tab:appendix-greyscott-errorstd}.

\subsection{Additional Tests and Limitations} \label{sec:limitations}
To evaluate the behavior of PANNs on PDEs whose solutions possess limited smoothness, we conducted an additional experiment on the well-studied 2D Darcy-flow ``five-strip" manufactured problem~\cite{trask2017fivestrip, nakshatrala2006stabilized, masud2002stabilized, cooley2025hyrespinns}. 
In this benchmark, the domain is partitioned into five vertical subregions, each assigned a distinct permeability value. 
The resulting coefficient jumps induce sharp variations in the pressure field and lower its smoothness across strip interfaces.
We used the same training procedure as in the previous experiments and compared a standard PINN (width = 100, depth = 3) with its PANN counterpart (c = 0.0001, PINN [w = 100, d = 3]). 
In this setting, the standard PINN achieved a relative error of $3.2e-5$ while the PANN---using the default orthogonality strength $\lambda_c=0.001$ with constraint $C_E$---produced a higher error of $3.0e-4$. Theorizing that this higher error was due to the influence of the high-degree polynomial bases (which would produce spurious oscillations around solution jumps), we reduced the influence of the polynomial on the PANN by reducing the orthogonality penalty; recall that this penalty can be viewed as enforcing a form of weak polynomial reproduction on the PANN.
Reducing the orthogonality penalty to $\lambda_c=0.00001$ improved the PANN error to $3.8e-5$, which is comparable to, though not better than, the baseline PINN.
These results indicate that when the PDE solution exhibits discontinuities, a strongly enforced orthogonality constraint (and the corresponding stricter polynomial reproduction) can restrict the flexibility of the DNN component, potentially over-regularizing the DNN bases.
This experiment suggests that future work should explore adaptive or localized polynomial bases or smoothness-driven constraint tuning strategies to better accommodate problems with non-smooth solutions.
\section{Conclusion and Future Work}\label{sec:conclusion}
This paper proposes an effective and applicable method of augmenting neural networks with a trainable polynomial layer.
Additionally, we provide a suite of novel discrete orthogonality constraints enforced through the loss function during optimization. 
Through a suite of numerical experiments, we show that---although simple---our methods result in higher accuracy across a broad range of test problems and apply to many domains, such as predicting solutions to PDEs. 
The experiments show that while our methods increase accuracy, including the polynomial preconditioning increases training times. 
Investigating efficient polynomial preconditioners for polynomials used in neural network architectures would be an interesting future research direction. 
% Additionally, investigating PI-\methodnames~ to solve space-time PDEs is left for future work.

\appendix
\section{Notation} \label{sec:apx-notation}
\begin{table}[h!]
\centering
\small
\caption{Summary of notation used throughout the paper.}
\renewcommand{\arraystretch}{1.0}
\begin{tabular}{ll}
\hline
\textbf{Symbol} & \textbf{Description} \\
\hline
$d$                           & Input dimension of the problem. \\
$N_{\text{data}}$             & Number of supervised training points. \\
$N_{\text{PDE}}$              & Number of PDE collocation points used for residual enforcement. \\
$x_i$, $x_i^b$, $x_j^r$       & Indexed training point, boundary point, and residual/collocation point. \\
$u(x)$                        & Ground-truth target function or PDE solution. \\
$u_\theta(x)$                 & Model prediction parameterized by weights $\theta$. \\
$\theta$                      & All model parameters (DNN and polynomial). \\
\hline
$w$                           & Width of the last hidden layer of the DNN. \\
$m$                           & Number of polynomial basis functions (“polynomial width”). \\
$a_j$                         & Coefficients for the DNN basis functions. \\
$b_k$                         & Coefficients for the polynomial basis functions. \\
$\psi_j(x)$                   & $j$th DNN basis function (activation after hidden layers). \\
$\phi_k(x)$                   & $k$th polynomial basis function (Legendre, TP/TD/HC). \\
\hline
$a$                         & DNN basis coefficient vector. \\
$b$                         & Polynomial basis coefficient vector. \\
$\Psi \in \mathbb{R}^{N \times w}$ 
                              & DNN basis matrix, $\Psi_{i,j} = \psi_j(x_i)$. \\
$\Phi \in \mathbb{R}^{N \times m}$ 
                              & Polynomial basis matrix, $\Phi_{i,k} = \phi_k(x_i)$. \\
$f \in \mathbb{R}^{N}$        & Vector of training outputs $f_i = u(x_i)$. \\
$K \in \mathbb{R}^{N \times N}$  
                              & Diagonal preconditioning matrix, $K_{ii}$ given by Eq.~\eqref{eq:poly-precond-const}. \\
$\lambda$                     & Weight balancing data loss vs.\ PDE residual loss. \\
$\lambda_r$                   & Regularization/Lagrange multiplier in preconditioned objective. \\
$\lambda_c$                   & Orthogonality constraint strength. \\
\hline
$\mathcal{F}[u_\theta]$       & Differential operator applied to model prediction. \\
$\Omega$, $\partial\Omega$    & Domain and boundary of the PDE problem. \\
\hline
\end{tabular} \label{tab:notation}
\end{table}

\section{Experimental Details} \label{sec:apx-expdetails}
The experiments were conducted on a GeForce RTX 3090 GPU with CUDA version 12.3, running on Ubuntu 20.04.6 LTS. 
We used total degree Legendre polynomial bases in all relevant models. 
The orthogonality strength ($\lambda_c$) in PANNs is $0.001$, and the basis coefficient truncation threshold is $0.0001$. 
For the regression experiments, we optimized each model using $20,000$ Adam~\cite{kingma2015iclr} iterations with an initial learning rate of $0.001$ and $400$ LBFGS iterations with an initial learning rate set to $1.0$.
We also employed the cosine learning rate annealing method~\cite{loshchilov2016sgdr}.
We used the Gauss-Legendre quadrature for all $L^2$ projection experiments.
We used a similar optimization approach for the PDE experiments as in~\cite{wang2023expert},
and provide problem specific details in Tables~\ref{tab:poisson_ac_impl_details}, \ref{tab:kdv_ks_impl_details}, and~\ref{tab:gs_impl_details}. 
Experiments comparing PANNs (c=$\;\cdot$, Base [w=$\;\cdot$, d=$\;\cdot$]) to their corresponding Base architectures, use identical hyperparameters for a fair comparison. 
Specifically, all Random Fourier Feature (RFF) models---including both baseline RFF-PINNs and PANNs employing RFF as their base (PANN (c=$\cdot$, RFF [w=$\cdot$, d=$\cdot$]))---use the same embedding configuration. 
Each RFF layer employs 100 Fourier features with frequencies sampled from a standard normal distribution and scaled by a factor of $1.0$. 
This scaling was fixed across all experiments and applied identically to both baseline and augmented variants.
This ensures that any observed performance differences arise solely from the inclusion of the polynomial layer rather than from differences in RFF tuning.
For coupled systems such as the Grey-Scott problem, we employed independent polynomial bases for each output field (\(u, v\)), while sharing the same DNN backbone. 

\begin{table}[!htpb]
\centering
{\small
\caption{\textit{Poisson and Allen-Cahn example problems implementation details}}
\begin{tabular}{|l|c|c|}
\hline
Implementation Details & Base Net & Base Net-PL \\ \hline
Sampling method & \multicolumn{2}{c|}{Equispaced and Poisson disk} \\ \hline 
Training collocation point sample sizes & \multicolumn{2}{c|}{$64, 256, 1024, 4096$} \\ \hline 
Training boundary point sample sizes & \multicolumn{2}{c|}{$400$} \\ \hline 
$\#$ of hidden layers     & \multicolumn{2}{c|}{$3,5$}\\ \hline
$\#$ of neurons per layer & \multicolumn{2}{c|}{$100, 50$} \\ \hline
$\#$ of polynomial basis ($m$) & --- & See Table~\ref{tab:poisson_ac_totaldegrees}  \\ \hline
Orthogonality type & --- & $C_E$ \\ \hline 
Preconditioning & --- & True \\ \hline 
Basis coeff. truncation threshold & --- & $0.0001$ \\ \hline 
Activation functions & \multicolumn{2}{c|}{Tanh} \\ \hline 
Training order & \multicolumn{2}{c|}{Adam then LBFGS} \\ \hline 
$\#$ Adam Iterations & \multicolumn{2}{c|}{100,000} \\ \hline 
\end{tabular} \label{tab:poisson_ac_impl_details}
}
\end{table}

\begin{table}[!htpb]
\centering
{\small
\caption{\textit{KDV and KS example problems implementation details}}
\begin{tabular}{|l|c|c|}
\hline
Implementation Details & Base Net & Base Net-PL \\ \hline
Sampling method & \multicolumn{2}{c|}{Equispaced and Poisson disk} \\ \hline 
Training collocation point sample sizes & \multicolumn{2}{c|}{$256, 1024, 4096, 16384$} \\ \hline 
Training boundary point sample sizes & \multicolumn{2}{c|}{$400$} \\ \hline 
$\#$ of hidden layers     & \multicolumn{2}{c|}{$3,5$}\\ \hline
$\#$ of neurons per layer & \multicolumn{2}{c|}{$100, 50$} \\ \hline
$\#$ of polynomial basis ($m$) & --- & See Table~\ref{tab:kdv_ks_totaldegrees}  \\ \hline
Orthogonality type & --- & $C_E$ \\ \hline 
Preconditioning & --- & True \\ \hline 
Basis coeff. truncation threshold & --- & $0.0001$ \\ \hline 
Activation functions & \multicolumn{2}{c|}{Tanh} \\ \hline 
Training order & \multicolumn{2}{c|}{Adam then LBFGS} \\ \hline 
$\#$ Adam Iterations & \multicolumn{2}{c|}{100,000} \\ \hline 
\end{tabular} \label{tab:kdv_ks_impl_details}
}
\end{table}

\begin{table}[!htpb]
\centering
{\small
\caption{\textit{Grey-Scott example problems implementation details}}
\begin{tabular}{|l|c|c|}
\hline
Implementation Details & Base Net & Base Net-PL \\ \hline
Sampling method & \multicolumn{2}{c|}{Poisson Disk} \\ \hline 
Training collocation point sample sizes & \multicolumn{2}{c|}{$8192, 32768, 131072$} \\ \hline 
Training boundary point sample sizes & \multicolumn{2}{c|}{$400$} \\ \hline 
$\#$ of hidden layers     & \multicolumn{2}{c|}{$3,5$}\\ \hline
$\#$ of neurons per layer & \multicolumn{2}{c|}{$100, 50$} \\ \hline
$\#$ of polynomial basis ($m$) & --- & See Table~\ref{tab:gs_exact_totaldegrees}  \\ \hline
Orthogonality type & --- & $C_E$ \\ \hline 
Preconditioning & --- & True \\ \hline 
Basis coeff. truncation threshold & --- & $0.0001$ \\ \hline 
Activation functions & \multicolumn{2}{c|}{Tanh} \\ \hline 
Training order & \multicolumn{2}{c|}{Adam then LBFGS} \\ \hline 
$\#$ Adam Iterations & \multicolumn{2}{c|}{100,000} \\ \hline 
\end{tabular} \label{tab:gs_impl_details}
}
\end{table}

\begin{table}[!htpb]
\small
\caption{\textit{Suite of total degrees $\ell$ and corresponding count of polynomial terms $m$ in polynomial bases used in the exact solution recovery ($u(x)=P_{10}(x_1)P_{10}(x_2)$) example. 
}}\label{tab:exact_totaldegrees}
\begin{center}
\resizebox{0.9\textwidth}{!}{%
\begin{tabular}{|l|c|c|c|c||c|c|c|c||c|c|c|c|} \hline
    & \multicolumn{4}{c||}{$\mathbf{c=0.001}$} 
    & \multicolumn{4}{c||}{$\mathbf{c=0.002}$} 
    & \multicolumn{4}{c|}{$\mathbf{c=0.003}$}   \\ \hline
$\mathbf{N}$ & $256$ & $1024$ & $4096$ & $16384$ 
    & $256$ & $1024$ & $4096$ & $16384$ 
    & $256$ & $1024$ & $4096$ & $16384$ \\
$\mathbf{\ell}$ & $18$ & $20$ & $26$ & $50$ 
    & $18$ & $22$ & $34$ & $82$ 
    & $18$ & $24$ & $42$ & $116$ \\ 
$\mathbf{m}$ & $190$ & $231$ & $378$ & $1326$ 
    & $190$ & $276$ & $630$ & $3486$ 
    & $190$ & $325$ & $946$ & $6903$ \\ 
\hline
\end{tabular} 
}
\end{center}
\end{table}

\begin{table}[!htpb]
\small
\caption{\textit{Suite of total degrees $\ell$ and corresponding count of polynomial terms $m$ in polynomial bases used in the non-smooth function  
($u(x)=x_1^2x_2^2\sin(1/x_1)\sin(1/x_2)$) 
approximation example.}}\label{tab:diff_totaldegrees}
\begin{center}
\resizebox{0.9\textwidth}{!}{%
\begin{tabular}{|l|c|c|c|c||c|c|c|c||c|c|c|c|} \hline
    & \multicolumn{4}{c||}{$\mathbf{c=0.001}$} 
    & \multicolumn{4}{c||}{$\mathbf{c=0.002}$}
    & \multicolumn{4}{c|}{$\mathbf{c=0.003}$} \\ \hline
$\mathbf{N}$ & $256$ & $1024$ & $4096$ & $16384$ 
    & $256$ & $1024$ & $4096$ & $16384$ 
    & $256$ & $1024$ & $4096$ & $16384$ \\
$\mathbf{\ell}$ & $9$ & $10$ & $13$ & $25$ 
    & $9$ & $11$ & $17$ & $41$ 
    & $9$ & $12$ & $21$ & $58$ \\ 
$\mathbf{m}$ & $55$ & $66$ & $105$ & $351$ 
    & $55$ & $78$ & $171$ & $903$ 
    & $55$ & $91$ & $253$ & $1770$ \\ 
\hline
\end{tabular} 
}
\end{center}
\end{table}

\begin{table}[!htpb]
\small
\caption{\textit{Suite of total degrees $\ell$ and corresponding count of polynomial terms $m$ in polynomial bases used in the Poisson and Allen-Cahn PDE examples.}}\label{tab:poisson_ac_totaldegrees}
\begin{center}
\resizebox{0.6\textwidth}{!}{%
\begin{tabular}{|l|c|c|c|c||c|c|c|c|} \hline
   & \multicolumn{4}{c||}{$\mathbf{c=0.003}$} 
   & \multicolumn{4}{c|}{$\mathbf{c=0.004}$} \\ \hline
$\mathbf{N}$    & 64 & 256 & 1024 & 4096 &  64 & 256 & 1024 & 4096  \\  
$\mathbf{\ell}$ & 9 & 9 & 12 & 21        &  9  & 10 & 13 & 25  \\ 
$\mathbf{m}$    & 55 & 55 & 91 & 253     &  55 & 66 & 105 & 351  \\  
\hline
\end{tabular}
}
\end{center}
\end{table}

\begin{table}[!htpb]
\small
\caption{\textit{Suite of total degrees $\ell$ and corresponding count of polynomial terms $m$ in polynomial bases used in the Korteweg De-Vries and Kuramoto Sivashinsky PDE examples.}}\label{tab:kdv_ks_totaldegrees}
\begin{center}
\resizebox{0.6\textwidth}{!}{%
\begin{tabular}{|l|c|c|c|c||c|c|c|c|} \hline
   & \multicolumn{4}{c||}{$\mathbf{c=0.0002}$} 
   & \multicolumn{4}{c|}{$\mathbf{c=0.0004}$} \\ \hline
$\mathbf{N}$    & 256 & 1024 & 4096 & 16384 &  256 & 1024 & 4096 & 16384 \\  
$\mathbf{\ell}$ & 9 & 9 & 9 & 12        &  9  & 9 & 10 & 15  \\ 
$\mathbf{m}$    & 10 & 10 & 10 & 13     &  10 & 10 & 11 & 16  \\  
\hline
\end{tabular}
}
\end{center}
\end{table}

\begin{table}[!htpb]
\small
\caption{\textit{Suite of total degrees $\ell$ and corresponding count of polynomial terms $m$ in polynomial bases used in the Grey-Scott example. 
}}\label{tab:gs_exact_totaldegrees}
\begin{center}
\resizebox{0.6\textwidth}{!}{%
\begin{tabular}{|l|c|c|c||c|c|c||c|c|c|} \hline
    & \multicolumn{3}{c|}{$\mathbf{c=0.001}$} 
    & \multicolumn{3}{c|}{$\mathbf{c=0.002}$} \\ \hline
$\mathbf{N}$ & $8192$ & $32768$ & $131072$ 
             & $8192$ & $32768$ & $131072$  \\
$\mathbf{\ell}$ & $9$ & $10$ & $13$ 
                & $9$ & $11$ & $17$   \\ 
$\mathbf{m}$ & $55$ & $66$ & $105$
             & $55$ & $78$ & $171$  \\ 
\hline
\end{tabular} 
}
\end{center}
\end{table}

\section{Additional Numerical Results} \label{sec:apx-additresults}
We compared our results against a variety of standard regression models to benchmark performance. 
These models include:
AdaBoost, Bagging, Bayesian Ridge, Elastic Net, Gradient boosting, Huber regression, linear SVR, MLP, Nu SVR, SVR, and kNeighbors regression.
Tables~\ref{tab:test2_baseline_comparisons}, \ref{tab:test4_baseline_comparisons}, \ref{tab:highdim_baseline_comparisons}, and~\ref{tab:realworld_baseline_comparisons_full} list the models whose relative errors fall below at least $0.6$ for each example. 

\begin{table}[!htpb]
\small
\caption{\textit{Relative $\ell_2$ errors and standard deviations of baseline methods for predicting the two-dimensional Legendre function problem in~\Cref{sec:exact_solution_recovery}}}\label{tab:test2_baseline_comparisons}
\begin{center}
\resizebox{0.9\textwidth}{!}{%
\begin{tabular}{|l|c|c|c|c|c|}
 \hline
 Model & 64 & 256 & 1024 & 4096 & 16384 \\ \hline
 Bagging & $0.916\pm0.076$ & $0.854\pm0.014$ & $0.881\pm0.021$ & $0.734\pm0.020$ & $0.433\pm0.017$\\
 kNeighbors & $1.064\pm0.000$ & $0.972\pm0.000$ & $0.876\pm0.000$ & $0.619\pm0.000$ & $0.333\pm0.000$\\
 \hline
\end{tabular}
}
\end{center}
\end{table}

\begin{table}[!htpb]
\small
\caption{\textit{Relative $\ell_2$ errors and standard deviations of baseline methods for predicting the non-smooth synthetic function in~\Cref{sec:non_smooth_recovery}.}} \label{tab:test4_baseline_comparisons}
\begin{center}
\resizebox{0.75\textwidth}{!}{%
 \begin{tabular}{|l|c|c|c|c|c|}
 \hline
 Model & 64 & 256 & 1024 & 4096 & 16384 \\ \hline
 Bagging & $1.190\pm0.220$ & $0.462\pm0.084$ & $0.180\pm0.022$ & $0.073\pm0.003$ & $0.033\pm0.006$\\
 Grad Boost & $0.846\pm0.010$ & $0.572\pm0.003$ & $0.240\pm0.001$ & $0.184\pm0.000$ & $0.217\pm0.000$\\
 MLP & $0.778\pm0.068$ & $0.864\pm0.191$ & $0.528\pm0.164$ & $0.359\pm0.077$ & $0.313\pm0.042$\\
 Nu SVR & $0.418\pm0.000$ & $0.151\pm0.000$ & $0.077\pm0.000$ & $0.069\pm0.000$ & $0.060\pm0.000$\\
 SVR & $0.690\pm0.000$ & $0.346\pm0.000$ & $0.217\pm0.000$ & $0.190\pm0.000$ & $0.170\pm0.000$\\
 kNeighbors & $0.776\pm0.000$ & $0.189\pm0.000$ & $0.081\pm0.000$ & $0.041\pm0.000$ & $0.018\pm0.000$\\
 \hline
 \end{tabular} 
 }
\end{center}
\end{table}

\begin{table}[!htpb]
\small
\caption{\textit{Relative $\ell_2$ errors and standard deviations of baseline methods for increasing dimensions from~\Cref{sec:highdim-results}.}} \label{tab:highdim_baseline_comparisons}
\begin{center}
\resizebox{0.75\textwidth}{!}{%
 \begin{tabular}{|l|c|c|c|}
 \hline
 Model & 2 & 3 & 4 \\ \hline
 Bagging & $0.122\pm0.004$ & --- & $0.750\pm0.265$ \\
 Bayesian Ridge & $1.005\pm0.010$ & $0.237\pm0.329$ & ---\\
 Elastic Net & $1.000\pm0.000$ & $0.550\pm0.606$ & ---\\
 kNeighbors & $0.175\pm0.000$ & --- & $0.479\pm0.035$ \\
 \hline
 \end{tabular} 
 }
\end{center}
\end{table}

\begin{table}[!htpb]
\small
\caption{\textit{Relative $\ell_2$ errors and standard deviations of baseline methods for the real-world dataset from~\Cref{sec:realworld-results}.}} \label{tab:realworld_baseline_comparisons_full}
\begin{center}
\resizebox{0.75\textwidth}{!}{%
\begin{tabular}{|l|c||l|c|}
 \hline
 Model & Relative $\ell_2$ Error & Model & Relative $\ell_2$ Error \\ \hline
 Adaboost & $0.368\pm0.016$ & MLP & $0.474\pm0.084$ \\
 Bagging & \textbf{ $0.225\pm0.003$} & Nu SVR & $0.263\pm0.003$\\
 Elastic Net & $0.487\pm0.002$ & SVR & $0.264\pm0.003$\\
 Grad Boosting & \textbf{$0.225\pm0.004$} & Theil Sen & $0.511\pm0.263$\\
 Huber & $0.342\pm0.012$ & Tweedie & $0.320\pm0.011$\\
 Linear SVR & $0.287\pm0.005$ & kNeighbors & $0.255\pm0.001$\\
 \hline
\end{tabular} 
}
\end{center}
\end{table}

\begin{table*}[h!]
\centering
\caption{
\textbf{(2D Poisson)} Relative $\ell_2$ errors (mean and standard deviation in parentheses) and average runtime (s) for all PINN variations for $N=4096$.}
\label{tab:appendix-poisson-errorstd}
\scriptsize
\renewcommand{\arraystretch}{1.2}
\setlength{\tabcolsep}{4pt}
\begin{tabular}{@{}lcccc|c@{}}
\toprule
\textbf{Equispaced Points} & \multicolumn{4}{c|}{\textbf{Relative $\mathbf{\ell_2}$ Errors}} & \textbf{Runtime (s)} \\
\cmidrule(lr){2-5}
& $N = 64$ & $N = 256$ & $N = 1024$ & $N = 4096$ & \\ 
\midrule

\midrule
\multicolumn{6}{l}{\textbf{PL (c=$\;\cdot$)}} \\
c-0.003      & 82.447 (10.33) & 15.563 (0.04) & 4.548 (0.27) & 4.34e-06 (3.15e-06) &  8.00e+01 (4.50e+01) \\
c-0.004      & 82.447 (10.33) & 11.691 (0.01) & 4.167 (0.00) & 5.01e-06 (3.88e-06) &  8.00e+01 (4.50e+01) \\

\midrule
\multicolumn{6}{l}{\textbf{PINN (w=$\;\cdot$, d=$\;\cdot$)}} \\
w-100, d-3  & 5.712 (0.99)  & 9.778 (0.04)  & 2.256 (1.33) & 3.19e-01 (1.46e-01) &  2.50e+02 (2.10e+02) \\
w-50, d-3   & 9.244 (3.42)  & 10.886 (2.92) & 3.167 (0.41) & 2.56e+00 (3.70e-01) &  2.50e+02 (2.10e+02) \\
w-50, d-5   & 16.685 (4.36) & 4.893 (0.76)  & 3.653 (0.32) & 8.90e-01 (1.50e-01) &  4.00e+02 (2.80e+02) \\

\multicolumn{6}{l}{\textbf{PANN (PINN [c=$\;\cdot$, w=$\;\cdot$, d=$\;\cdot$])}} \\
c-0.003, w-100, d-3 & 12.479 (4.59)  & 7.275 (2.86) & 1.107 (0.35) & 3.74e-03 (5.52e-03) &  5.20e+02 (2.00e+02) \\
c-0.003, w-50, d-3  & 41.162 (41.64) & 23.98 (4.33) & 3.415 (2.97) & 6.00e-04 (1.34e-03) &  5.00e+02 (2.20e+02) \\
c-0.003, w-50, d-5  & 39.889 (34.82) & 5.330 (1.90) & 2.298 (1.67) & \textbf{2.69e-06 (1.37e-07)} &  6.30e+02 (2.40e+02) \\
c-0.004, w-100, d-3 & 12.479 (4.59)  & 7.821 (1.56) & 1.192 (0.58) & 3.33e-02 (1.09e-02) &  5.50e+02 (2.20e+02) \\
c-0.004, w-50, d-3  & 41.162 (42.64) & 20.88 (4.36) & 3.115 (1.88) & 2.68e-02 (1.25e-02) &  5.50e+02 (2.20e+02) \\
c-0.004, w-50, d-5  & 39.889 (34.82) & 8.308 (3.14) & 1.316 (0.60) & 2.18e-02 (1.86e-02) &  7.60e+02 (3.30e+02) \\

\midrule
\multicolumn{6}{l}{\textbf{Baselines (w=100, d=3)}} \\
RFF                   & 25.634 (24.98) & 5.474 (2.20)  & 0.447 (0.17) & 2.88e-02 (4.60e-03) &  2.00e+02 (1.40e+00) \\
GradNorm              & 15.156 (8.99)  & 11.460 (4.94) & 3.206 (1.88) & 4.70e-01 (3.72e-01) &  1.10e+02 (2.40e+00) \\
ModMLP                & 38.012 (27.17) & 9.996 (9.71)  & 1.052 (0.15) & 1.85e-01 (6.28e-02) &  6.00e+02 (9.70e-02) \\

\multicolumn{6}{l}{\textbf{PANN (Baseline [c=0.003, w=100, d=3])}} \\
RFF     & 19.410 (3.33)  & 4.237 (1.18)  & 0.535 (0.11) & 9.81e-03 (9.12e-03) &  7.00e+02 (2.20e+02) \\
GradNorm & 17.468 (6.70)  & 12.393 (2.20) & 1.753 (0.67) & 2.81e-04 (8.60e-07) &  5.10e+02 (1.50e+02) \\
ModMLP   & 23.515 (5.50)  & 9.676 (3.50)  & 1.463 (0.74) & 7.51e-03 (6.64e-03) &  1.50e+03 (4.10e+02) \\

\midrule
\midrule
\multicolumn{6}{l}{\textbf{Poisson Disk Points}} \\
\midrule
\multicolumn{6}{l}{\textbf{PL (c=$\;\cdot$)}} \\
c-0.003      & 34.391 (9.30) & 8.552 (0.15) & 4.529 (0.95) & 7.16e-06 (5.88e-06) &  8.00e+01 (4.50e+01) \\
c-0.004      & 34.391 (9.30) & 8.337 (0.15) & 4.374 (0.56) & 5.64e-06 (2.62e-06) &  8.00e+01 (4.50e+01) \\

\midrule
\multicolumn{6}{l}{\textbf{PINN (w=$\;\cdot$, d=$\;\cdot$)}} \\
w-100, d-3  & 17.109 (6.4) & 13.599 (2.04) & 1.780 (0.42) & 1.66e-01 (5.68e-02) &  2.50e+02 (2.10e+02) \\
w-50, d-3   & 12.684 (2.9) & 13.353 (7.02) & 2.640 (2.25) & 5.20e-01 (4.74e-02) &  2.50e+02 (2.10e+02) \\
w-50, d-5   & 9.360 (1.5)  & 6.852 (0.82)  & 2.126 (0.22) & 3.44e-01 (8.99e-02) &  4.10e+02 (3.00e+02) \\

\multicolumn{6}{l}{\textbf{PANN (PINN [c=$\;\cdot$, w=$\;\cdot$, d=$\;\cdot$])}} \\
c-0.003, w-100, d-3 & 8.243 (4.34)   & 17.156 (6.27) & 2.573 (1.12) & 8.28e-04 (1.84e-03) &  4.50e+02 (1.70e+02) \\
c-0.003, w-50, d-3  & 17.818 (10.01) & 18.843 (1.59) & 4.214 (2.79) & 5.10e-06 (2.72e-07) &  4.30e+02 (1.90e+02) \\
c-0.003, w-50, d-5  & 12.584 (5.11)  & 11.411 (5.40) & 1.627 (0.55) & 5.15e-06 (2.72e-07) &  6.10e+02 (2.50e+02) \\
c-0.004, w-100, d-3 & 8.243 (3.11)   & 18.301 (7.76) & 2.738 (1.11) & 1.22e-03 (1.83e-03) &  5.30e+02 (2.20e+02) \\
c-0.004, w-50, d-3  & 17.818 (10.01) & 18.819 (3.61) & 3.282 (0.95) & 1.32e-03 (2.93e-03) &  4.60e+02 (2.00e+02) \\
c-0.004, w-50, d-5  & 12.584 (5.11)  & 13.415 (4.41) & 1.997 (0.46) & 3.93e-04 (8.67e-04) &  6.80e+02 (3.40e+02) \\

\midrule
\multicolumn{6}{l}{\textbf{Baselines (w=100, d=3)}} \\
RFF                   & 18.124 (8.85)  & 5.955 (1.44) & 1.129 (0.10) & 1.25e-01 (2.80e-02) &  2.00e+02 (4.70e-02) \\
GradNorm              & 32.647 (8.75)  & 12.646 (4.10) & 1.969 (0.30) & 1.95e-01 (6.96e-02) &  1.10e+02 (2.30e+00) \\
ModMLP                & 25.881 (11.97) & 4.813 (1.68) & 1.854 (0.54) & 1.91e-01 (3.26e-02) &  6.00e+02 (1.80e-01) \\

\multicolumn{6}{l}{\textbf{PANN (Baseline [c=0.003, w=100, d=3])}} \\
RFF      & 9.485 (3.19)   & 5.611 (2.18) & 1.757 (0.33) & 2.41e-03 (1.32e-03) &  6.80e+02 (2.10e+02) \\
GradNorm & 32.064 (16.82) & 33.980 (14.16) & 2.996 (1.52) & 5.18e-04 (2.09e-07) &  5.20e+02 (1.50e+02) \\
ModMLP   & 20.473 (13.04) & 7.486 (3.81) & 2.345 (1.64) & \textbf{5.02e-06 (2.00e-07)} &  1.40e+03 (4.00e+02) \\

\bottomrule
\end{tabular}
\end{table*}

\begin{table*}[h!]
\centering
\caption{
\textbf{(2D Allen--Cahn)} Relative $\ell_2$ errors (mean and standard deviation in parentheses) and average runtime (s) for all PINN variations for $N=4096$. 
}
\label{tab:appendix-allencahn-errorstd}
\scriptsize
\renewcommand{\arraystretch}{1.2}
\setlength{\tabcolsep}{4pt}
\resizebox{\textwidth}{!}{%
\begin{tabular}{@{}lcccc|c@{}}
\toprule
\textbf{Equispaced Points} & \multicolumn{4}{c|}{\textbf{Relative $\mathbf{\ell_2}$ Errors}} & \textbf{Runtime (s)} \\
\cmidrule(lr){2-5}
& $N = 64$ & $N = 256$ & $N = 1024$ & $N = 4096$ & \\ 
\midrule

\midrule
\multicolumn{6}{l}{\textbf{PL (c=$\;\cdot$)}} \\
PL (c=0.003)     & 1.99e+00 (3.40e-01) & 5.99e-01 (8.17e-02) & 1.02e-01 (1.80e-02) & 1.50e-03 (1.03e-02) & 2.21e+02 (3.12e+01) \\
PL (c=0.004)     & 1.99e+00 (5.01e-01) & 5.99e-01 (8.17e-02) & 4.47e-02 (7.14e-04) & 3.34e-02 (1.83e-02) & 1.71e+02 (1.60e+01) \\

\midrule
\multicolumn{6}{l}{\textbf{PINN (w=$\;\cdot$, d=$\;\cdot$)}} \\
w=100, d=3          & 1.72e+00 (3.80e-01) & 3.35e-01 (2.68e-01) & 9.15e-01 (8.22e-01) & 1.20e+00 (2.82e-01) & 5.51e+00 (1.62e+00) \\
w=50, d=3           & 1.65e+00 (3.41e-01) & 7.89e-01 (6.67e-01) & 5.76e-01 (5.21e-01) & 4.70e-01 (6.46e-01) & 7.57e+01 (6.56e+01) \\
w=50, d=5           & 1.83e+00 (2.45e-01) & 1.52e+00 (3.63e-01) & 9.88e-01 (7.19e-01) & 7.77e-01 (6.54e-01) & 3.65e+01 (6.11e+01) \\

\multicolumn{6}{l}{\textbf{PANN (PINN [c=$\;\cdot$, w=$\;\cdot$, d=$\;\cdot$])}} \\
c=0.003, w=100, d=3 & 5.22e-01 (1.26e-03) & 1.11e-01 (5.44e-05) & 5.34e-03 (5.91e-05) & 7.65e-05 (4.96e-05) & 2.16e+02 (2.29e+01) \\
c=0.003, w=50, d=3  & 5.21e-01 (9.43e-04) & 1.11e-01 (6.97e-05) & 5.33e-03 (5.22e-05) & 2.15e-04 (1.07e-04) & 2.38e+02 (3.01e+01) \\
c=0.003, w=50, d=5  & 5.21e-01 (1.03e-03) & 1.12e-01 (7.67e-05) & 5.32e-03 (1.83e-05) & \textbf{7.20e-05 (2.21e-05)} & 2.11e+02 (3.08e+01) \\
c=0.004, w=100, d=3 & 5.22e-01 (1.26e-03) & 5.37e-02 (2.54e-05) & 9.30e-04 (6.74e-06) & 6.69e-04 (5.31e-05) & 3.14e+02 (2.55e+00) \\
c=0.004, w=50, d=3  & 5.21e-01 (9.43e-04) & \textbf{5.38e-02 (1.51e-04)} & 9.30e-04 (2.11e-05) & 4.02e-04 (1.90e-04) & 3.20e+02 (3.44e+01) \\
c=0.004, w=50, d=5  & 5.21e-01 (1.03e-03) & \textbf{5.38e-02 (7.36e-05)} & 9.19e-04 (1.36e-05) & 1.85e-04 (1.07e-04) & 2.93e+02 (2.90e+01) \\

\midrule
\multicolumn{6}{l}{\textbf{Baselines (w=100, d=3)}} \\
RFF                 & 5.76e-01 (1.26e-04) & 1.14e-01 (5.34e-02) & \textbf{5.06e-04 (1.80e-02)} & 5.92e-01 (1.80e-02) & 6.62e+02 (2.90e+01) \\
GradNorm            & 9.10e-01 (1.26e-03) & 6.64e-01 (1.94e-04) & 1.73e+00 (1.07e-04) & 7.36e-04 (4.07e-04) & 5.00e+02 (3.17e+02) \\
ModMLP              & 1.28e+00 (1.26e-02) & 1.15e+00 (6.44e-05) & 6.45e-01 (4.36e-05) & 4.45e-01 (4.36e-05) & 5.10e+03 (3.17e+02) \\

\multicolumn{6}{l}{\textbf{PANN (Baseline [c=0.003, w=100, d=3])}} \\
RFF       & \textbf{4.16e-01 (4.07e-03)} & 1.02e-01 (2.24e-02) & 2.23e-04 (1.03e-03) & 4.82e-01 (3.80e-03) & 7.98e+02 (3.08e+01) \\
GradNorm   & 8.05e-01 (7.89e-04) & 1.69e-01 (3.87e-02) & 1.02e-01 (2.98e-02) & 9.61e-04 (1.07e-03) & 1.19e+03 (1.17e+03) \\
ModMLP     & 2.23e+00 (1.26e-02) & 2.28e-01 (1.50e-02) & 5.15e-01 (1.80e-02) & 2.45e-01 (1.07e-04) & 5.72e+02 (1.80e+02) \\

\midrule
\midrule
\multicolumn{6}{l}{\textbf{Poisson Disk Points}} \\
\midrule
\multicolumn{6}{l}{\textbf{PL (c=$\;\cdot$)}} \\
PL (c-0.003)       &  1.51e+00 (3.28e-01) &  3.53e-01 (7.07e-06) &  1.66e-01 (1.33e-02) &  1.50e-02 (1.00e-02) &  2.20e+02 (3.02e+01) \\
PL (c-0.004)       &  1.51e+00 (1.08e-01) &  3.53e-01 (7.07e-06) &  3.66e-02 (6.33e-04) &  4.20e-02 (1.27e-01) & 2.86e+02 (1.60e+01) \\

\midrule
\multicolumn{6}{l}{\textbf{PINN (w=$\;\cdot$, d=$\;\cdot$)}} \\
w-100, d-3            &  1.94e+00 (2.04e-01) &  1.30e+00 (1.02e-01) &  1.34e+00 (7.62e-02) &  1.43e+00 (1.50e-01) &  5.53e+00 (1.88e+00) \\
w-50, d-3             &  1.99e+00 (3.28e-01) &  1.25e+00 (1.68e-01) &  9.78e-01 (1.04e-01) &  1.05e+00 (4.96e-01) &  5.47e+00 (2.29e+00) \\
w-50, d-5             &  1.66e+00 (8.02e-02) &  1.53e+00 (3.00e-01) &  1.39e+00 (1.85e-01) &  1.45e+00 (1.27e-01) &  5.76e+00 (1.57e+00) \\

\multicolumn{6}{l}{\textbf{PANN (PINN [c=$\;\cdot$, w=$\;\cdot$, d=$\;\cdot$])}} \\
c-0.003, w-100, d-3   &  9.94e-01 (8.89e-03) &  1.03e-01 (2.90e-05) &  1.25e-02 (3.60e-04) &  6.03e-03 (1.56e-03) &  2.02e+02 (2.23e+01) \\
c-0.003, w-50, d-3    &  1.02e+00 (4.07e-02) &  1.03e-01 (5.35e-05) &  1.21e-02 (2.49e-04) &  1.77e-03 (1.31e-03) &  2.86e+02 (1.20e+01) \\
c-0.003, w-50, d-5    &  1.03e+00 (1.36e-02) &  1.03e-01 (1.27e-05) &  1.25e-02 (5.00e-04) &  5.63e-03 (2.57e-03) &  1.98e+02 (1.71e+01) \\
c-0.004, w-100, d-3   &  9.94e-01 (8.89e-03) &  1.20e-01 (9.27e-05) &  1.52e-03 (7.01e-05) &  5.09e-03 (4.08e-04) &  3.31e+02 (9.89e-01) \\
c-0.004, w-50, d-3    &  1.02e+00 (4.07e-02) &  1.20e-01 (5.42e-05) &  1.51e-03 (6.72e-05) &  5.59e-03 (5.72e-04) &  3.45e+02 (1.79e+00) \\
c-0.004, w-50, d-5    &  1.03e+00 (1.36e-02) &  1.20e-01 (1.93e-04) &  1.48e-03 (6.46e-05) &  4.27e-03 (4.87e-04) &  3.26e+02 (2.58e+00) \\

\midrule
\multicolumn{6}{l}{\textbf{Baselines (w=100, d=3)}} \\
RFF                   &  7.25e-01 (2.78e-03) &  3.36e-01 (6.11e-03) &  \textbf{1.08e-02 (8.38e-04)} &  \textbf{4.51e-04 (5.52e-03)} &       7.68e+02 (2.88e+01) \\
GradNorm              &  1.18e+00 (8.66e-03) &  1.23e+00 (5.06e-03) &  6.14e-03 (6.59e-03) &  7.48e-04 (6.51e-03) &       4.78e+02 (4.90e+01) \\
ModMLP                &  1.49e+00 (7.65e-03) &  1.44e+00 (2.19e-03) &  5.13e-01 (8.19e-03) &  9.64e-01 (9.73e-03) &       1.40e+03 (2.01e+01) \\

\multicolumn{6}{l}{\textbf{PANN (Baselines [c=0.003, w=100, d=3])}} \\
RFF      &  \textbf{2.10e-02 (6.57e-03)} &  9.39e-02 (5.99e-03) &  8.83e-02 (3.91e-03) &  3.04e-02 (5.74e-03) &       9.98e+02 (2.90e+01) \\
GradNorm &  8.03e-02 (9.06e-03) &  2.56e-01 (1.92e-02) &  2.92e-02 (2.44e-02) &  4.95e-03 (9.05e-03) &       9.24e+02 (1.79e+01) \\
ModMLP   &  4.56e-02 (1.65e-03) &  \textbf{3.63e-02 (2.11e-04)} &  4.24e-02 (7.44e-03) &  6.94e-02 (3.02e-03) &       4.87e+03 (1.87e+02) \\

\bottomrule
\end{tabular}
}
\end{table*}

\begin{table*}[h!]
\centering
\caption{
\textbf{(Korteweg-De Vries)} Relative $\ell_2$ errors (mean and standard deviation in parentheses) and average runtime (s) for all PINN variations for $N=16384$.
}
\label{tab:appendix-kdv-errorstd}
\scriptsize
\renewcommand{\arraystretch}{1.2}
\setlength{\tabcolsep}{4pt}
\resizebox{\textwidth}{!}{%
\begin{tabular}{@{}lcccc|c@{}}
\toprule
\textbf{Equispaced Points} & \multicolumn{4}{c|}{\textbf{Relative $\mathbf{\ell_2}$ Errors}} & \textbf{Runtime (s)} \\
\cmidrule(lr){2-5}
& $N = 256$ & $N = 1024$ & $N = 4096$ & $N=16384$ &  \\ 
\midrule 

\multicolumn{6}{l}{\textbf{PINN (w=$\;\cdot$, d=$\;\cdot$)}} \\
w-100, d-3            &  \textbf{6.89e-01 (8.94e-03)} &  6.66e-01 (6.43e-02) &  7.16e-01 (3.01e-02) &  3.94e-01 (1.80e-01) &  7.92e+02 (1.10e+01) \\
w-50, d-3             &  7.02e-01 (8.33e-03) &  6.93e-01 (2.45e-02) &  6.94e-01 (8.16e-03) &  5.39e-01 (2.06e-01) &  7.75e+02 (2.27e+01) \\
w-50, d-5             &  6.89e-01 (4.49e-02) &  6.94e-01 (4.92e-02) &  6.94e-01 (6.29e-02) &  2.19e-01 (5.80e-02) &  1.19e+03 (3.88e+01) \\

\multicolumn{6}{l}{\textbf{PANN (PINN [c=$\;\cdot$, w=$\;\cdot$, d=$\;\cdot$])}} \\
c-0.0002, w-100, d-3   &  9.98e-01 (8.45e-02) &  7.18e-01 (8.13e-02) &  5.06e-01 (8.05e-02) &  1.34e-01 (9.15e-02) &  2.04e+03 (2.22e+02) \\
c-0.0002, w-50, d-3    &  1.14e+00 (9.77e-02) &  7.28e-01 (9.72e-02) &  6.96e-01 (8.09e-02) &  3.26e-01 (8.98e-02) &  1.34e+03 (2.15e+02) \\
c-0.0002, w-50, d-5    &  1.16e+00 (8.40e-02) &  7.75e-01 (6.19e-02) &  6.08e-01 (5.80e-02) &  1.68e-01 (9.87e-02) &  3.35e+03 (2.18e+02) \\
c-0.0004, w-100, d-3   &  1.15e+00 (5.65e-02) &  7.71e-01 (7.20e-02) &  6.63e-01 (9.42e-02) &  1.79e-01 (6.46e-02) &  2.65e+03 (2.75e+02) \\
c-0.0004, w-50, d-3    &  1.09e+00 (7.11e-02) &  7.96e-01 (5.49e-02) &  6.57e-01 (6.17e-02) &  2.44e-01 (6.41e-02) &  1.84e+03 (2.68e+02) \\
c-0.0004, w-50, d-5    &  1.05e+00 (7.60e-02) &  7.35e-01 (8.08e-02) &  5.81e-01 (9.71e-02) &  2.15e-01 (8.12e-02) &  1.34e+03 (2.90e+02) \\

\midrule
\multicolumn{6}{l}{\textbf{Baselines (w=100, d=3)}} \\
RFF                   &  7.02e-01 (1.69e-02) &  6.93e-01 (4.34e-02) &  7.26e-01 (6.42e-02) &  3.95e-01 (2.62e-01) &  7.36e+02 (2.56e+02) \\
GradNorm              &  6.90e-01 (2.35e-02) &  6.69e-01 (4.84e-02) &  7.19e-01 (3.70e-02) &  5.07e-01 (1.82e-01) &  7.87e+02 (1.26e+01) \\
ModMLP                &  7.95e-01 (1.88e-01) &  7.49e-01 (7.96e-02) &  6.95e-01 (5.04e-02) &  6.09e-01 (1.48e-01) &  2.52e+03 (3.74e+01) \\
Causal                &  9.34e-01 (3.59e-03) &  8.39e-01 (5.05e-03) &  6.51e-01 (4.27e-03) &  6.52e-01 (2.89e-03) &  6.00e+02 (8.70e-02) \\

\multicolumn{6}{l}{\textbf{PANN (Baseline [c=0.003, w=100, d=3])}} \\
RFF       &  1.08e+00 (9.94e-02) &  \textbf{6.49e-01 (9.60e-02)} &  \textbf{4.00e-01 (6.68e-02)} &  \textbf{1.05e-01 (5.15e-02)} &  3.90e+03 (2.23e+02) \\
GradNorm  &  1.07e+00 (8.36e-02) &  1.13e+00 (8.84e-02) &  8.67e-01 (7.34e-02) &  2.35e-01 (8.42e-02) &  2.90e+03 (2.51e+02) \\
ModMLP    &  1.11e+00 (5.47e-02) &  7.29e-01 (8.93e-02) &  6.15e-01 (8.09e-02) &  1.10e-01 (7.49e-02) &  3.36e+03 (2.94e+02) \\
Causal    &  1.19e+00 (7.07e-02) &  7.53e-01 (9.67e-02) &  4.56e-01 (7.00e-02) &  2.07e-01 (6.88e-02) &  1.11e+03 (2.57e+02) \\

\midrule
\midrule
\multicolumn{6}{l}{\textbf{Poisson Disk Points}} \\

\midrule
\multicolumn{6}{l}{\textbf{PINN (w=$\;\cdot$, d=$\;\cdot$)}} \\
w-100, d-3            &  \textbf{6.76e-01 (1.89e-02)} &  6.99e-01 (3.69e-02) &  7.89e-01 (2.64e-02) &  3.78e-01 (1.83e-01) &  7.86e+02 (1.52e+01) \\
w-50, d-3             &  6.81e-01 (2.19e-02) &  7.36e-01 (3.76e-02) &  7.46e-01 (4.96e-02) &  4.73e-01 (1.85e-01) &  7.89e+02 (1.15e+01) \\
w-50, d-5             &  8.84e-01 (1.31e-01) &  7.60e-01 (3.22e-02) &  6.17e-01 (1.36e-01) &  2.73e-01 (1.60e-01) &  1.16e+03 (2.76e+01) \\

\multicolumn{6}{l}{\textbf{PANN (PINN [c=$\;\cdot$, w=$\;\cdot$, d=$\;\cdot$])}} \\
c-0.003, w-100, d-3   &  1.17e+00 (8.33e-02) &  7.71e-01 (8.50e-02) &  5.39e-01 (6.10e-02) &  1.54e-01 (8.79e-02) &  2.92e+03 (2.97e+02) \\
c-0.003, w-50, d-3    &  1.18e+00 (6.09e-02) &  7.76e-01 (9.33e-02) &  6.80e-01 (6.19e-02) &  3.69e-01 (7.55e-02) &  3.61e+03 (2.92e+02) \\
c-0.003, w-50, d-5    &  1.16e+00 (8.46e-02) &  7.48e-01 (7.65e-02) &  5.66e-01 (7.08e-02) &  1.34e-01 (5.26e-02) &  3.57e+03 (2.30e+02) \\
c-0.004, w-100, d-3   &  1.01e+00 (7.73e-02) &  7.02e-01 (9.65e-02) &  6.69e-01 (9.56e-02) &  1.60e-01 (9.79e-02) &  2.83e+03 (2.55e+02) \\
c-0.004, w-50, d-3    &  1.03e+00 (6.86e-02) &  7.99e-01 (9.43e-02) &  6.99e-01 (5.73e-02) &  2.50e-01 (6.71e-02) &  1.04e+03 (2.60e+02) \\
c-0.004, w-50, d-5    &  1.10e+00 (9.24e-02) &  7.85e-01 (7.12e-02) &  5.79e-01 (9.31e-02) &  1.49e-01 (9.90e-02) &  2.30e+03 (2.89e+02) \\

\midrule
\multicolumn{6}{l}{\textbf{Baselines (w=100, d=3)}} \\
RFF                   &  7.52e-01 (8.47e-02) &  7.16e-01 (2.92e-02) &  7.44e-01 (6.31e-02) &  3.94e-01 (2.52e-01) &  7.35e+02 (2.56e+02) \\
GradNorm              &  6.93e-01 (3.26e-02) &  6.70e-01 (2.23e-02) &  8.15e-01 (2.65e-02) &  5.28e-01 (6.39e-02) &  7.86e+02 (8.95e+00) \\
ModMLP                &  7.67e-01 (9.67e-02) &  6.91e-01 (3.91e-02) &  6.91e-01 (4.42e-02) &  5.90e-01 (1.85e-01) &  2.51e+03 (5.77e+01) \\
Causal                &  1.09e+00 (1.53e-01) &  6.60e-01 (1.47e-03) &  6.53e-01 (2.45e-03) &  6.51e-01 (2.50e-03) &  6.00e+02 (2.31e-01) \\

\multicolumn{6}{l}{\textbf{PANN (Baselines [c=0.0002, w=100, d=3])}} \\
RFF       &  1.15e+00 (8.70e-02) &  \textbf{6.17e-01 (9.22e-02)} &  4.74e-01 (8.64e-02) &  \textbf{1.01e-01 (7.41e-02)} &  2.13e+03 (2.65e+02) \\
GradNorm  &  1.11e+00 (7.14e-02) &  1.14e+00 (7.41e-02) &  8.61e-01 (8.36e-02) &  3.43e-01 (9.35e-02) &  1.64e+03 (2.83e+02) \\
ModMLP    &  1.08e+00 (5.22e-02) &  7.47e-01 (8.70e-02) &  6.28e-01 (9.54e-02) &  1.59e-01 (5.55e-02) &  3.45e+03 (2.60e+02) \\
Causal    &  1.10e+00 (8.48e-02) &  7.18e-01 (9.31e-02) &  \textbf{4.04e-01 (9.79e-02)} &  2.41e-01 (6.88e-02) &  3.77e+03 (2.75e+02) \\

\bottomrule
\end{tabular}
}
\end{table*}

\begin{table*}[h!]
\centering
\caption{
\textbf{(Kuramoto-Sivashinsky)} Relative $\ell_2$ errors (mean and standard deviation in parentheses) and average runtime (s) for all PINN variations for $N=16382$. 
}
\label{tab:appendix-kuramoto-errorstd}
\scriptsize
\renewcommand{\arraystretch}{1.2}
\setlength{\tabcolsep}{4pt}
\resizebox{\textwidth}{!}{%
\begin{tabular}{@{}lcccc|c@{}}
\toprule
\textbf{Equispaced Points} & \multicolumn{4}{c|}{\textbf{Relative $\mathbf{\ell_2}$ Errors}} & \textbf{Runtime (s)} \\
\cmidrule(lr){2-5}
& $N = 256$ & $N = 1024$ & $N = 4096$ &  $N = 16384$ & \\ 
\midrule

\midrule
\multicolumn{6}{l}{\textbf{PINN (w=$\;\cdot$, d=$\;\cdot$)}} \\
w-100, d-3            &  1.18e+00 (1.12e-01) &  9.10e-01 (1.17e-01) &  8.98e-01 (1.14e-01) &  9.80e-01 (1.10e-01) &  1.76e+03 (2.43e+02) \\
w-50, d-3             &  1.08e+00 (9.12e-02) &  9.09e-01 (1.02e-01) &  8.23e-01 (9.35e-02) &  7.65e-01 (9.84e-02) &  1.59e+03 (2.85e+02) \\
w-50, d-5             &  1.19e+00 (1.08e-01) &  9.55e-01 (1.17e-01) &  8.75e-01 (1.12e-01) &  7.80e-01 (9.00e-02) &  6.54e+03 (2.71e+02) \\

\multicolumn{6}{l}{\textbf{PANN (PINN [c=$\;\cdot$, w=$\;\cdot$, d=$\;\cdot$])}} \\
c-0.0001, w-100, d-3   &  1.07e+00 (1.10e-01) &  9.93e-01 (1.14e-01) &  9.15e-01 (1.09e-01) &  7.25e-01 (9.79e-02) &  2.26e+03 (2.36e+02) \\
c-0.0001, w-50, d-3    &  1.12e+00 (1.03e-01) &  9.02e-01 (1.16e-01) &  6.37e-01 (9.07e-02) &  \textbf{5.70e-01 (1.18e-01)} &  2.26e+03 (2.68e+02) \\
c-0.0001, w-50, d-5    &  1.05e+00 (1.14e-01)  &  9.86e-01 (1.02e-01) &  9.34e-01 (9.79e-02) &  5.94e-01 (1.12e-01) &  2.26e+03 (2.53e+02) \\
c-0.0002, w-100, d-3   &  9.95e-01 (1.16e-01) &  9.84e-01 (1.04e-01) &  9.26e-01 (9.70e-02) &  6.97e-01 (1.10e-01) &  1.85e+03 (2.81e+02) \\
c-0.0002, w-50, d-3    &  1.07e+00 (1.14e-01) &  9.81e-01 (1.17e-01) &  9.90e-01 (1.14e-01) &  6.87e-01 (9.84e-02) &  2.26e+03 (2.43e+02) \\
c-0.0002, w-50, d-5    &  1.09e+00 (1.15e-01) &  9.83e-01 (1.16e-01) &  9.80e-01 (1.08e-01) &  6.97e-01 (1.01e-01) &  2.26e+03 (2.99e+02) \\

\midrule
\multicolumn{6}{l}{\textbf{Baselines (w=100, d=3)}} \\
RFF                   &  1.10e+00 (1.20e-01) &  9.84e-01 (1.06e-01) &  8.91e-01 (1.04e-01) &  6.90e-01 (1.13e-01) &  2.76e+03 (2.15e+02) \\
GradNorm              &  1.13e+00 (1.13e-01) &  9.21e-01 (9.50e-02) &  8.16e-01 (1.02e-01) &  7.60e-01 (1.19e-01) &  2.76e+03 (2.74e+02) \\
ModMLP                &  1.19e+00 (1.08e-01) &  9.35e-01 (1.10e-01) &  8.65e-01 (1.13e-01) &  7.71e-01 (1.12e-01) &  2.76e+03 (2.47e+02) \\
Causal                &  1.15e+00 (1.07e-01) &  9.39e-01 (1.20e-01) &  8.98e-01 (1.18e-01) &  7.70e-01 (9.70e-02) &  2.76e+03 (2.34e+02) \\

\multicolumn{6}{l}{\textbf{PANN (Baseline [c=0.0002, w=100, d=3])}} \\
RFF      &  1.16e+00 (9.32e-02) &  \textbf{9.07e-01 (1.03e-01)} &  8.89e-01 (1.06e-01) &  7.00e-01 (1.03e-01) &  9.63e+03 (2.85e+02) \\
GradNorm &  1.08e+00 (1.20e-01) &  1.20e+00 (1.18e-01) &  \textbf{8.21e-01 (1.00e-01)} &  7.09e-01 (9.09e-02) &  3.07e+03 (2.22e+02) \\
ModMLP   &  1.13e+00 (1.05e-01) &  1.12e+00 (1.10e-01) &  8.58e-01 (9.32e-02) &  6.50e-01 (1.03e-01) &  3.05e+03 (2.24e+02) \\
Causal   &  1.01e+00 (1.14e-01) &  1.04e+00 (1.14e-01) &  8.70e-01 (9.61e-02) &  6.21e-01 (9.08e-02) &  4.15e+03 (2.40e+02) \\

\midrule
\midrule

\multicolumn{6}{l}{\textbf{Poisson Disk Points}} \\
\midrule
\multicolumn{6}{l}{\textbf{PINN (w=$\;\cdot$, d=$\;\cdot$)}} \\
w-100, d-3            &  1.09e+00 (9.09e-02) &  9.75e-01 (1.02e-01) &  8.74e-01 (1.16e-01) &  9.01e-01 (1.15e-01) &  1.46e+03 (2.04e+02) \\
w-50, d-3             &  1.03e+00 (1.04e-01) &  9.80e-01 (1.13e-01) &  8.75e-01 (1.12e-01) &  7.75e-01 (1.00e-01) &  1.47e+03 (2.84e+02) \\
w-50, d-5             &  1.03e+00 (1.04e-01) &  9.80e-01 (1.13e-01) &  8.75e-01 (1.12e-01) &  7.90e-01 (1.00e-01) &  3.78e+03 (2.84e+02) \\

\multicolumn{6}{l}{\textbf{PANN (PINN [c=$\;\cdot$, w=$\;\cdot$, d=$\;\cdot$])}} \\
c-0.0001, w-100, d-3   &  1.08e+00 (9.40e-02) &  \textbf{9.43e-01 (9.45e-02)} &  9.60e-01 (1.05e-01) &  7.42e-01 (1.10e-01) &  3.70e+03 (2.98e+02) \\
c-0.0001, w-50, d-3    &  1.19e+00 (9.94e-02) &  9.88e-01 (1.11e-01) &  9.67e-01 (1.20e-01) &  6.15e-01 (1.14e-01) &  3.77e+03 (3.12e+02) \\
c-0.0001, w-50, d-5    &  1.15e+00 (9.92e-02) &  9.89e-01 (1.01e-01) &  9.59e-01 (1.03e-01) &  6.82e-01 (1.15e-01) &  3.55e+03 (3.98e+02) \\
c-0.0001, w-100, d-3   &  1.05e+00 (9.03e-02) &  9.82e-01 (9.09e-02) &  9.06e-01 (1.07e-01) &  6.95e-01 (9.85e-02) &  2.99e+03 (4.70e+02) \\
c-0.0002, w-50, d-3    &  1.13e+00 (1.16e-01) &  9.85e-01 (1.10e-01) &  9.80e-01 (1.11e-01) &  6.74e-01 (1.09e-01) &  2.15e+03 (2.08e+02) \\
c-0.0002, w-50, d-5    &  1.15e+00 (1.09e-01) &  9.80e-01 (1.17e-01) &  9.88e-01 (1.03e-01) &  7.00e-01 (1.04e-01) &  5.63e+03 (1.98e+02) \\

\midrule
\multicolumn{6}{l}{\textbf{Baselines (w=100, d=3)}} \\
RFF                   &  1.03e+00 (1.04e-01) &  9.80e-01 (1.45e-01) &  8.55e-01 (1.12e-01) &  7.04e-01 (1.00e-01) &  2.13e+03 (1.98e+02) \\
GradNorm              &  1.03e+00 (1.04e-01) &  9.80e-01 (1.87e-01) &  9.75e-01 (2.04e-01) &  7.50e-01 (1.00e-01) &  2.94e+03 (2.84e+02) \\
ModMLP                &  1.03e+00 (1.04e-01) &  9.80e-01 (2.13e-01) &  8.80e-01 (1.57e-01) &  7.82e-01 (1.00e-01) &  2.31e+03 (1.98e+02) \\
Causal                &  1.03e+00 (1.04e-01) &  9.90e-01 (1.03e-01) &  8.75e-01 (2.12e-01) &  7.37e-01 (1.00e-01) &  1.99e+03 (1.98e+02) \\

\multicolumn{6}{l}{\textbf{PANN (Baseline [c=0.0002, w=100, d=3])}} \\
RFF       &  1.03e+00 (1.04e-01) &  9.80e-01 (1.13e-01) &  8.75e-01 (2.72e-01) &  7.10e-01 (1.00e-01) &  4.41e+03 (2.84e+02) \\
GradNorm  &  1.15e+00 (1.12e-01) &  1.06e+00 (1.15e-01) &  9.66e-01 (1.12e-01) &  6.80e-01 (1.04e-01) &  3.90e+03 (1.98e+02) \\
ModMLP    &  1.15e+00 (1.12e-01) &  1.02e+00 (1.11e-01) &  8.76e-01 (2.02e-01) &  \textbf{6.11e-01 (1.04e-01)} &  4.00e+03 (1.98e+02) \\
Causal    &  1.15e+00 (1.12e-01) &  1.06e+00 (1.13e-01) &  \textbf{7.68e-01 (1.09e-01)} &  6.31e-01 (1.04e-01) &  4.30e+03 (1.98e+02) \\

\bottomrule
\end{tabular}
}
\end{table*}

\begin{table*}[h!]
\centering
\caption{
\textbf{(Grey-Scott)} Relative $\ell_2$ errors (mean and standard deviation in parentheses) and average runtime (s) for all PINN variations for $N=131072$.
}
\label{tab:appendix-greyscott-errorstd}
\scriptsize
\renewcommand{\arraystretch}{1.1}
\setlength{\tabcolsep}{4pt}
\begin{tabular}{@{}lcccc|c@{}}
\toprule
\textbf{$u$ Predictions} & \multicolumn{3}{c|}{\textbf{Relative $\mathbf{\ell_2}$ Errors}} & \textbf{Runtime (s)} \\
\cmidrule(lr){2-4}
& $N = 8192$ & $N = 32768$ & $N = 131072$ & \\ 
\midrule

\midrule
\multicolumn{5}{l}{\textbf{PINN (w=$\;\cdot$, d=$\;\cdot$)}} \\
w-100, d-3            &  3.88e-01 (1.86e-01) &  3.91e-01 (1.28e-01) &  3.82e-01 (1.89e-01) &  2.60e+03 (1.30e+02) \\
w-50, d-3             &  3.95e-01 (1.86e-01) &  3.97e-01 (1.28e-01) &  3.97e-01 (1.89e-01) &  2.01e+03 (1.30e+02) \\
w-50, d-5             &  3.97e-01 (1.86e-01) &  4.01e-01 (1.28e-01) &  6.00e-01 (1.89e-01) &  3.41e+03 (1.30e+02) \\

\multicolumn{5}{l}{\textbf{PANN (PINN [c=$\;\cdot$, w=$\;\cdot$, d=$\;\cdot$])}} \\
c-0.001, w-100, d-3   &  2.45e-01 (1.36e-01) &  2.06e-01 (1.61e-01) &  1.86e-01 (1.64e-01) &  5.77e+03 (1.66e+02) \\
c-0.001, w-50, d-3    &  2.95e-01 (1.74e-01) &  2.17e-01 (1.21e-01) &  2.27e-01 (1.42e-01) &  4.26e+03 (1.51e+02) \\
c-0.001, w-50, d-5    &  2.97e-01 (1.61e-01) &  2.11e-01 (1.88e-01) &  \textbf{1.43e-01 (1.29e-01)} &  7.23e+03 (1.86e+02) \\
c-0.002, w-100, d-3   &  \textbf{2.55e-01 (1.29e-01)} &  2.32e-02 (1.93e-01) &  2.52e-02 (1.41e-01) &  6.50e+03 (1.47e+02) \\
c-0.002, w-50, d-3    &  2.60e-01 (1.94e-01) &  5.23e-01 (1.15e-01) &  2.23e-01 (1.36e-01) &  5.65e+03 (1.05e+02) \\
c-0.002, w-50, d-5    &  2.78e-01 (1.62e-01) &  2.32e-01 (1.57e-01) &  1.95e-01 (1.74e-01) &  8.24e+03 (1.02e+02) \\

\midrule
\multicolumn{5}{l}{\textbf{Baselines (w=100, d=3)}} \\
RFF                   &  5.98e-01 (1.86e-01) &  5.98e-01 (1.28e-01) &  4.82e-01 (1.89e-01) &  2.69e+03 (1.30e+02) \\
GradNorm              &  5.96e-01 (1.79e-01) &  5.97e-01 (1.42e-01) &  5.96e-01 (1.52e-01) &  2.59e+03 (1.43e+02) \\
ModMLP                &  3.96e-01 (1.64e-01) &  3.97e-01 (1.62e-01) &  3.97e-01 (1.64e-01) &  2.70e+03 (1.97e+02) \\
Causal                &  3.90e-01 (1.74e-01) &  3.89e-01 (1.41e-01) &  3.85e-01 (1.06e-01) &  2.63e+03 (1.13e+02) \\

\multicolumn{5}{l}{\textbf{PANN (Baseline, [c=0.001, w=100, d=3])}} \\
RFF      &  5.59e-01 (1.79e-01) &  5.70e-01 (1.52e-01) &  4.24e-01 (1.94e-01) &  5.72e+03 (1.12e+02) \\
GradNorm &  5.45e-01 (1.36e-01) &  4.13e-01 (1.61e-01) &  4.71e-01 (1.64e-01) &  5.76e+03 (1.66e+02) \\
ModMLP   &  5.79e-01 (1.79e-01) &  4.90e-01 (1.52e-01) &  2.20e-01 (1.94e-01) &  5.70e+03 (1.12e+02) \\
Causal   &  3.13e-01 (1.95e-01) &  \textbf{2.02e-01 (1.17e-01)} &  1.71e-01 (1.27e-01) &  5.52e+03 (1.42e+02) \\

\midrule
\midrule
\multicolumn{5}{l}{\textbf{$v$ Predictions}} \\

\midrule
\multicolumn{5}{l}{\textbf{PINN (w=$\;\cdot$, d=$\;\cdot$)}} \\
w-100, d-3            &  7.26e-01 (1.99e-01) &  7.29e-01 (1.47e-01) &  7.19e-01 (1.79e-01) &  2.60e+03 (1.89e+02) \\
w-50, d-3             &  7.34e-01 (1.99e-01) &  7.37e-01 (1.47e-01) &  7.37e-01 (1.79e-01) &  2.01e+03 (1.89e+02) \\
w-50, d-5             &  7.38e-01 (1.99e-01) &  7.42e-01 (1.47e-01) &  1.00e+00 (1.79e-01) &  3.41e+03 (1.89e+02) \\

\multicolumn{5}{l}{\textbf{PANN (PINN [c=$\;\cdot$, w=$\;\cdot$, d=$\;\cdot$])}} \\
c-0.001, w-100, d-3   &  6.62e-01 (1.88e-01) &  2.97e-01 (1.61e-01) &  3.59e-01 (1.88e-01) &  5.77e+03 (1.82e+02) \\
c-0.001, w-50, d-3    &  6.04e-01 (1.83e-01) &  5.45e-01 (1.74e-01) &  6.05e-01 (1.70e-01) &  4.26e+03 (1.57e+02) \\
c-0.001, w-50, d-5    &  5.29e-01 (1.93e-01) &  5.34e-01 (1.41e-01) &  \textbf{3.06e-01 (1.13e-01)} &  7.23e+03 (1.37e+02) \\
c-0.002, w-100, d-3   &  \textbf{5.06e-01 (1.13e-01)} &  \textbf{3.78e-01 (1.62e-01)} &  2.99e-01 (1.57e-01) &  6.50e+03 (1.13e+02) \\
c-0.002, w-50, d-3    &  5.13e-01 (1.61e-01) &  6.71e-01 (1.64e-01) &  6.25e-01 (1.88e-01) &  5.65e+03 (1.64e+02) \\
c-0.002, w-50, d-5    &  5.17e-01 (1.21e-01) &  4.27e-01 (1.42e-01) &  3.52e-01 (1.83e-01) &  8.24e+03 (1.78e+02) \\

\midrule
\multicolumn{5}{l}{\textbf{Baselines (w=100, d=3)}} \\
RFF                   &  9.96e-01 (1.99e-01) &  9.96e-01 (1.47e-01) &  8.59e-01 (1.79e-01) &  2.69e+03 (1.89e+02) \\
GradNorm              &  9.90e-01 (1.20e-01) &  9.92e-01 (1.12e-01) &  9.90e-01 (1.15e-01) &  2.59e+03 (1.67e+02) \\
ModMLP                &  7.38e-01 (1.61e-01) &  7.38e-01 (1.11e-01) &  7.38e-01 (1.82e-01) &  2.70e+03 (1.55e+02) \\
Causal                &  7.29e-01 (1.37e-01) &  7.28e-01 (1.62e-01) &  7.23e-01 (1.64e-01) &  2.63e+03 (1.57e+02) \\

\multicolumn{5}{l}{\textbf{PANN (Baseline, [c=0.001, w=100, d=3])}} \\
RFF       &  8.12e-01 (1.15e-01) &  7.91e-01 (1.36e-01) &  8.04e-01 (1.61e-01) &  5.72e+03 (1.83e+02) \\
GradNorm  &  8.62e-01 (1.88e-01) &  6.97e-01 (1.61e-01) &  6.11e-01 (1.88e-01) &  5.76e+03 (1.82e+02) \\
ModMLP    &  8.22e-01 (1.15e-01) &  7.82e-01 (1.36e-01) &  3.06e-01 (1.61e-01) &  3.46e+03 (1.86e+02) \\
Causal    &  6.51e-01 (1.52e-01) &  6.56e-01 (1.22e-01) &  3.20e-01 (1.53e-01) &  5.52e+03 (1.70e+02) \\

\bottomrule
\end{tabular}
\end{table*}

\bibliographystyle{siamplain}
\bibliography{references}

\end{document}